\documentclass{article}

\usepackage[main, final]{neurips_2026}

\usepackage[utf8]{inputenc} %
\usepackage[T1]{fontenc}    %
\usepackage{hyperref}       %
\usepackage{url}            %
\usepackage{booktabs}       %
\usepackage{amsfonts}       %
\usepackage{nicefrac}       %
\usepackage{microtype}      %
\usepackage{xcolor}         %

\usepackage{microtype}
\usepackage{graphicx}
\usepackage{subcaption}
\usepackage{booktabs}
\usepackage{hyperref}
\hypersetup{
    colorlinks,
    linkcolor={blue!60!black},
    citecolor={cyan!80!black},
    urlcolor={magenta!60!black}
}

\usepackage{amsmath}
\usepackage{amssymb}
\usepackage{mathtools}

\usepackage{amsthm}

\usepackage[capitalize,noabbrev]{cleveref}

\usepackage{nameref}

\usepackage[framemethod=TikZ]{mdframed}

\usepackage{algorithm}
\usepackage{algorithmic}

\definecolor{theoremcolor}{rgb}{0.931176471, 0.958627451, 1.0}

\mdfsetup{
    backgroundcolor=theoremcolor,
    linecolor=black,      %
    linewidth=0.5pt,      %
    frametitlebackgroundcolor=theoremcolor, %
    leftline=true,
    topline=true,
    rightline=true,
    bottomline=true,
    skipabove=10pt,
    skipbelow=10pt,
    roundcorner=2pt       %
}

\newmdtheoremenv{definition}{Definition}
\newmdtheoremenv{proposition}{Proposition}
\newmdtheoremenv{corollary}{Corollary}
\newmdtheoremenv{theorem}{Theorem}
\newmdtheoremenv{lemma}{Lemma}
\newmdtheoremenv{example}{Example}
\newmdtheoremenv{remark}{Remark}

\usepackage[dvipsnames]{xcolor}
\usepackage{bbm}
\usepackage{tikz}

\usepackage{makecell}

\usepackage{enumitem}

\usetikzlibrary{matrix, arrows.meta}

\usepackage{amsmath,amsfonts,bm}

\def\1{\bm{1}}
\def\0{\bm{0}}

\def\RR{{\mathbb{R}}}
\def\NN{{\mathbb{N}}}
\def\EE{{\mathbb{E}}}

\def\conv{{\mathsf{conv}}}

\def\cE{{\mathcal{E}}}

\def\cG{{\mathcal{G}}}

\def\cL{{\mathcal{L}}}

\def\cN{{\mathcal{N}}}
\def\cO{{\mathcal{O}}}
\def\cP{{\mathcal{P}}}

\def\cS{{\mathcal{S}}}

\def\cV{{\mathcal{V}}}

\def\cY{{\mathcal{Y}}}

\def\thetav{{\bm{\theta}}}
\def\Thetav{{\bm{\Theta}}}
\def\muv{{\bm{\mu}}}
\def\sigmav{{\bm{\sigma}}}
\def\etav{{\bm{\eta}}}
\def\phiv{{\bm{\phi}}}

\def\e{{\bm{e}}}

\def\k{{\bm{k}}}

\def\q{{\bm{q}}}

\def\s{{\bm{s}}}

\def\u{{\bm{u}}}
\def\v{{\bm{v}}}
\def\w{{\bm{w}}}
\def\x{{\bm{x}}}
\def\y{{\bm{y}}}
\def\z{{\bm{z}}}

\def\B{{\bm{B}}}

\def\I{{\bm{I}}}

\def\M{{\bm{M}}}

\def\Z{{\bm{Z}}}

\def\maxo{\mathsf{max}_\Omega}

\def\ri{\mathsf{ri}}
\def\range{\mathsf{range}}

\def\Vw{V^{\bm{w}}}

\def\Vwo{V^{\bm{w}}_{\Omega}}
\def\VVwo{\bm{V}^{\bm{w}}_{\Omega}}
\def\Vdwo{\dot{V}^{\bm{w}}_{\Omega}}
\def\VVdwo{\dot{\bm{V}}^{\bm{w}}_{\Omega}}
\def\Ew{E^{\bm{w}}}
\def\EEw{\bm{E}^{\bm{w}}}
\def\Ewo{E^{\bm{w}}_{\Omega}}
\def\EEwo{\bm{E}^{\bm{w}}_{\Omega}}
\def\Edwo{\dot{E}^{\bm{w}}_{\Omega}}
\def\EEdwo{\dot{\bm{E}}^{\bm{w}}_{\Omega}}
\def\Qw{Q^{\bm{w}}}
\def\QQw{\bm{Q}^{\bm{w}}}
\def\Qwo{Q^{\bm{w}}_{\Omega}}
\def\QQwo{\bm{Q}^{\bm{w}}_{\Omega}}

\def\Vk{V^{\bm{1}}}
\def\Vko{V^{\bm{1}}_{\Omega}}

\def\yk{{\bm{y}^k_{\bm{1}}}}
\def\Yk{{\mathcal{Y}^k_{\bm{1}}}}
\def\maxk{{\mathsf{max}^k_{\bm{1}}}}
\def\yko{\bm{y}^k_{\bm{1},\Omega}}
\def\maxko{\mathsf{max}^k_{\bm{1},\Omega}}

\def\ycw{{\bm{y}^C_{\bm{w}}}}
\def\Ycw{{\mathcal{Y}^C_{\bm{w}}}}

\def\maxcw{{\mathsf{max}^C_{\bm{w}}}}
\def\ycwo{\bm{y}^C_{\bm{w},\Omega}}
\def\maxcwo{\mathsf{max}^C_{\bm{w},\Omega}}

\def\picwo{\pi^{\bm{w},C}_{\bm{\theta},\Omega}}
\def\piko{\pi^{\bm{1},k}_{\bm{\theta},\Omega}}

\def\Omegacw{{\Omega^C_{\bm{w}}}}
\def\Omegak{{\Omega^k_{\bm{1}}}}

\DeclareMathOperator*{\argmax}{\mathsf{argmax}}

\DeclareMathOperator{\clip}{{\mathsf{clip}}}

\title{Differentiable Knapsack and Top-$\k$ Operators via Dynamic Programming}

\author{
  Germain Vivier-Ardisson \\
  Google DeepMind \& CERMICS\\
  Paris, France\\
  \texttt{gvivier@google.com}\\
  \And
  Michaël E. Sander \\
  Google DeepMind\\
  Paris, France \\
  \texttt{michaelsander@google.com}\\
  \AND
  Axel Parmentier \\
  CERMICS, CNRS, ENPC, IPP\\
  Marne-la-Vallée, France\\
  \texttt{axel.parmentier@enpc.fr}\\
  \And
  Mathieu Blondel \\
  Google DeepMind\\
  Paris, France\\
  \texttt{mblondel@google.com}
}

\begin{document}

\maketitle

\begin{abstract}
Knapsack and Top-$k$ operators are useful for selecting discrete subsets of variables. However, their integration into neural networks is challenging as they are piecewise constant, yielding gradients that are zero almost everywhere. 
In this paper, we propose a unified framework casting these operators as dynamic programs, and derive differentiable relaxations by smoothing the underlying recursions. 
On the algorithmic side, we develop efficient parallel algorithms supporting both deterministic and stochastic forward passes, and vector-Jacobian products for the backward pass.
On the theoretical side, we prove that Shannon's negative entropy is the unique separable regularization choice that yields permutation-equivariant operators, and characterize regularizers inducing sparse selections. 
On the experimental side, we demonstrate our framework on a benchmark on learning to predict Knapsack solutions, an extension of discrete VAEs, and a constrained dynamic assortment RL problem.
\end{abstract}

\section{Introduction}
\label{sec:intro}

Many learning tasks rely on selecting a discrete subset of variables, or items, whether to learn sparse latent representations or to optimize resource allocation. The standard Top-$k$ operator selects subsets under a fixed cardinality constraint, and the \emph{Knapsack} problem \citep{martello_knapsack_1990} generalizes this structure by supporting non-uniform item weights, enforcing a total capacity limit on their sum. While the structure of these constraints provides very useful inductive bias, the resulting operators are piecewise constant, with either zero or undefined Jacobian. As a result, integrating them into neural networks is challenging as they break the differentiable computation graph, preventing the backpropagation of meaningful gradients.

\begin{figure*}[t]
    \centering
    \includegraphics[width=0.99\textwidth]{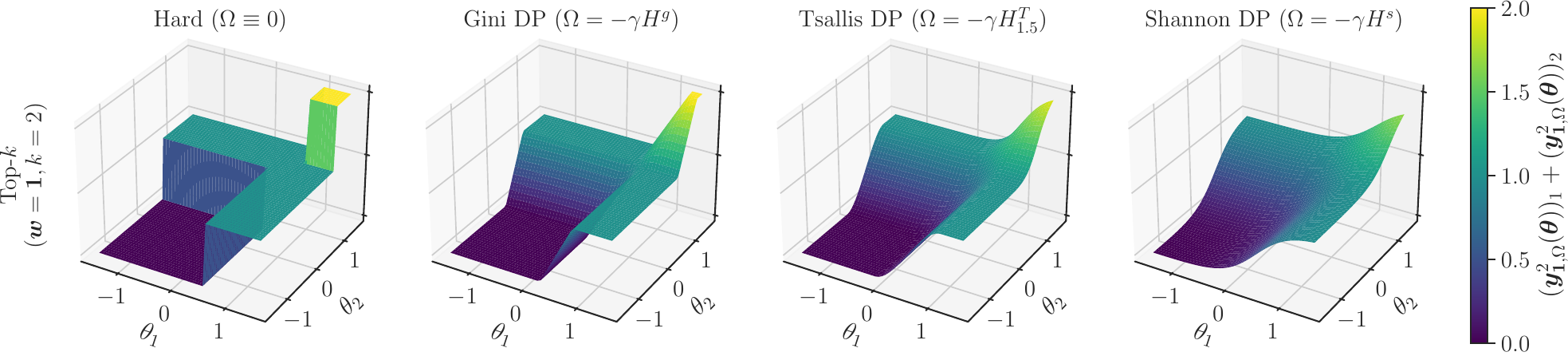}
    \par
    \includegraphics[width=0.99\textwidth]{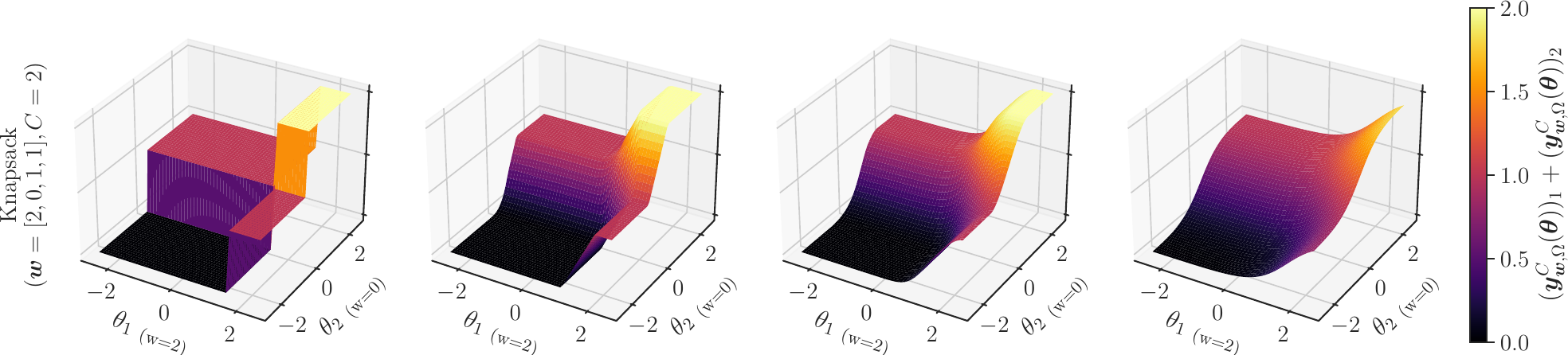}
    \caption{\textbf{Illustration of our relaxed Top-$k$ and Knapsack operators.} We plot the sum of the first two coordinates of the operator for $\thetav=(\theta_1,\theta_2,\frac{1}{2},1)^\top$. In the first row, we use $\yko$ with $k=2$ and $\w=\1$. In the second row, we use $\ycwo$ with $C=2$ and $\w=(2,0,1,1)^\top$, breaking the symmetry in $\theta_1$ and $\theta_2$. Using $\Omega\equiv0$ yields piecewise-constant outputs. Gini and Tsallis regularization yield a sparse and differentiable operator, while Shannon regularization yields a dense and differentiable one.}
    \label{fig:topk_knapsack_3D}
\end{figure*}

To bridge this gap, current research often relies on continuous relaxations or regularization for gradient estimation. Solver-agnostic, \emph{black-box} approaches, such as perturbation with stochastic noise \citep{berthet_learning_2020, niepert_implicit_2021} or piecewise affine interpolation \citep{pogancic_differentiation_2020}, are flexible but computationally expensive, often requiring multiple solver calls. Conversely, specialized differentiable layers for ranking and Top-$k$ operations, often based on optimal transport or regularized linear programming \citep{cuturi_differentiable_2019, blondel_fast_2020}, can be efficient but lack the generality to handle the non-uniform weights inherent to Knapsack constraints.
Furthermore, many existing methods restrict the choice of regularization (e.g., to Shannon entropy), limiting control over key properties like sparsity.

Moreover, most existing relaxations are purely deterministic (yielding a soft, expected selection in the forward pass), whereas stochastic sampling over feasible subsets serves two essential purposes. First, conditional-computation pipelines \citep{li_knapsack_2026, cha_knapspec_2026} strictly require hard $0/1$ selections, as a computation is either executed or skipped. Second, generative and exploratory paradigms, such as in discrete VAEs \citep{rolfe_discrete_2016,ahmed_simple_2023} or structured reinforcement learning \citep{hoppe_structured_2025}, inherently rely on sampling discrete solutions from a tractable distribution.

In this work, we address these limitations by revisiting the structural similarities between Knapsack and Top-$k$ problems, casting both as instances of dynamic programming (DP) and building upon \citet{mensch_differentiable_2018} to propose a unified differentiable formulation. In the Knapsack setting, we introduce the first specialized differentiable layer for this problem: this unlocks the end-to-end training of models that require weighted, capacity-constrained discrete selections, while allowing them to benefit from the full diversity of our framework (supporting both stochastic and deterministic forward passes, as well as sparse and equivariant modes). In the Top-$k$ setting, we provide the first unified framework for differentiable operators that natively jointly combines exact sparsity (or permutation equivariance), exact VJPs, and exact sampling.

Specifically, we make the following technical contributions:
\begin{itemize}[topsep=0pt, itemsep=1pt, parsep=3pt, leftmargin=1em]
    \item In \cref{sec:algorithmic_smoothing}, we derive differentiable relaxations of the Knapsack and Top-$k$ problems by regularizing the $\max$ operators within the underlying Bellman DP recursions.
    \item In \cref{sec:regularization_choice}, we show that Shannon's entropy is the \emph{unique} separable regularizer yielding \emph{permutation-equivariant} operators, and characterize those inducing \emph{sparsity} (\cref{prop:equivariance,prop:sparsity}).
    \item In \cref{sec:continuous_forward}, we derive efficient vector-Jacobian product computations, allowing the integration of the proposed operators into differentiable programming pipelines.
    \item In \cref{sec:stochastic_forward}, we prove the existence of a distribution underlying the proposed operators (\cref{prop:distribution}),  provide an ancestral sampling algorithm for stochastic forward passes, and propose unbiased first-order surrogate gradients to bypass the non-differentiability of sampling.
    \item We provide a principled supervised learning approach using our operators as output layers in \cref{sec:output_layer}, by computing gradients of associated Fenchel-Young losses \citep{blondel_learning_2020}.
    \item In \cref{sec:experiments}, we evaluate our DP-based losses on a knapsack problem, benchmark our proposed differentiable Top-$k$ operators on a Fenchel-Young extension of discrete VAEs, and evaluate our Knapsack operators on a constrained dynamic assortment RL problem.
\end{itemize}
Code is available at \,\url{https://github.com/google-deepmind/diff-knapsack}.

\section{Background and Related Work}
\label{sec:background_and_related_work}

\paragraph{Knapsack operators.}

The 0/1 Knapsack problem for $n$ items with values given by $\thetav \in\RR^n$, weights $\w\in\NN^n$, and capacity $C\in\NN$ such that $w_i\leq C\;\forall i\in[n]$, corresponds to %
finding a subset of items with maximal value and a total weight restricted by $C$ \citep{martello_knapsack_1990, kellerer_knapsack_2004}:
\begin{equation}
\label{pb:knapsack}
    \max_{\y\in\{0,1\}^n} \sum_{i=1}^n \theta_i y_i\quad \text{s.t.} \;\sum_{i=1}^n w_i y_i\leq C\,.
\end{equation}
We denote by $\Ycw\triangleq\left\{ \y\in\{0,1\}^n\mid \langle\w,\y\rangle\leq C \right\}$ the set of feasible item selections. %
The optimal value of Problem~(\ref{pb:knapsack}) is denoted by $\maxcw(\thetav)\triangleq \max_{\y\in\Ycw}\langle\thetav,\y\rangle$, and $\ycw(\thetav)\triangleq\argmax_{\y\in\Ycw}\langle\thetav,\y\rangle$ is the corresponding maximizer, or \emph{Knapsack operator}. Danskin's theorem \citep{danskin_theory_1966} gives $\nabla\maxcw(\thetav)=\ycw(\thetav)$ when the maximizer is unique. However, since $\ycw$ is piecewise constant, its Jacobian $\nabla_\thetav\ycw(\thetav)$ is either zero or undefined, preventing the backpropagation of gradients.

\paragraph{Top-$k$ operators.}

The variational form of the Top-$k$ selection problem is a notable slight variation of Problem~(\ref{pb:knapsack}), where item weights are given by $\w=\1$, and the capacity is $C=k$ for some $k$ such that $1\leq k<n$. Moreover, the inequality constraint becomes an equality, so that the problem writes:
\begin{equation}
    \label{pb:topk}
    \max_{\y\in\{0,1\}^n} \sum_{i=1}^n \theta_i y_i \quad \text{s.t.}\; \sum_{i=1}^n y_i=k\,.
\end{equation}
With a slight abuse of notation (overloading Knapsack notations for $\w=\1$ and $C=k$), we respectively denote by $\Yk\,$, $\maxk\,$, and $\yk$ the feasible set, value and operator in this case. Note that $\yk(\thetav)=(\mathbbm{1}_{\{i\in\text{Top-$k$}(\thetav)\}})_{i=1}^n$ is also known as the \emph{Top-$k$ mask} of $\thetav$.

\paragraph{Smoothed maximum operators.} Let $\Omega:\triangle^d\to\RR$ be a strictly convex regularization function. The \emph{smoothed maximum operator} $\maxo\!:\!\RR^d\!\to\!\RR$ is defined as $\maxo(\thetav)\!\triangleq\!\max_{\q\in\triangle^d}\left\{ \langle\thetav,\!\q\rangle\!-\!\Omega(\q) \right\}$. Typical choices for $\Omega$ include Shannon's negative entropy, which yields the \emph{log-sum-exp} operator and \emph{softmax} gradients, and quadratic (or Gini's negative entropy) regularization, which leads to the \emph{sparsemax} operator \citep{martins_softmax_2016}. This choice controls the smoothness and sparsity of the resulting operator: we discuss its consequences in our setting in \cref{sec:regularization_choice}.

\paragraph{Differentiating through combinatorial solvers.}
Integrating discrete combinatorial solvers into differentiable pipelines faces the challenge of uninformative gradients, as their output is piecewise constant. A line of research differentiates the Karush-Kuhn-Tucker conditions of linear or mixed integer programs \citep{mandi_interior_2020,ferber_mipaal_2020}. One can also treat the solver as a \emph{black-box} oracle. \citet{pogancic_differentiation_2020} propose a method to compute gradients via piecewise affine interpolation of the solver's output.
Alternatively, perturbation-based methods \citep{berthet_learning_2020, niepert_implicit_2021} smooth the operator by adding noise to inputs and estimate gradients via Monte-Carlo. \citet{cordonnier_differentiable_2021} apply these techniques specifically to Top-$k$ selection. While flexible, these approaches can be computationally expensive due to the need for multiple solver calls. 

\paragraph{Differentiable Top-$k$ and sorting.}
Existing approaches typically relax Problem~(\ref{pb:topk}) using regularized optimal transport \citep{cuturi_differentiable_2019,xie_differentiable_2020}, regularized linear programming \citep{amos_limited_2019,blondel_fast_2020,qian_multivector_2022,sander_fast_2023}, or 
smoothed sorting networks \citep{petersen_differentiable_2021,petersen_differentiable_2022}.
These methods, however, are deterministic and cannot be used as a stochastic layer. Closest to our work, \citet{ahmed_simple_2023} derive a differentiable Top-$k$ operator based on factorizing the entropy of distributions on $k$-subsets. By adopting a broader DP-based perspective, we strictly generalize that framework: we recover similar algorithms with Shannon entropy-based regularization, and obtain new sparse operators with Gini or Tsallis regularization. Crucially, unlike all aforementioned approaches, our framework naturally handles Knapsack constraints and non-uniform weights. Furthermore, it supports both deterministic and stochastic layers.

\paragraph{DP for structured prediction.}

In NLP, DP with the standard $(\max, +)$ semiring has been used for $k$-subset selection \citep{mcdonald_discriminative_2006,niculae_lpsparsemap_2020}.
Dynamic programs can be smoothed out using second-order expectation semirings \citep{li_first_2009}
or by replacing $\max$ operators in the Bellman recursion with smoothed $\maxo$ operators \citep{mensch_differentiable_2018}.

\section{Knapsack and Top-$k$ as Dynamic Programs}
\label{sec:dynamic}

\paragraph{Knapsack case.} Problem~(\ref{pb:knapsack}) famously exhibits optimal substructure, allowing for a DP solution \citep{martello_knapsack_1990,kellerer_knapsack_2004}. Indeed, let $\Vw[i, c]$ be the optimal value of the subproblem considering only the first $i\leq n$ items with a capacity $c\leq C$:

\vspace{-0.65cm}
\begin{align*}
    \Vw[i,c] \triangleq
    \max_{\y\in\{0,1\}^i} \;\sum_{l=1}^i \theta_l y_l\quad
\text{s.t.} \;\sum_{l=1}^i w_l y_l\leq c\,.
\end{align*}
To compute this, we can compare two possible decisions for item $i$. We can skip it and set $y_i\!=\!0$, yielding value $\Vw[i-1,c]$, or we can pick it (if $w_i \leq c$) and set $y_i=1$, yielding $\theta_i+\Vw[i-1, c-w_i]$:
\begin{align}
\label{eq:recursion_knapsack}
    \Vw[i, c] = 
        \Vw[i-1, c] \quad\text{if $w_{i}>c$,}
        \quad\max\bigl( \Vw[i\!-\!1, c],\theta_{i}+\Vw[i-1, c-w_{i}] \bigr)\quad \text{else.}
\end{align}
The optimal value of Problem~(\ref{pb:knapsack}) is $\Vw[n, C]=\maxcw(\thetav)$. This recursion runs in $\mathcal{O}(nC)$ time and space, and is initialized with $\Vw[0,\colon] = 0$. An illustration is given in \cref{fig:knapsack_dp_tables}.

\paragraph{Top-$k$ case.} For \eqref{pb:topk}, let $\Vk[i,j]$ be the optimal value when selecting \emph{exactly} $j$ items from the first $i$:

\vspace{-0.65cm}
\begin{align*}
    \Vk[i,j] \triangleq 
    \max_{\y\in\{0,1\}^i} \;\sum_{l=1}^i \theta_l y_l\quad
\text{s.t.} \;\sum_{l=1}^i y_l = j\,.
\end{align*}
The equality constraint implies that any subproblem $(i,j)$ with $j>i$ is infeasible. Consequently, if item $i$ is required to reach the target count (i.e., $i=j$), we are forced to pick it, as skipping would lead to an infeasible state. We enforce this behavior by initializing $\Vk[0,0]=0$ and $\Vk[0, 1\colon]=-\infty$, which naturally propagates to infeasible states. With $\Vk[\colon,0]=0$, for $j\geq 1$ the recursion writes:%

\vspace{-0.4cm}
\begin{equation*}
    \Vk[i,j] = 
    \max\left(\theta_i+ \Vk[i\!-\!1, j\!-\!1],\;\Vk[i\!-\!1, j]\right).
\end{equation*}
\noindent The optimal value of Problem~(\ref{pb:topk}) is therefore given by $\Vk[n, k] = \maxk(\thetav)$. This recursion runs in $\mathcal{O}(nk)$ time and space, and  we illustrate it on an example in \cref{fig:topk_dp_tables}.

\paragraph{Unified framework.}
Since in Top-$k$ we have that $w_i=1\leq c$ for $c\geq 1$ (and $w_i>0$ keeps $\Vk[\colon,0]=0$), \cref{eq:recursion_knapsack} serves as a unified recursion for Problems (\ref{pb:knapsack}) and (\ref{pb:topk}). The Top-$k$ case is then simply a specific instance of the Knapsack with $\w\!=\!\1$ and $C\!=\!k$, the only distinction being that the row $[0,1\colon]$ of the DP table is initialized with $-\infty$ rather than $0$ to enforce the equality constraint.

\paragraph{Backtracking.} These recursions yield the optimal values of Problems (\ref{pb:knapsack}) and (\ref{pb:topk}). To recover the maximizing mask $\ycw(\thetav)$ from $\Vw[n,C]$, one can \emph{backtrack} from $n$ down to $1$, setting $y_i$ to $0$ or $1$ based on which transition was active. Since $\ycw(\thetav)$ is a subgradient of $\maxcw(\thetav)$, this backtracking procedure is a special case of backpropagation \citep{blondel_elements_2025} (see \ref{sec:backtracking} for a derivation).

\begin{figure}[t!]
    \begin{minipage}{0.48\textwidth}
        \vspace{0pt}
        \begin{algorithm}[H]
        \caption{Computation of $\maxcwo(\thetav)$}
        \label{algo:value}
        \begin{algorithmic}
        \INPUT Item values $\thetav\!\in\!\RR^n$, item weights $\w\!\in\!\NN^n$, capacity $C\!\in\!\NN$, regularization function $\Omega$.
        \OUTPUT Smoothed value $\maxcwo(\thetav) \in \RR$.
        \STATE Initialize $\VVwo\!\in\!\RR^{(n+1)\times(C+1)}$ by setting $\Vwo[0,\!0]\!\gets\!0$, and either $\Vwo[0,1\colon]\!\gets\!0$ (Knapsack) or $\Vwo[0,1\colon]\!\gets\!-\infty$ (Top-$k$).
        \STATE Initialize $\QQwo\gets\0\in\RR^{(n+1)\times(C+1)}$.
        \FOR{$i=1$ {\bfseries to} $n$}
            \FOR{$c=0$ {\bfseries to} $C$ \COMMENT{parallel loop}}
                \STATE
                \vspace{-0.55cm}
                \begin{align*}
                    \!\!\!\Vwo[i,c]\!\gets\!\begin{cases}
                         \Vwo[i-1,c] &\text{if {$w_i\!>\!c$,}}\\
                         \maxo\bigl(\theta_i\!+\!\Vwo[i\!-\!1,c\!-\!w_i],\!\!\!\!\!\!\!\!\!\!\!\!\!\!\!\!\!\!\!\!\!\!\!\!\!\!\!\!\\
                    \quad\Vwo[i-1,c]\bigr) &\text{else.}
                    \end{cases}
                \end{align*}
                \STATE \!\!\!$\Qwo[i,c]\gets\frac{\partial\Vwo[i,c]}{\partial\theta_i}$
            \ENDFOR
        \ENDFOR
        \STATE \textbf{return} $\maxcwo(\thetav)\!=\!\Vwo[n,\!C]$ and $(\VVwo\!\!,\!\QQwo)$ (intermediate results for \cref{algo:layer,algo:sample,algo:vjp}).
        \end{algorithmic}
        \end{algorithm}
    \end{minipage}
    \begin{minipage}[c]{0.005\textwidth}
    \vspace*{11pt}
    \rule{0.4pt}{8.2cm}
    \end{minipage}
    \begin{minipage}{0.48\textwidth}
        \vspace{0pt} %
        \begin{algorithm}[H]
        \caption{Computation of $\ycwo(\thetav)$}
        \label{algo:layer}
        \begin{algorithmic}
        \INPUT Item values $\thetav\!\in\!\RR^n$, item weights $\w\!\in\!\NN^n$, capacity $C\!\in\!\NN$, regularization function $\Omega$, outputs $\VVwo,\QQwo$ of \cref{algo:value}.
        \OUTPUT Differentiable operator $\ycwo(\thetav)$.
        \STATE Initialize $\EEwo\!\gets\0\in\RR^{(n+1)\times(C+1)}$\! by setting $\Ewo[n,C]\gets1$.
        \FOR{$i=n-1$ {\bfseries to} $1$}
            \FOR{$c=0$ {\bfseries to} $C$ \COMMENT{parallel loop}}
                \STATE
                \vspace{-0.5cm}
                \begin{align*}
                    \Ewo[i,c]\gets&\Ewo[i+1,c]\\
                    \times&(1-\Qwo[i+1,c])\\
                    + &\Ewo[i+1,c+w_{i+1}]
                    \\\times&\Qwo[i+1,c+w_{i+1}]
                \end{align*}
            \vspace{-0.5cm}
            \ENDFOR
        \ENDFOR
        \STATE $\ycwo(\thetav) \gets(\EEwo[1\colon,\colon]\circ\QQwo[1\colon,\colon])\cdot  \1$
        \STATE \textbf{return} $\ycwo(\thetav)$ and $(\VVwo,\QQwo,\EEwo)$ (intermediate results for \cref{algo:vjp}).
        \end{algorithmic}
        \end{algorithm}
    \end{minipage}
\end{figure}

\section{Differentiable Knapsack and Top-$k$ Operators}
\label{sec:differentiable_operators}

\subsection{Algorithmic smoothing}
\label{sec:algorithmic_smoothing}

We now algorithmically smooth the dynamic programs defined in \cref{sec:dynamic}, by regularizing the $\max$ operator in \eqref{eq:recursion_knapsack} into $\maxo$ using a strictly convex function $\Omega:\triangle^2\to\RR$:%
\begin{align}
\label{eq:smoothed_recursion}
    \Vwo[i,c] \triangleq \Vwo[i-1, c] \quad \text{if $w_{i}>c$,}\quad 
        \maxo\bigl(\theta_{i}+\Vwo[i-1, c-w_{i}],\Vwo[i-1, c]\bigr) \quad \text{else.}
\end{align}
In \cref{sec:dag_ddp}, we relate this approach to the framework of \citet{mensch_differentiable_2018}. We now define our differentiable Knapsack and Top-$k$ operators using the smoothed recursion in \cref{eq:smoothed_recursion}.

\begin{definition}[Smoothed DP values and operators.]
Let $\thetav\in\RR^n$. We define the following relaxations of the Knapsack and Top-$k$ problem values and the corresponding operators:
\begin{align*}
    \maxcwo(\thetav)&\triangleq \Vwo[n,C],\quad &&\maxko(\thetav)\triangleq \Vko[n,k],\\
    \ycwo(\thetav)&\triangleq \nabla\maxcwo(\thetav)\in\conv(\Ycw),\quad
    &&\yko(\thetav)\triangleq \nabla\maxko(\thetav)\in\conv(\Yk).
\end{align*}
\vspace{-0.4cm}
\end{definition}

Importantly, the smoothed values are convex in $\thetav$, allowing us to define convex supervised learning losses in \cref{sec:output_layer}. We prove this and other properties of the relaxed operators in \cref{sec:properties}.

\paragraph{Parallel implementation.}
Observing that values $\Vwo[i, \cdot]$ depend only on row $i\!-\!1$, we exploit \emph{wavefront parallelism} \citep{muraoka_parallelism_1971} by vectorizing over capacities. This reduces the effective time complexity from $\cO(nC)$ to $\cO(n)$ parallel steps while incurring an $\cO(nC)$ memory footprint (see \hyperref[par:limitations]{Limitations and scalability} for large-$C$ mitigations). We provide pseudo-code in \cref{algo:value,algo:layer}, with detailed derivations given in \cref{sec:layer_derivation}. To achieve high-performance execution without custom CUDA kernels, we provide a just-in-time compiled implementation in \texttt{Numba} \citep{lam_numba_2015}, efficiently parallelized over batch and capacity dimensions.

\subsection{Regularization choice}
\label{sec:regularization_choice}

We now show how regularization impacts the \emph{equivariance} and \emph{sparsity} of our operators. We discuss three choices for $\Omega$, based on  the Shannon, Gini, and $1.5$-Tsallis entropies \citep{blondel_learning_2019}:
\begin{align*}
    H^{s}(\q) \triangleq -\left(q_1\log q_1 + q_2\log q_2\right),\quad H^{g}(\q)\triangleq  \frac{1}{2}\left(1 -q_1^2-q_2^2\right),\quad H^{T}_{1.5}(\q)\triangleq  \frac{4}{3}\left(1-q_1^{\frac{3}{2}} -q_2^{\frac{3}{2}}\right).
\end{align*}

We use a regularization strength parameter $\gamma>0$, and consider $\Omega\in\{-\gamma H^{s},-\gamma H^{g},-\gamma H_{1.5}^{T}\}$.
Shannon's negative entropy $\Omega=-\gamma H^{s}$ yields closed forms for $\maxcwo$ and $\ycwo$ (see \cref{sec:closed_form}). Within the $\alpha$-Tsallis family \citep{tsallis_possible_1988}, the $1.5$-Tsallis entropy can be seen as a trade-off between the Shannon and Gini entropies (recovered as $\alpha\to 1$ and $\alpha=2$ respectively): it allows us to combine both smoothness and exact sparsity (see \cref{fig:topk_knapsack_3D}), while retaining efficient, closed-form expressions for the smoothed maximum and its derivatives (see \cref{sec:explicit_derivations} and \cref{tab:instantiations}).

In \cref{prop:sparsity} and \cref{sec:properties}, we show that the bias induced by regularization (i.e., the distance between hard operators and their smoothed counterparts) decays exponentially with $\gamma$ under Shannon regularization, and reaches exactly zero under a finite regularization threshold under Gini and Tsallis.

\paragraph{Equivariance.} We now show that $\Omega=-\gamma H^{s}$ is the \emph{only} separable regularization choice ensuring the permutation-equivariance of the smoothed Knapsack and Top-$k$ operators.

\begin{proposition}[Characterization of equivariance]
\label{prop:equivariance}
    Let $\Omega:\triangle^2\to\RR$ be a convex, \emph{separable} function such that $\Omega(\e_1)=\Omega(\e_2)=0$, and let $S_n$ be the group of permutations. We have:
    \begin{align*}
        &\sigma \circ \ycwo = \y^C_{\sigma(\w),\Omega}\circ\sigma \quad \forall (\sigma,\w) \in S_n\times\NN^n\quad \textbf{if and only if} \quad  \Omega=-\gamma H^{s} \quad \text{for some}\; \gamma\geq 0.
    \end{align*}
\end{proposition}

Note that in the Top-$k$ case we have $\sigma(\w)=\w=\1$. The proof of \cref{prop:equivariance}, given in \cref{proof:equivariance}, relies on two ingredients: first, the equivalence between the permutation-invariance of a function and the permutation-equivariance of its gradient, and second, the fact that only $\Omega=-\gamma H^{s}$ yields an \emph{associative} smoothed maximum operator $\maxo$.

\paragraph{Sparsity.} We now characterize the regularizers that enable the relaxed operators $\ycwo$ to map to vertices $\y\in\Ycw$ of the moment polytope $\conv(\Ycw)$, i.e., to produce sparse outputs. 

\begin{proposition}[Characterization of sparsity]
\label{prop:sparsity}
    Let $\Omega(\q) \triangleq \sum_{i=1}^2 \omega(q_i)$ be a separable, strictly convex regularizer with $\omega$ differentiable on $(0,1)$.
    The following statements are equivalent:
    \begin{enumerate}
        \item $\lim_{t\to0^+}\omega'(t) > -\infty \quad \text{and} \quad \lim_{t\to1^-}\omega'(t) < +\infty$.
        \item $\ycwo$ can output hard discrete selections, i.e., $\Ycw \subseteq \range(\ycwo)$.
        \item \emph{Exact recovery:} for any $\thetav\in\RR^n$ yielding a unique hard maximizer $\ycw(\thetav)$, there exists a regularization threshold $\gamma^* > 0$ such that $\y^C_{\w,\gamma\Omega}(\thetav) = \ycw(\thetav)$ for all $\gamma \in (0, \gamma^*]$.
    \end{enumerate}
\end{proposition}
The proof is given in \cref{proof:sparsity}. Both Gini ($\omega'(q)=\gamma q$) and $1.5$-Tsallis ($\omega'(q)=2\gamma\sqrt{q}$) satisfy condition (1), whereas even above $\gamma^*$, bounded endpoint derivatives allow individual DP transitions to saturate to $\{0,1\}$, placing $\ycwo(\thetav)$ on boundary faces of $\conv(\Ycw)$ (inducing both hard $0/1$ coordinates and soft ones). Conversely, in the Shannon-regularized setting $\Omega=-\gamma H^{s}$, and more generally for any Legendre-type regularizer \citep{rockafellar_convex_1970}, whose gradients explode at the boundary of $\triangle^2$, the outputs of $\ycwo$ are forced to lie in the relative interior $\ri(\conv(\Ycw))\subseteq (0,1)^n$, and can therefore never be sparse.

We illustrate our operators in \cref{fig:topk_knapsack_3D}, and numerically demonstrate \cref{prop:equivariance,prop:sparsity} in \cref{fig:equivariance_and_sparsity}.

\section{Layer Integration}
\label{sec:integration}

In \cref{sec:continuous_forward,sec:stochastic_forward,sec:output_layer}, we describe three principled scenarios for integration of the proposed operators into a differentiable programming framework: as a deterministic hidden layer, as a stochastic hidden layer, and as an output layer. 

We denote by $\thetav\in\RR^n$ the layer's input, which is typically obtained as $\thetav=f_{W_1}(\x)$ where $f_{W_1}$ is an upstream model with learnable parameters $W_1$. Similarly, we denote the downstream computation graph by $f_{W_2}$.

\subsection{Deterministic forward passes and VJPs}
\label{sec:continuous_forward}

In the hidden layer case, the most direct approach is to use the relaxed operator $\ycwo(\thetav)$ as a deterministic, continuous layer $\thetav\mapsto\ycwo(\thetav)\in\conv(\Ycw)$ during training, directly feeding its output to the downstream function $f_{W_2}$.

To learn the upstream parameters $W_1$ via backpropagation, we must compute the vector-Jacobian product (VJP) of the layer.
Since $\ycwo=\nabla\maxcwo$, the Jacobian $\nabla\ycwo(\thetav)$ is the Hessian $\nabla^2\maxcwo(\thetav)$, which is symmetric. Thus, for any cotangent vector $\z\in\RR^n$, the VJP $\z^\top(\nabla_\thetav\ycwo(\thetav))$ is the transpose of the JVP $(\nabla_\thetav\ycwo(\thetav))\z$.

We compute it using a \emph{reverse-over-forward} approach, by backpropagating through the computation of the directional derivative $\langle \nabla_\thetav\maxcwo(\thetav),\z\rangle$, which we derive in \cref{sec:directional_derivation}. The special structure of our DAG, which has repeated edges and constant ones, leads to important changes compared to the general case of \citet{mensch_differentiable_2018} (see \cref{sec:dag_ddp}). The pseudo-code is given in \cref{algo:vjp}, with derivations in \cref{sec:vjp_derivation}.

In \cref{sec:explicit_derivations}, we derive explicit formulas for the quantities needed to instantiate \cref{algo:vjp}  with Shannon, Gini, and $1.5$-Tsallis regularization, which we summarize in \cref{tab:instantiations}.

\subsection{Stochastic forward passes and surrogate gradients}
\label{sec:stochastic_forward}

The method described in \cref{sec:continuous_forward} is susceptible to a train-test mismatch. Although the network is trained on continuous vectors $\ycwo(\thetav)\in\conv(\Ycw)$, inference often requires hard item selections $\ycw(\thetav) \in \Ycw$. This is particularly true for semantically discrete tasks where passing a soft, averaged vector $\ycwo(\thetav)$ to $f_{W_2}$ would be nonsensical. We propose to bridge this gap via sampling.

\begin{proposition}[Underlying distribution]
\label{prop:distribution}
    Let $\QQwo$ be the output of \cref{algo:value}. Let $\picwo$ be the distribution on $\{0,1\}^n$ defined autoregressively by $\picwo(Y_i=1|Y_{i+1},\dots, Y_n)\triangleq \Qwo\left[i,\,C_i\right]$ for all $i\in\{n,\dots,1\}$, where $C_i\triangleq C-\sum_{j=i+1}^n Y_j w_j$ is the remaining capacity for items $[i]$. 
    
    The support of $\picwo$ is a subset of $\Ycw\,$, and its expectation matches the relaxed operator: $\ycwo(\thetav) = \EE_{\picwo}[Y]\in\conv(\Ycw)$.
    Moreover, in the Shannon-regularized case $\Omega=-\gamma H^{s}$, we recover the Gibbs distribution $\picwo(\y) \propto \exp(\langle\thetav,\y\rangle/\gamma)$.
\end{proposition}

\cref{prop:distribution}, proved in \cref{proof:distribution}, shows the existence of a distribution $\picwo$ underlying the proposed operators, and yields a tractable $\cO(n)$ ancestral sampling algorithm to sample from it. We give the corresponding pseudo-code in \cref{algo:sample}.

\begin{algorithm}
\caption{Sampling from $\picwo$}
\label{algo:sample}
\begin{algorithmic}
\INPUT Values $\thetav \in \RR^n$, weights $\w\in\NN^n$, capacity $C \in \NN$, regularizer $\Omega$, output $\QQwo$ of \cref{algo:value}.
\OUTPUT Sample $\y\sim\picwo$.
\STATE Initialize $\y\in\RR^n$, $c\gets C$.
\FOR{$i=n$ {\bfseries to} $1$}
    \STATE $y_i\sim \mathrm{Bernoulli}\left( \Qwo[i,c] \right)$
    \STATE $c\gets c - y_iw_i$
\ENDFOR
\STATE \textbf{return} $\y$
\end{algorithmic}
\end{algorithm}

The structure of $\picwo$ enables to efficiently compute the probability of any selection $\y\in\Ycw$ as $\picwo(\y) = \prod_{i=1}^n\left( y_i\Qwo[i,c_i] + (1-y_i)(1-\Qwo[i,c_i]) \right)$,
where $c_i\triangleq C - \sum_{j=i+1}^ny_j w_j$.%
This is very useful as it enables assessing the probability of a given Top-$k$ or Knapsack assignment.
 
\paragraph{Surrogate gradients.} To enforce hard item selections during training, we replace the relaxed output $\ycwo(\thetav)\in\conv(\Ycw)$ with a discrete sample $\y \in \Ycw$ from $\picwo$ in the forward pass.
To bypass the non-differentiability of the sampling operation, we adopt the following surrogate gradient approach:
\begin{enumerate}
    \item During the forward pass, we treat the layer as a stochastic node by sampling $\y\sim\picwo$ using \cref{algo:sample}. We can then compute any loss $\cL(\x, f_{W_2}(\y))$.
    \item In the backward pass, we treat the layer as the deterministic node $\thetav\!\mapsto\!\ycwo(\thetav)\!=\!\EE_{\picwo}[Y]$, with VJP computed by \cref{algo:vjp}. Thus, we use
    \begin{align*}
        \nabla_\thetav \cL(\x, f_{W_2}(\y))\triangleq \left(\nabla_\thetav\ycwo(\thetav)\right)\cdot\nabla_\y \cL(\x, f_{W_2}(\y)),
    \end{align*}
    and backpropagate the cotangent $\nabla_\thetav \cL(\x,f_{W_2}(\y))$ to upstream learnable parameters $W_1$.
\end{enumerate}
A similar approach is also proposed in \citet{ahmed_simple_2023} (although limited to the Shannon-regularized Top-$k$ setting). In \cref{sec:surrogate_properties}, we show that this surrogate gradient is an \emph{unbiased first-order estimator} of the expectation of the loss, following the analysis in \citet{liu_bridging_2023}. We also show its superiority to alternatives using hard item selections during forward passes in \cref{fig:mock_gradient}.
\newpage

\subsection{Output layers and Fenchel-Young losses}
\label{sec:output_layer}

We now give principled losses for supervised learning with the proposed operators as output layers.

\paragraph{Moment polytope regularization.} For any $\Omega:\triangle^2\to\RR$ used to smooth the DP recursion \eqref{eq:recursion_knapsack} into \cref{eq:smoothed_recursion}, define the regularization function $\Omegacw$ as the Fenchel conjugate of $\maxcwo$ with domain $\conv(\Ycw)$. The corresponding \emph{Fenchel-Young loss} \citep{blondel_learning_2020} is given by:
\begin{align*}
    L_{\Omegacw}\!: \RR^n\!\times\!\conv(\Ycw)\longrightarrow\RR,\quad \text{with} \quad 
    L_{\Omegacw}(\thetav\,;\y)\triangleq \maxcwo(\thetav)+\Omegacw(\y)\!-\!\langle\thetav,\y\rangle.
\end{align*}
This loss function has several desirable properties \citep[Proposition 2]{blondel_learning_2020}:
\begin{itemize}
    \item $\thetav\mapsto L_{\Omegacw}(\thetav\,;\y)$ is convex for all $\y\in\Ycw$,
    \item $\thetav\mapsto L_{\Omegacw}(\thetav\,;\y)$ is differentiable with gradient given by $\nabla_\thetav L_{\Omegacw}(\thetav\,;\y)=\ycwo(\thetav)-\y$,
    \item $L_{\Omegacw}(\thetav\,;\y)\geq0$ for all $(\thetav,\y)\in\RR^n\times\conv(\Ycw)$,
    \item $L_{\Omegacw}(\thetav\,;\y)=0$ if and only if $\thetav$ is such that $\ycwo(\thetav)=\y$.
\end{itemize}
Thus, computing gradients reduces to computing a forward pass of the relaxed layer $\ycwo(\thetav)$ using \cref{algo:layer}, and backpropagating the difference between its output and the ground-truth target $\y$.

\begin{figure}[t!]
    \centering
    \begin{subfigure}[b]{0.49\textwidth}
        \centering
        \includegraphics[width=\textwidth]{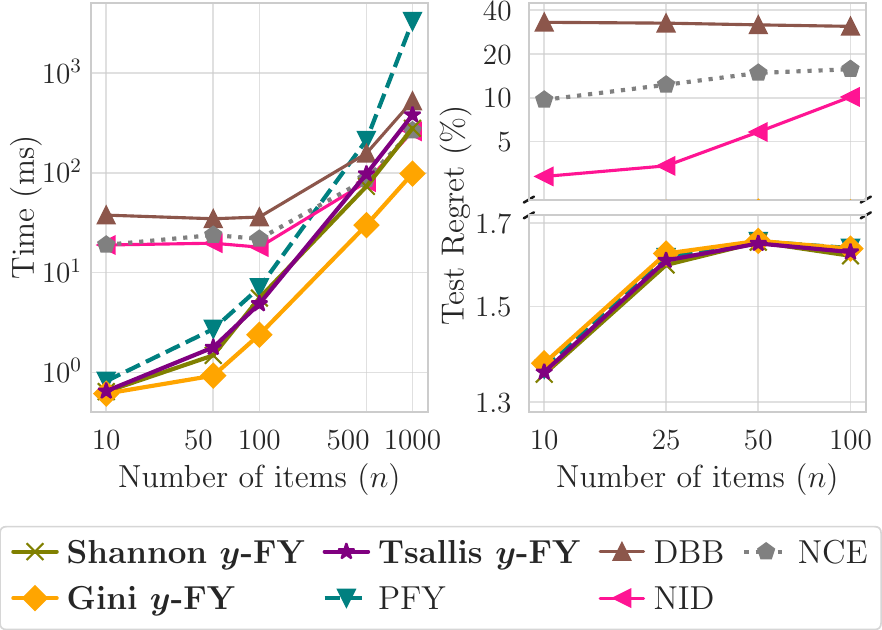}
        \caption{\textbf{Scaling and performance of $\ycwo$ as an output layer.}
    Lower computational time and test relative regret are better.}
        \label{fig:pyepo_scale}
    \end{subfigure}
    \hfill
    \begin{subfigure}[b]{0.49\textwidth}
        \centering
        \includegraphics[width=\textwidth]{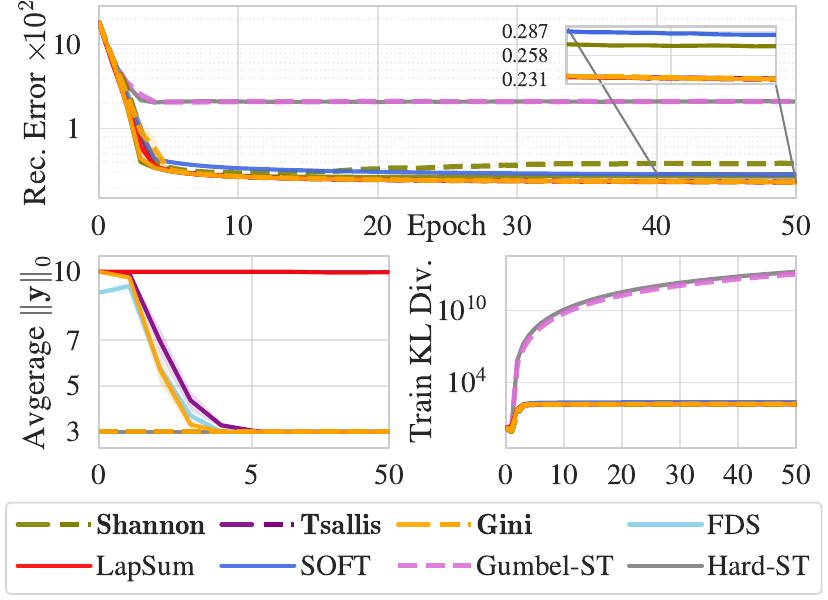}
        \caption{\textbf{DVAE training dynamics:} test MSE, sparsity of latent representations $\y$ and KL divergence during training (dashed lines indicate stochastic forward passes).}
        \label{fig:curves_dvae}
    \end{subfigure}
    \caption{Evaluation of our DP-based operators as output layers on Knapsack prediction (a) and as stochastic and deterministic hidden layers in discrete VAEs (b).}
    \label{fig:main}
\end{figure}

\section{Experiments}
\label{sec:experiments}

\subsection{Learning to predict Knapsack solutions}
\label{sec:experiments_output}

We compare the losses proposed in \cref{sec:output_layer} against four established baselines: \texttt{PFY} \citep{berthet_learning_2020}, \texttt{DBB} \citep{pogancic_differentiation_2020}, \texttt{NCE} \citep{mulamba_contrastive_2021}, and \texttt{NID} \citep{sahoo_backpropagation_2022} (all are detailed in \cref{sec:pyepo_appendix}) on the \texttt{PyEPO} benchmark \citep{tang_pyepo_2024}. Given a dataset $(\x^{(i)},\thetav^{(i)},\y^{(i)})_{i=1}^N$, the goal is to predict item values $\hat{\thetav}^{(i)}$ from features $\x^{(i)}$ such that the subsequent Knapsack solution $\ycw(\hat{\thetav}^{(i)})$ matches the ground-truth selection $\y^{(i)}$.

We measure performance using relative regret on a test set,  defined as $\frac{\langle \thetav^{(i)}, \y^{(i)}\rangle - \langle \thetav^{(i)}, \ycw(\hat{\thetav}^{(i)})\rangle}{|\langle \thetav^{(i)}, \y^{(i)}\rangle |}$.

\paragraph{Results.}
We parameterize $\hat{\thetav}=f_{W_1}(\x)$ as a standard feed-forward neural network, and vary the number of items $n$. We measure the average computation time of each evaluated loss function (forward and backward pass up to $\thetav$), and the best test relative regret.
The results are gathered in \cref{fig:pyepo_scale}. Our DP-based Fenchel-Young losses consistently outperform the baselines in terms of regret while remaining computationally efficient. Full experimental details are given in \cref{sec:pyepo_appendix}.

\subsection{Fenchel-Young discrete VAE}
\label{sec:experiments_dvae}

We now evaluate our relaxed Top-$k$ operators within a discrete variational auto-encoder (DVAE, \citet{rolfe_discrete_2016}) framework. The goal is to learn representations that disentangle continuous style attributes from discrete categorical identity. The encoder $E_{W_1}$ maps an input $\x$ to latent style parameters $\z_1, \dots, \z_n \in \RR^{d_z}$ and selection logits $\thetav \in \RR^n$.
The discrete latent selection is obtained via either $\y = \yko(\thetav)$ or $\y\sim\piko$ (following \cref{sec:continuous_forward,sec:stochastic_forward}). The reconstruction $\hat{\x}$ is then generated by a shared decoder $D_{W_2}$, which processes learnable class embeddings $\e_i$ conditioned on style variables $\z_i$, aggregated by the selection weights $\y$, i.e., $\hat{\x} \triangleq \frac{1}{k} \sum_{i=1}^n y_i \cdot D_{W_2}([\e_i, \z_i])$.

To train this model, we minimize the sum of a MSE reconstruction loss on $\hat{\x}$, a KL regularization term for the latent style variables $\z$, and a Fenchel-Young regularization term $L_{\Omegak}(\0\,; \yko(\thetav))$ for the latent selection logits $\thetav$ using the loss proposed in \cref{sec:output_layer}. The gradient of this regularization term is given by $\nabla_\thetav L_{\Omegak}(\0\,;\yko(\thetav))=(\nabla_\thetav\yko(\thetav))\cdot\thetav$, i.e., it is the JVP of $\yko$ in direction $\thetav$, efficiently computed via \cref{algo:vjp} (see \cref{sec:dvae_appendix} for the derivation).

The term $L_{\Omegak}$ equals the standard KL divergence against a uniform prior in the $\Omega=-H^{s}$ setting, recovering the framework of SIMPLE \citep{ahmed_simple_2023}, while enabling more general regularizers that induce sparse latent representations, such as Gini and Tsallis entropies. This formulation also extends the Fenchel-Young variational inference framework of \citet{sklaviadis_fenchelyoung_2025}, from continuous VAEs to DVAEs with latent distributions on $k$-subsets. \cref{tab:dvae_fy_ablation} presents an ablation of this regularization term, illustrating the classic rate-distortion trade-off typical of VAEs.

All derivations and experimental details are given in \cref{sec:dvae_appendix}.

\paragraph{Experimental setup.}
We use a stacked version of the MNIST dataset \citep{lecun_mnist_2005}, where inputs are pixel-wise averages of $k\!=\!3$ distinct digit images. The model must recover the identities of the constituent digits via the latent $k$-subset $\y$. We compare our deterministic and stochastic Top-$k$ operators against those of \citet{sander_fast_2023} (FDS), \citet{struski_lapsum_2025} (LapSum), \citet{xie_differentiable_2020} (SOFT), and to hard and Gumbel-perturbed Top-$k$ baselines \citep{_gumbelmax_,kool_stochastic_2019} using straight-through estimators (ST, \citet{bengio_estimating_2013}) for the backward pass.

\paragraph{Results.}
\cref{fig:curves_dvae} illustrates the training dynamics, highlighting three results. First, our DP-based operators match or outperform baselines in reconstruction quality; in particular, our variants equipped with sparse regularizers (Gini and Tsallis) surpass the Shannon-based SIMPLE framework \citep{ahmed_simple_2023} across both deterministic and stochastic forward passes. Second, unlike the SIMPLE, LapSum, or SOFT Top-$k$ operators, Gini and Tsallis regularization induce sparse latent representations, quickly converging to hard $k$-subsets to explain the data. Third, our methods exhibit superior stability compared to ST-based baselines, which lead to a diverging KL regularization term.

\begin{figure*}[t!]
    \centering
    \includegraphics[width=0.99\linewidth]{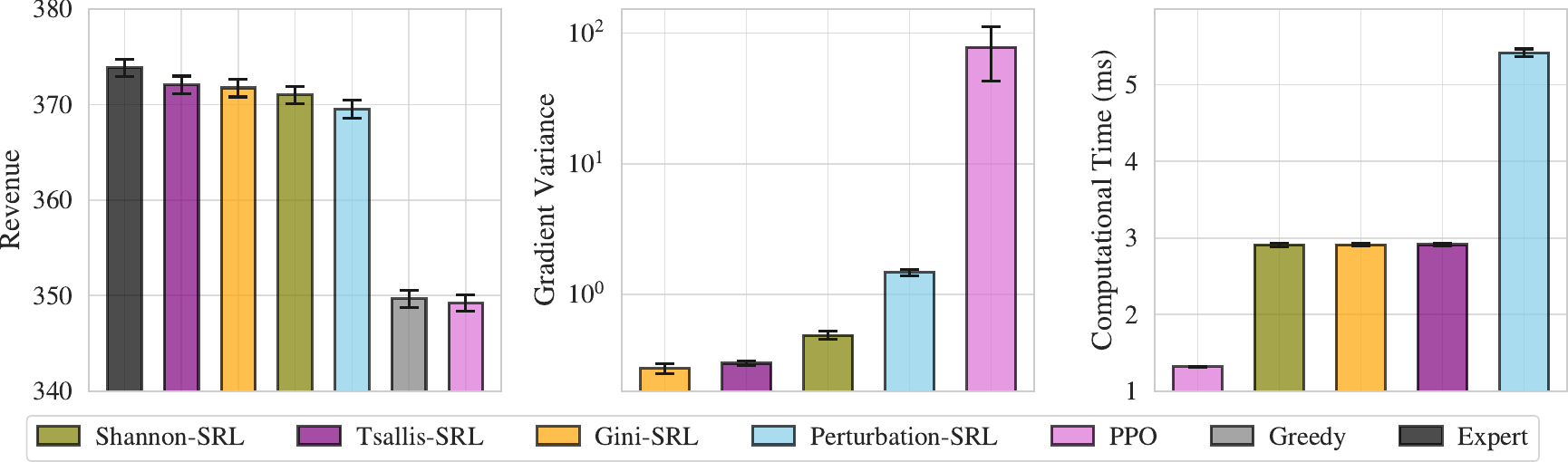}
    \caption{\textbf{Constrained dynamic assortment.} \textbf{Left:} Expected revenue on $10^4$ test episodes. Our DP framework outperforms realizable baselines and approaches the full-information expert oracle. \textbf{Middle:} Expected gradient variance. Sparse regularizers (Gini, Tsallis) significantly reduce variance by inducing a sparse support for $\picwo$. \textbf{Right:} Gradient estimation wall-clock time. By avoiding repeated solver calls and reusing intermediate outputs from \cref{algo:value} for sampling and differentiation via \cref{algo:layer,algo:sample}, our approach achieves notable speedups over perturbation-based SRL.}
    \label{fig:srl}
\end{figure*}

\subsection{Dynamic assortment with Knapsack constraints}
\label{sec:experiments_srl}

We evaluate our differentiable Knapsack operators on a dynamic assortment problem \citep{talluri_theory_2004,chen_dynamic_2020}. The agent manages a store of $n=20$ items and selects subsets to display over $80$ time steps, subject to capacity and inventory constraints. The environment features endogenous uncertainty driven by a hidden customer choice model, parameterized by unknown preferences $\phiv$. Rewards correspond to sold item prices, which deplete inventory and update features. %

\paragraph{Method.}
We employ the structured reinforcement learning (SRL) framework of \citet{hoppe_structured_2025}, where the actor maps states to item logits $\thetav$. While the original method relies on the perturbation framework of \citet{berthet_learning_2020} for exploration and gradient estimation, our DP-based formulation yields \emph{exact} Fenchel-Young gradients $\nabla_\thetav L_{\Omegacw}(\thetav; \hat{\y}) \!=\! \ycwo(\thetav) \!-\! \hat{\y}$. Furthermore, we substitute perturbed optimization with ancestral sampling using \cref{algo:sample} to generate targets $\hat{\y}\sim\picwo$.

\paragraph{Baselines.}
We compare our method against standard perturbation-based SRL and proximal policy optimization (PPO, \citet{schulman_proximal_2017}). We also include a \emph{Greedy} policy, which maximizes the cumulative price of displayed objects, and an \emph{Expert} oracle. This expert ``cheats'' by accessing the hidden customer model $\phiv$ to compute the exact expected immediate revenue of every feasible assortment $\y\in\Ycw$, and chooses the maximizing one. While this policy is not globally optimal (it ignores inventory and customer choice dynamics), it serves as a strong performance ceiling. Note that it is not computationally scalable, as it requires enumerating all feasible actions at every step, and $|\Ycw|$ grows exponentially with $n$.

\paragraph{Results.}
\cref{fig:srl} shows that our proposed operators consistently outperform realizable baselines in expected revenue while reducing gradient variance and runtime over perturbation-based SRL. Although this pipeline blends stochastic sampling and FY gradients, their respective benefits remain clear. First, our sparse regularizers uniquely induce a sparse distribution support for $\picwo$ for the sampling phase. Second, our exact and deterministic computation of the FY gradient negates the need for Monte-Carlo sampling, further reducing variance. Third, our regularized DP has the same complexity as a single unregularized call needed by Monte-Carlo sampling for perturbation-based SRL (and reuses \cref{algo:value} outputs across \cref{algo:layer,algo:sample}), which in turn reduces the required computational time. Full experimental details are in \cref{sec:srl_appendix}.

\section*{Conclusion and Limitations}

We proposed a unified DP framework for differentiable Knapsack and Top-$k$ operators, showing that the choice of regularization dictates key properties like permutation equivariance and sparsity. Our operators seamlessly enforce combinatorial constraints in differentiable pipelines as deterministic or stochastic hidden layers, or as output layers with Fenchel-Young losses for supervised learning.

\paragraph{Limitations and scalability.}
\label{par:limitations} 
While the DP formulation has an $\mathcal{O}(nC)$ complexity, wavefront parallelization reduces execution to $\mathcal{O}(n)$ parallel steps, making time complexity competitive on modern hardware. The $\mathcal{O}(nC)$ memory footprint poses no issue for our targeted applications, but exceptionally large $C$ could create bottlenecks. To mitigate this, one can trade compute for memory by storing only a fraction of the DP table and recomputing the rest in the backward. Alternatively, one can reduce the state space by dividing item weights and the capacity by a common factor (this is exact if using their greatest common divisor, or trades slight precision for memory if rounding).

\section*{Acknowledgments and Disclosures}
This work was fully supported by Google DeepMind, where Germain Vivier-Ardisson is a full-time employee and PhD candidate. The authors have no competing interests or financial relationships outside the submitted work to disclose.

\bibliography{main}
\bibliographystyle{abbrvnat}

\appendix
\onecolumn

\paragraph{Notation.} We denote $[n] \triangleq \{1, \dots, n\}$. Vectors and matrices use boldface letters (e.g., $\v, \1, \0,\M$). The probability simplex in $\RR^d$ is $\triangle^d\triangleq\!\{\q\in\RR^d_{\geq0}\mid\sum_{i}q_i\!=\!1\}$. We use $\langle \cdot, \!\cdot \rangle$ and $\circ$ for inner and element-wise products. The convex hull and relative interior of a set $\cS$ are $\conv(\cS)$ and $\ri(\cS)$. For a function $f$, $\range(f)$ and $\nabla f$ %
are its range and gradient. %

\section{Proofs}
\label{sec:proofs}

\subsection{Closed forms in the Shannon-entropy regularized case}
\label{sec:closed_form}
For $\Omega=-\gamma H^{s}$, we have:
\begin{align*}
    \maxcwo(\thetav) &= \gamma\cdot \log\!\!\sum_{\y\in\Ycw}\exp\left(\langle \thetav,\y\rangle/\gamma\right)\,,\\
    \maxko(\thetav) &= \gamma\cdot \log\!\!\sum_{\y\in\Yk}\exp\left(\langle \thetav,\y\rangle/\gamma\right)\,,\\
    \ycwo(\thetav) &= \sum_{\y\in\Ycw}\frac{\exp\left(\langle \thetav,\y\rangle/\gamma\right)}{\sum_{\y'\in\Ycw}\exp\left(\langle \thetav,\y'\rangle/\gamma\right)}\cdot \y\,,\\
    \yko(\thetav) &= \sum_{\y\in\Yk}\frac{\exp\left(\langle \thetav,\y\rangle/\gamma\right)}{\sum_{\y'\in\Yk}\exp\left(\langle \thetav,\y'\rangle/\gamma\right)}\cdot \y\,.
\end{align*}
\begin{proof}
For any integer $d\geq2$, define $\Omega_d:\triangle^{d}\to\RR$ as $\Omega_d(\q)\triangleq \gamma \sum_{i=1}^dq_i\log q_i$. \citet[Proposition 2]{mensch_differentiable_2018} gives that this definition of the family $(\Omega_d)_{d\geq 2}$ is such that:
\begin{align*}
    \mathsf{max}^C_{\w,\Omega_2}(\thetav) = \mathsf{max}_{\Omega_{|\Ycw|}}(\s_{\thetav}),
\end{align*}
where we define the score vector $\s_{\thetav}\triangleq \left(\langle\thetav,\y\rangle\right)_{\y\in\Ycw}$. To make the bridges between notations clearer, \citet{mensch_differentiable_2018} use a generic, dimension-agnostic definition of the regularizer $\Omega$, and $\mathsf{max}^C_{\w,\Omega_2}(\thetav)$ and $\max_{\Omega_{|\Ycw|}}(\s_{\thetav})$ are denoted by $\mathrm{DP}_\Omega(\thetav)$ and $\mathrm{LP}_\Omega(\thetav)$, respectively.

Moreover, since we have the closed form $\mathsf{max}_{\Omega_d}(\q)=\gamma \log\sum_{i=1}^d \exp(q_i/\gamma)$ for any $d\geq2$, we get:

\begin{align*}
    \y^C_{\w,\Omega_2}(\thetav) &= \nabla_\thetav\mathsf{max}^C_{\w,\Omega_2}(\thetav)\\
    &= \nabla_\thetav \mathsf{max}_{\Omega_{|\Ycw|}}(\s_{\thetav})\\
    &= \nabla_\thetav\left( \gamma\log\!\!\!\sum_{\y\in\Ycw}\!\!\exp\left( \langle\thetav,\y\rangle/\gamma \right) \right)\\
    &= \sum_{\y\in\Ycw}\frac{\exp\left(\langle \thetav,\y\rangle/\gamma\right)}{\sum_{\y'\in\Ycw}\exp\left(\langle \thetav,\y'\rangle/\gamma\right)}\cdot \y\,.
\end{align*}

While we wrote the proof only in Knapsack notation, since the differences in the padding values initializing the table $\VVwo$ (which is the only distinction between the Knapsack and the Top-$k$ cases) do not impact it, the exact same arguments naturally hold for $\yko$.
\end{proof}

\subsection{Proof of \cref{prop:equivariance} (\nameref{prop:equivariance})}
\label{proof:equivariance}

\begin{proof}
$\bullet$  ($\implies$)

First, we show the equivariance of $\ycwo$ with respect to permutations of $\thetav$ and $\w$ for $\Omega=-\gamma H^{s}$. First, we assume $\gamma >0$. In this setting, we have the following closed form:
\begin{align*}
    \ycwo(\thetav) &= \sum_{\y\in\Ycw}\frac{\exp\left(\langle \thetav,\y\rangle/\gamma\right)}{\sum_{\y'\in\Ycw}\exp\left(\langle \thetav,\y'\rangle/\gamma\right)}\cdot \y\,.
\end{align*}
Let $\sigma\in S_n$. We then have:
\begin{align*}
    \y^C_{\sigma(\w),\Omega}\left(\sigma(\thetav)\right) = \sum_{\y\in\mathcal{Y}^C_{\sigma(\w)}}\frac{\exp\left(\langle \sigma(\thetav),\y\rangle/\gamma\right)}{\sum_{\y'\in\mathcal{Y}^C_{\sigma(\w)}}\exp\left(\langle \sigma(\thetav),\y'\rangle/\gamma\right)}\cdot \y\,.
\end{align*}
Let us recall that the action of the permutation $\sigma$ on $\RR^n$ is that of an orthogonal endomorphism, giving:
\begin{align*}
    \forall \u,\v\in\RR^n,\quad \langle\sigma(\u),\v\rangle=\langle\u,\sigma^{-1}(\v)\rangle,\quad \text{and} \quad \langle\sigma(\u),\sigma(\v)\rangle=\langle\u,\v\rangle.
\end{align*}
Then, by definition of the feasible item selection set, we have:
\begin{align*}
    \mathcal{Y}^C_{\sigma(\w)} &= \left\{ \y\in\{0,1\}^n\mid \langle\sigma(\w),\y\rangle\leq C \right\}\\
    &= \left\{ \y\in\{0,1\}^n\mid \langle\w,\sigma^{-1}(\y)\rangle\leq C \right\}\\
    &= \left\{ \sigma(\y')\mid \y'\in\{0,1\}^n,\; \langle\w,\y'\rangle\leq C \right\} &&\text{(since $\sigma^{-1}(\{0,1\}^n)=\{0,1\}^n$)}\\
    &= \left\{ \sigma(\y') \mid \y'\in \Ycw \right\},
\end{align*}
where we denote $\sigma^{-1}(\y)\triangleq\y'$ for clarity.

Thus, we have in fact:
\begin{align*}
    \y^C_{\sigma(\w),\Omega}\left(\sigma(\thetav)\right) &= \sum_{\y\in\mathcal{Y}^C_{\sigma(\w)}}\frac{\exp\left(\langle \sigma(\thetav),\y\rangle/\gamma\right)}{\sum_{\y'\in\mathcal{Y}^C_{\sigma(\w)}}\exp\left(\langle \sigma(\thetav),\y'\rangle/\gamma\right)}\cdot \y\\
    &= \sum_{\y'\in\Ycw}\frac{\exp\left(\langle \sigma(\thetav),\sigma(\y')\rangle/\gamma\right)}{\sum_{\y''\in\Ycw}\exp\left(\langle \sigma(\thetav),\sigma(\y'')\rangle/\gamma\right)}\cdot \sigma(\y')\\
    &= \sum_{\y'\in\Ycw}\frac{\exp\left(\langle \thetav,\y'\rangle/\gamma\right)}{\sum_{\y''\in\Ycw}\exp\left(\langle \thetav,\y''\rangle/\gamma\right)}\cdot \sigma(\y')\\
    &= \sigma\left(\sum_{\y'\in\Ycw}\frac{\exp\left(\langle \thetav,\y'\rangle/\gamma\right)}{\sum_{\y''\in\Ycw}\exp\left(\langle \thetav,\y''\rangle/\gamma\right)}\cdot \y'\right)\\
    &= \sigma\left( \ycwo(\thetav) \right).
\end{align*}
For the Top-$k$ case, the exact same arguments hold, and the final expression of equivariance is only simplified because $\sigma(\w)=\w$ as $\w=\1$.

Second, for the case $\gamma=0$, we recover the unregularized operators, i.e., $\ycwo=\ycw$ and $\yko=\yk$, so that the equivariance property is simply directly implied by their definition as maximizers in Problems (\ref{pb:knapsack}) and (\ref{pb:topk}).

$\bullet$  ($\impliedby$)

We now show that Shannon's negative entropy is the only choice of a convex, separable regularizer yielding permutation-equivariant relaxed operators.

Let $\Omega:\triangle^2\to\RR$ be a convex regularization function. We assume it is separable, i.e., such that $\Omega(\q)= \omega(q_1)+\omega(q_2)$ for some lower semi-continuous convex function $\omega:[0,1]\to\RR$. Moreover, we assume that $\Omega(\q)=0$ for $\q\in\{\e_1,\e_2\}$ (i.e., $\omega(0)+\omega(1)=0$), and that $\Omega$ yields the permutation-equivariance property for the relaxed layers, i.e., that we have:
\begin{align*}
    \forall(\sigma,\thetav,\w)\in S_n\times \RR^n\times \NN^n, \quad \y^C_{\sigma(\w),\Omega}(\sigma(\thetav)) &= \sigma\left(\y^C_{\w,\Omega}(\thetav)\right),\\
    \y^k_{\1,\Omega}(\sigma(\thetav)) &= \sigma\left(\y^k_{\1,\Omega}(\thetav)\right).
\end{align*}

We must show that we then necessarily have $\Omega=-\gamma H^{s}$ for some $\gamma\geq 0$. In fact, we will show that $\Omega$ must necessarily yield an \emph{associative} smoothed maximum operator $\maxo$: indeed, both are equivalent by Lemma 3 in \citet{mensch_differentiable_2018}.

More precisely, we will assume that $\maxo$ is not associative for the sake of contradiction, and show that the permutation-equivariance property cannot be verified.

To do so, \cref{lemma:limit,lemma:equiv} will prove useful.

\begin{lemma}
\label{lemma:limit}
    Let $\Omega:\triangle^2\to\RR$ be convex, with $\Omega(\q)=\omega(q_1)+\omega(q_2)$, where $\omega:[0,1]\to\RR$ is lower semi-continuous, and such that $\Omega(\q)=0$ for $\q\in\{\e_1,\e_2\}$. We have, for any $x\in\RR$:
    \begin{align*}
        \lim_{c\to+\infty}\maxo(x+c, 0)-c = x\,.
    \end{align*}
\end{lemma}

\begin{lemma}
\label{lemma:equiv}
    Let $f:\RR^n\to\RR$ be any differentiable function. Then, we have:
    \begin{align*}
        f\circ\sigma=f \quad \text{for all $\sigma\in S_n$} \quad \iff \quad \nabla f\circ \sigma =\sigma\circ\nabla f \quad \text{for all $\sigma\in S_n$\,.}
    \end{align*}
    That is, a differentiable function is permutation-invariant if and only if its gradient is permutation-equivariant.
\end{lemma}

We give a proof of \cref{lemma:limit,lemma:equiv} in \cref{proof:limit,proof:equiv} for completeness. We now build simple instances of the Top-$k$ and Knapsack problems for which we show that the associativity of $\maxo$ is implied by the permutation-equivariance assumption for $\y^C_{\w,\Omega}$ and $\y^k_{\1,\Omega}$.

\begin{itemize}
    \item \textbf{Knapsack case.} Let $n=3$, $\w=\1$, $C=1$.  The smoothed dynamic program for solving the corresponding Knapsack problem with $\x=(x_1,x_2,x_3)\in\RR^3$ is depicted in \cref{fig:knapsack_and_topk_example_dags}. Unrolling the smoothed recursion from \cref{eq:smoothed_recursion} gives:
    \begin{align*}
        \maxcwo(\x) &= \Vwo[3,1]\\
        &= \maxo\left( x_3 + 0, \Vwo[2,1] \right)\\
        &= \maxo\left( x_3, \maxo\left( x_2 + 0, \Vwo[1,1] \right) \right)\\
        &= \maxo\left( x_3, \maxo\left( x_2, \maxo\left( x_1, 0\right) \right) \right).
    \end{align*}
    Moreover, since $\Omega$ is separable, it is commutative, so that $\maxo$ is also commutative. Thus, we get:
    \begin{align*}
        \maxcwo(\x) = \maxo\left( \maxo\left(\maxo\left(x_1, 0\right) , x_2\right), x_3\right).
    \end{align*}
    Now let $\sigma\in S_3$ be defined by $\sigma((x_1, x_2, x_3)) = (x_3, x_2, x_1)$ for all $\x\in\RR^3$. We have, similarly:
    \begin{align*}
        \mathsf{max}^C_{\sigma(\w),\Omega}(\sigma(\x))&=\maxcwo(\sigma(\x)) &&\text{(since $\w=\1$, so $\sigma(\w)=\w$)} \\
        &= \maxo\left( x_1, \maxo\left( x_2, \maxo\left( x_3, 0\right) \right) \right).
    \end{align*}
    Since we assumed for the sake of contradiction that $\maxo$ is not associative, we can find $(\theta_1,\theta_2,\theta_3)\in\RR^3$ such that $\maxo\left(\maxo\left( \theta_1, \theta_2 \right),\theta_3 \right)\neq \maxo\left(\theta_1,\maxo\left( \theta_2, \theta_3 \right) \right)$. 
    However, from what precedes, and using the distributivity of $+$ over $\maxo$, we also get, for any $c\in\RR$:
    \begin{align*}
        \maxcwo(\thetav+c\1) &= \maxo\left( \maxo\left(\maxo\left(\theta_1+c, 0\right) , \theta_2+c\right), \theta_3+c\right)\\
        &= c+\maxo\left( \maxo\left(\maxo\left(\theta_1+c, 0\right)-c , \theta_2\right), \theta_3\right),\\
        \maxcwo(\sigma(\thetav+c\1)) &= \maxo\left(\theta_1+c, \maxo\left(\theta_2+c, \maxo\left(\theta_3+c, 0\right) \right) \right)\\
        &= c+\maxo\left(\theta_1, \maxo\left(\theta_2, \maxo\left(\theta_3+c, 0\right)-c \right) \right).
    \end{align*}
    Thus, we have, for any $c\in\RR$:
    \begin{align*}
        \maxcwo(\thetav+c\1)-c
        &= \maxo\left( \maxo\left(\maxo\left(\theta_1+c, 0\right)-c , \theta_2\right), \theta_3\right),\\
        \maxcwo(\sigma(\thetav+c\1)) -c 
        &= \maxo\left(\theta_1, \maxo\left(\theta_2, \maxo\left(\theta_3+c, 0\right)-c \right) \right).
    \end{align*}
    By \cref{lemma:limit}, we have:
    \begin{align*}
        \maxo\left(\theta_1+c, 0\right)-c&\xrightarrow[c\to+\infty]{}\theta_1,\\
        \maxo\left(\theta_3+c, 0\right)-c&\xrightarrow[c\to+\infty]{}\theta_3.
    \end{align*}
    Since $\maxo$ is continuous, we can take the limit of the previous expressions:
    \begin{align*}
        \lim_{c\to+\infty}\maxcwo(\thetav+c\1)-c &= \maxo\left( \maxo\left(\theta_1 , \theta_2\right), \theta_3\right),\\
        \lim_{c\to+\infty}\maxcwo(\sigma(\thetav+c\1)) -c  &= \maxo\left(\theta_1, \maxo\left(\theta_2, \theta_3 \right) \right).
    \end{align*}
    Thus, by our definition of $\thetav$, we have:
    \begin{align*}
        \lim_{c\to+\infty}\maxcwo(\thetav+c\1)-c \neq \lim_{c\to+\infty}\maxcwo(\sigma(\thetav+c\1)) -c.
    \end{align*}
    
    Thus, we necessarily have:
    \begin{align*}
        \exists c\in\RR: \quad \maxcwo(\thetav+c\1)\neq\maxcwo(\sigma(\thetav+c\1)).
    \end{align*}
    Further, since $\ycwo=\nabla\maxcwo$, \cref{lemma:equiv} gives that this prevents us from having $\sigma(\ycwo(\x))=\ycwo(\sigma(\x))$ for all $\x\in\RR^n$, which contradicts the permutation-equivariance assumption (since we have $\w=\sigma(\w)=\1$).

    \item \textbf{Top-$k$ case.}
    Let $n=3$, $k=1$.  The smoothed dynamic program for solving the corresponding Top-$k$ problem with input $\x=(x_1,x_2,x_3)\in\RR^3$ is depicted in \cref{fig:knapsack_and_topk_example_dags}. Unrolling the smoothed recursion from \cref{eq:smoothed_recursion} gives:
    \begin{align*}
        \maxko(\x) &= \Vko[3,1]\\
        &= \maxo\left( x_3 + 0, \Vko[2,1] \right)\\
        &= \maxo\left( x_3, \maxo\left( x_2 + 0, \Vko[1,1] \right) \right)\\
        &= \maxo\left( x_3, \maxo\left( x_2, \maxo\left( x_1, -\infty\right) \right) \right).
    \end{align*}
    From the definition of smoothed maximum operators, we can easily see that $\maxo(x,-\infty)=x - \Omega(\e_1)$ for all $x\in\RR$, since the maximizing distribution in the definition of $\maxo$ is necessarily $\q=\e_1$. Moreover, since we assumed $\Omega(\q)=0$ when $\q\in\{\e_1,\e_2\}$, we have $\maxo(x,-\infty)=x$. Thus, we get in fact:
    \begin{align*}
        \maxko(\x) &= \maxo\left( x_3, \maxo\left( x_2, x_1\right) \right)\\
        &= \maxo\left(\maxo\left( x_1, x_2 \right),x_3 \right),
    \end{align*}
    where we also used commutativity of $\maxo$.
    Now let $\sigma\in S_3$ be defined by $\sigma((x_1, x_2, x_3)) = (x_3, x_2, x_1)$ for all $\x\in\RR^3$. We have, similarly:
    \begin{align*}
        \maxko(\sigma(\x)) &= \maxo\left(\maxo\left( x_3, x_2 \right),x_1 \right)\\
        &= \maxo\left(x_1,\maxo\left(x_2, x_3\right) \right).
    \end{align*}
    Since we assumed for the sake of contradiction that $\maxo$ is not associative, we can find $(\theta_1,\theta_2,\theta_3)\in\RR^3$ such that $\maxo\left(\maxo\left( \theta_1, \theta_2 \right),\theta_3 \right)\neq \maxo\left(\theta_1,\maxo\left( \theta_2, \theta_3 \right) \right)$. Thus, from previous calculations, we then have $\maxko(\thetav)\neq \maxko(\sigma(\thetav))$. Further, from \cref{lemma:equiv}, since $\yko=\nabla\maxko$, this prevents us from having $\yko(\sigma(\thetav))=\sigma(\yko(\thetav))$ for all $\thetav\in\RR^3$, which contradicts the permutation-equivariance assumption.
\end{itemize}

\begin{figure*}[t!]
\centering
\scalebox{0.9}{\begin{tikzpicture}[
    cell/.style={draw, minimum width=1.4cm, minimum height=0.8cm, anchor=center},
    oldarrow/.style={gray, dashed, thick, ->, >=Latex},
    newarrow/.style={blue, thick, ->, >=Latex},
    textcell/.style={minimum width=1cm, minimum height=1cm, anchor=center, align=left},
]

\matrix (m) [matrix of nodes, nodes={cell}, column sep=-\pgflinewidth, row sep=-\pgflinewidth] {
  0 & 0 \\
  0 & \footnotesize$\Vwo\![1,1]$ \\
  0 & \footnotesize$\Vwo\![2,1]$ \\
  0 & \footnotesize$\Vwo\![3,1]$ \\
};

\draw[oldarrow, shorten >=5pt, shorten <=7pt] (m-1-2.center) -- (m-2-2.center);
\draw[oldarrow, shorten >=12pt, shorten <=12pt] (m-1-1.center) -- (m-2-2.center);

\draw[oldarrow, shorten >=5pt, shorten <=7pt] (m-2-2.center) -- (m-3-2.center);
\draw[oldarrow, shorten >=12pt, shorten <=12pt] (m-2-1.center) -- (m-3-2.center);

\draw[newarrow, shorten >=5pt, shorten <=7pt] (m-3-2.center) -- (m-4-2.center);
\draw[newarrow, shorten >=12pt, shorten <=12pt] (m-3-1.center) -- (m-4-2.center);

\node[textcell] at ([xshift=1.2cm]m-1-2.east) {$[C=1]$};
\node[textcell] at ([xshift=1.2cm]m-2-2.east) {(\small$x_1,\;w_1=1$)};
\node[textcell] at ([xshift=1.2cm]m-3-2.east) {(\small$x_2,\;w_2=1$)};
\node[textcell] at ([xshift=1.2cm]m-4-2.east) {(\small$x_3,\;w_3=1$)};

\end{tikzpicture}}
\hspace{0.2cm}
\scalebox{0.9}{\begin{tikzpicture}[
    cell/.style={draw, minimum width=1.4cm, minimum height=0.8cm, anchor=center},
    oldarrow/.style={gray, dashed, thick, ->, >=Latex},
    newarrow/.style={blue, thick, ->, >=Latex},
    textcell/.style={minimum width=1cm, minimum height=1cm, anchor=center, align=left},
]

\matrix (m) [matrix of nodes, nodes={cell}, column sep=-\pgflinewidth, row sep=-\pgflinewidth] {
  0 & $-\infty$ \\
  0 & \footnotesize$\Vko[1,1]$ \\
  0 & \footnotesize$\Vko[2,1]$ \\
  0 & \footnotesize$\Vko[3,1]$ \\
};

\draw[oldarrow, shorten >=5pt, shorten <=7pt] (m-1-2.center) -- (m-2-2.center);
\draw[oldarrow, shorten >=12pt, shorten <=12pt] (m-1-1.center) -- (m-2-2.center);

\draw[oldarrow, shorten >=5pt, shorten <=7pt] (m-2-2.center) -- (m-3-2.center);
\draw[oldarrow, shorten >=12pt, shorten <=12pt] (m-2-1.center) -- (m-3-2.center);

\draw[newarrow, shorten >=5pt, shorten <=7pt] (m-3-2.center) -- (m-4-2.center);
\draw[newarrow, shorten >=12pt, shorten <=12pt] (m-3-1.center) -- (m-4-2.center);

\node[textcell] at ([xshift=0.8cm]m-1-2.east) {$[k=1]$};
\node[textcell] at ([xshift=0.8cm]m-2-2.east) {(\small$x_1$)};
\node[textcell] at ([xshift=0.8cm]m-3-2.east) {(\small$x_2$)};
\node[textcell] at ([xshift=0.8cm]m-4-2.east) {(\small$x_3$)};

\end{tikzpicture}}
\caption{Example DP tables for the smoothed Knapsack (left) and Top-$k$ (right) recursions.}
\label{fig:knapsack_and_topk_example_dags}
\end{figure*}
\end{proof}

\subsection{Proof of \cref{prop:sparsity} (\nameref{prop:sparsity})}
\label{proof:sparsity}
\begin{proof}
We prove the equivalence via the cycle $\text{(1)} \implies \text{(3)} \implies \text{(2)} \implies \text{(1)}$ (assuming a non-trivial feasible set $|\Ycw| \geq 2$, with $1 \leq k < n$ in the Top-$k$ case, as otherwise $\conv(\Ycw)$ is a singleton and all statements hold trivially). We write the proof in Knapsack notation, noting where the $-\infty$ boundary initialization of the Top-$k$ case ($\w=\1$, $C=k$) plays a role.

\textbf{$\bullet$ (1) $\implies$ (3).}
Assume $\lim_{t\to 0^+} \omega'(t) > -\infty$ and $\lim_{t\to 1^-} \omega'(t) < +\infty$. By strict convexity of $\omega$, the gap $\tau_\Omega \triangleq \lim_{t\to 1^-}\omega'(t) - \lim_{t\to 0^+}\omega'(t)$ satisfies $0 < \tau_\Omega < \infty$. Moreover, since $\Omega:\triangle^2\to\RR$ is convex and closed (so that $\maxo$ attains its maximum on $\triangle^2$), it is continuous on the compact simplex $\triangle^2$ and thus bounded: $M_\Omega \triangleq \sup_{\q \in \triangle^2} |\Omega(\q)| < \infty$.

For any $\gamma > 0$, the scaled regularizer $\gamma\Omega$ has saturation threshold $\gamma\tau_\Omega$ and maximum absolute value $\gamma M_\Omega$. By \citet[Definition 3, Propositions 7 and 8]{blondel_learning_2020}, for any $a, b \in \RR$, the local gradient mapping saturates to a hard decision once $|a - b| \geq \gamma\tau_\Omega$:
\begin{align}
    \label{eq:saturation}
    a - b \geq \gamma\tau_\Omega \implies (\nabla \mathsf{max}_{\gamma\Omega}(a,b))_1 = 1, \qquad
    b - a \geq \gamma\tau_\Omega \implies (\nabla \mathsf{max}_{\gamma\Omega}(a,b))_1 = 0.
\end{align}

Next, we bound the discrepancy between the smoothed and unregularized DP value tables across all feasible recursion states $(i,c)$ (i.e., $0 \leq c \leq C$ for Knapsack, and $0 \leq c \leq \min(i, k)$ for Top-$k$).
\begin{lemma}
\label{lemma:value_bound}
    Let $M_\Omega \triangleq \sup_{\q \in \triangle^2} |\Omega(\q)| < \infty$. For any $\thetav \in \RR^n$, $i \in \{0, \dots, n\}$, and any feasible capacity $c$ at step $i$, we have:
    \begin{align*}
        \left\lvert \Vwo[i,c] - \Vw[i,c] \right\rvert \leq i M_\Omega.
    \end{align*}
\end{lemma}
We prove \cref{lemma:value_bound} by induction in \cref{proof:value_bound}.

Applying \cref{lemma:value_bound} to $\gamma\Omega$ gives $|V^{\w}_{\gamma\Omega}[i,c] - \Vw[i,c]| \leq i\gamma M_\Omega$ for every feasible state $(i, c)$.

Now let $\thetav \in \RR^n$ admit a unique hard maximizer $\y^* \triangleq \ycw(\thetav)$, and trace the sequence of remaining capacities along its optimal DP path: $c_n \triangleq C$ and $c_{i-1} \triangleq c_i - y^*_i w_i$ for $i = n, \dots, 1$. Partition $[n]$ into forced steps and active binary steps along this path:
\begin{itemize}[topsep=0pt, itemsep=1pt, parsep=3pt, leftmargin=1em]
    \item \emph{Forced steps:} if $w_i > c_i$ (which forces $y^*_i = 0$, and includes $c_i = 0$ in Top-$k$), \cref{eq:smoothed_recursion} sets $V^{\w}_{\gamma\Omega}[i, c_i] = V^{\w}_{\gamma\Omega}[i-1, c_i]$, so $Q^{\w}_{\gamma\Omega}[i, c_i] = 0 = y^*_i$ for all $\gamma > 0$. Likewise, in Top-$k$ when $c_i = i$ (which forces $y^*_i = 1$), the skip parent $(i-1, i)$ equals $-\infty$, so $Q^{\1}_{\gamma\Omega}[i, i] = 1 = y^*_i$ for all $\gamma > 0$.
    \item \emph{Active binary steps ($\mathcal{A}$):} let $\mathcal{A} \subseteq [n]$ be the remaining indices, where both parent states $(i-1, c_i)$ and $(i-1, c_i - w_i)$ are feasible. For each $i \in \mathcal{A}$, define the hard local skipping advantage:
    \begin{align*}
        \Delta^{\text{hard}}_{i,c_i} \triangleq \Vw[i-1, c_i] - \bigl(\theta_i + \Vw[i-1, c_i - w_i]\bigr).
    \end{align*}
    Since $\y^*$ is the unique global maximizer, every active decision along the optimal path must be strict (otherwise, taking the alternative choice $1 - y^*_i$ and completing items $\{1, \dots, i-1\}$ optimally would yield a distinct optimal solution $\tilde{\y} \in \Ycw \setminus \{\y^*\}$). Thus $\Delta^{\text{hard}}_{i,c_i} > 0$ if $y^*_i = 0$ and $\Delta^{\text{hard}}_{i,c_i} < 0$ if $y^*_i = 1$.
\end{itemize}
If $\mathcal{A} \!=\! \emptyset$, then $Q^{\w}_{\gamma\Omega}[i, c_i] \!=\! y^*_i$ already holds for any $\gamma \!>\! 0$. Otherwise, define $\Delta^\star \!\triangleq\! \min_{i \in \mathcal{A}} |\Delta^{\text{hard}}_{i,c_i}|\! >\! 0$ and set:
\begin{align*}
    \gamma^* \triangleq \frac{\Delta^\star}{\tau_\Omega + 2(n-1)M_\Omega} > 0.
\end{align*}
Fix any $\gamma \in (0, \gamma^*]$ and $i \in \mathcal{A}$, and let $\Delta^{\gamma\Omega}_{i,c_i} \triangleq V^{\w}_{\gamma\Omega}[i-1, c_i] - (\theta_i + V^{\w}_{\gamma\Omega}[i-1, c_i - w_i])$ be the smoothed local advantage. By the triangle inequality and \cref{lemma:value_bound},
\begin{align*}
    \left\lvert \Delta^{\gamma\Omega}_{i,c_i} - \Delta^{\text{hard}}_{i,c_i} \right\rvert
    &\leq \left\lvert V^{\w}_{\gamma\Omega}[i-1, c_i] - \Vw[i-1, c_i] \right\rvert
    + \left\lvert V^{\w}_{\gamma\Omega}[i-1, c_i - w_i] - \Vw[i-1, c_i - w_i] \right\rvert \\
    &\leq 2(i-1)\gamma M_\Omega \leq 2(n-1)\gamma M_\Omega.
\end{align*}
Because $\gamma \leq \gamma^*$ and $\tau_\Omega > 0$, we have $2(n-1)\gamma M_\Omega < \gamma(\tau_\Omega + 2(n-1)M_\Omega) \leq \Delta^\star \leq |\Delta^{\text{hard}}_{i,c_i}|$, ensuring that $\Delta^{\gamma\Omega}_{i,c_i}$ has the same sign as $\Delta^{\text{hard}}_{i,c_i}$. Moreover, by the reverse triangle inequality,
\begin{align*}
    \left\lvert \Delta^{\gamma\Omega}_{i,c_i} \right\rvert
    \geq \left\lvert \Delta^{\text{hard}}_{i,c_i} \right\rvert - 2(n-1)\gamma M_\Omega
    \geq \Delta^\star - 2(n-1)\gamma M_\Omega
    \geq \gamma\tau_\Omega.
\end{align*}
By \cref{eq:saturation}, the local gate saturates to $Q^{\w}_{\gamma\Omega}[i, c_i] = y^*_i$ for all $i \in \mathcal{A}$, and hence for all $i \in [n]$.

Finally, in the backward pass (\cref{eq:backward_recursion}), initialized with $E^{\w}_{\gamma\Omega}[n, C] = 1$ and $0$ elsewhere, downward induction from $n$ to $1$ routes all unit mass deterministically along $(i, c_i)$, giving $E^{\w}_{\gamma\Omega}[i, c_i] = 1$ and $E^{\w}_{\gamma\Omega}[i, c] = 0$ for $c \neq c_i$ (if $c_i = 0$ with positive weights, $y^*_l = 0$ and no mass remains on $c \geq 1$ for all $l \leq i$). Therefore, $(\y^C_{\w,\gamma\Omega}(\thetav))_i = \sum_c E^{\w}_{\gamma\Omega}[i, c] Q^{\w}_{\gamma\Omega}[i, c] = y^*_i$ for all $i \in [n]$, i.e., $\y^C_{\w,\gamma\Omega}(\thetav) = \ycw(\thetav)$.

\textbf{$\bullet$ (3) $\implies$ (2).}
By Property~3 of \cref{sec:properties}, we have $\y^C_{\w,\gamma\Omega}(\thetav) = \ycwo(\thetav/\gamma)$ for all $\gamma > 0$ and $\thetav \in \RR^n$.

Let $\y \in \Ycw$ be arbitrary and set $\thetav \triangleq 2\y - \1 \in \RR^n$. For any $\y' \in \Ycw \setminus \{\y\}$, we have
\begin{align*}
    \langle \thetav, \y - \y' \rangle = \sum_{i: y_i=1}(1 - y'_i) + \sum_{i: y_i=0} y'_i = \|\y - \y'\|_1 > 0,
\end{align*}
so $\y = \ycw(\thetav)$ is the unique hard maximizer. By (3), there exists a threshold $\gamma^* > 0$ such that $\ycwo(\thetav/\gamma^*) = \y^C_{\w,\gamma^*\Omega}(\thetav) = \ycw(\thetav) = \y$, and therefore $\y \in \range(\ycwo)$ and $\Ycw \subseteq \range(\ycwo)$.

\textbf{$\bullet$ (2) $\implies$ (1).}
We proceed by contraposition. Suppose (1) fails. Since $\omega$ is convex on $(0,1)$, $\omega'$ is non-decreasing, so $\omega'(0^+) < +\infty$ and $\omega'(1^-) > -\infty$; hence the negation of (1) implies either $\lim_{t\to 0^+}\omega'(t) = -\infty$ or $\lim_{t\to 1^-}\omega'(t) = +\infty$, and in both cases we have that the difference is $\lim_{q \to 1^-}\omega'(q) - \lim_{q \to 0^+}\omega'(q) = +\infty$.

For any finite $a, b \in \RR$, the local objective $f(q) \triangleq q a + (1-q)b - \omega(q) - \omega(1-q)$ has derivative $f'(q) = (a - b) - (\omega'(q) - \omega'(1-q))$ on $(0,1)$, satisfying $\lim_{q \to 0^+} f'(q) = +\infty$ and $\lim_{q \to 1^-} f'(q) = -\infty$. Consequently, its unique maximizer $q^* = (\nabla \maxo(a, b))_1$ lies strictly in $(0, 1)$ whenever $a, b \in \RR$ are finite.

Now fix any $\thetav \in \RR^n$:
\begin{itemize}[topsep=0pt, itemsep=1pt, parsep=3pt, leftmargin=1em]
    \item In the Top-$k$ case ($1 \leq k < n$), both parent states $(n-1, k-1)$ and $(n-1, k)$ of the sink $(n, k)$ are feasible ($0 \leq k-1 < k \leq n-1$), so their smoothed DP values are finite. Thus $\Qwo[n, k] \in (0, 1)$, and since $\Ewo[n, k] = 1$ (with $\Ewo[n, c] = 0$ for $c \neq k$), we get $(\yko(\thetav))_n = \Qwo[n, k] \in (0, 1)$.
    \item In the Knapsack case ($|\Ycw| \geq 2$), since $\0 \in \Ycw$, there exists $\y \in \Ycw \setminus \{\0\}$, so $\{i \in [n] \mid w_i \leq C\}$ is non-empty; let $i^\star \triangleq \max\{i \in [n] \mid w_i \leq C\}$. For all $j > i^\star$, $w_j > C$ implies $\Qwo[j, C] = 0$, so the backward recursion (\cref{eq:backward_recursion}) preserves $\Ewo[i^\star, C] = 1$ and $\Ewo[i^\star, c] = 0$ for $c \neq C$. At $(i^\star, C)$, $w_{i^\star} \leq C$ and both parent values $\Vwo[i^\star-1, C - w_{i^\star}]$ and $\Vwo[i^\star-1, C]$ are finite, giving $\Qwo[i^\star, C] \in (0, 1)$ and $(\ycwo(\thetav))_{i^\star} = \Qwo[i^\star, C] \in (0, 1)$.
\end{itemize}
In both cases, $\ycwo(\thetav) \notin \{0,1\}^n \;\forall\thetav \in \RR^n$, so $\range(\ycwo) \cap \Ycw = \emptyset$, and (2) cannot hold.
\end{proof}

\subsection{Proof of \cref{prop:distribution} (\nameref{prop:distribution})}
\label{proof:distribution}

\begin{proof}
We prove the three claims of the proposition sequentially. We adopt the Knapsack notation ($\maxcwo$, $\VVwo$, etc.), but the proof holds identically for the Top-$k$ case.

\paragraph{Support of the distribution.}
We wish to show that for any $\y$ sampled from $\picwo$, we have $\y \in \Ycw$ (i.e., $\langle \w, \y \rangle \leq C$).
The sampling procedure in \cref{algo:sample} samples $y_i$ sequentially from $i=n$ down to $1$. Note that the remaining capacity $C_i$ available for items $\{1, \dots, i\}$ is initialized at $C_n = C$ and updated as $C_{i-1} = C_i - y_i w_i$.

Consider the step for item $i$ with current remaining capacity $C_i$.
If $w_i > C_i$, the forward DP recursion in \cref{eq:smoothed_recursion} is defined via the first case:
\begin{align*}
    \Vwo[i, C_i] = \Vwo[i-1, C_i].
\end{align*}
Consequently, the derivative with respect to $\theta_i$ is zero:
\begin{align*}
    \Qwo[i, C_i] = \frac{\partial \Vwo[i, C_i]}{\partial \theta_i} = 0.
\end{align*}
By definition of $\picwo$, the probability of selecting item $i$ is given by $\picwo(y_i=1 \mid C_i) = \Qwo[i, C_i]$. Thus, if $w_i > C_i$, we have $\picwo(y_i=1 \mid C_i) = 0$, forcing $y_i=0$.
This ensures that the capacity constraint is never violated at any step. Thus, $\sum_{i=1}^n y_i w_i \leq C$, and the support of $\picwo$ is a subset of $\Ycw$.

\paragraph{Expectation as the relaxed operator.}
We now show that $\EE_{Y\sim\picwo}[Y] = \ycwo(\thetav)$. Since $\ycwo(\thetav) = \nabla \maxcwo(\thetav)$, and recalling from \cref{sec:layer_derivation} that the gradient is computed as:
\begin{align*}
    (\ycwo(\thetav))_i=(\nabla \maxcwo(\thetav))_i = \sum_{c=0}^C \Ewo[i,c] \Qwo[i,c],
\end{align*}
we must show that the marginal probability of selecting item $i$ under $\picwo$, denoted $\mu_i$, is equal to $\sum_{c=0}^C \Ewo[i,c] \Qwo[i,c]$.

Let $C_i$ be the random variable representing the remaining capacity for items $\{1 \dots i\}$ during the ancestral sampling process. The probability of selecting item $i$ is:
\begin{align*}
    \mu_i = \picwo(y_i=1) = \sum_{c=0}^C \picwo(y_i=1 \mid C_i=c) \picwo(C_i=c).
\end{align*}
By definition of the sampling distribution, $\picwo(y_i=1 \mid C_i=c) = \Qwo[i,c]$. Thus, it suffices to show that $\picwo(C_i=c) = \Ewo[i,c]$ holds for all $i\in[n]$ and $c\in\{0,\dots,C\}$.

We proceed by induction on $i$, moving backwards from $n$ to $1$.
\begin{itemize}
    \item \textbf{Base case ($i=n$):} The sampling always starts with capacity $C$. Thus $\picwo(C_n=C) = 1$ and $\picwo(C_n=c)=0$ for $c\neq C$.
    Matching this, the backward pass in \cref{algo:layer} initializes $\Ewo[n,C]=1$ and $\Ewo[n,c]=0$ otherwise. The base case holds.

    \item \textbf{Inductive step ($i<n$):} Assume $\picwo(C_{i+1}=c) = \Ewo[i+1,c]$ for all $c$.
    We express the probability of the capacity $C_i$ at step $i$. The capacity $c$ at step $i$ can be reached from step $i+1$ in two ways:
    \begin{enumerate}
        \item We had capacity $c$ at $i+1$ and chose $y_{i+1}=0$ (skip).
        \item We had capacity $c+w_{i+1}$ at $i+1$ and chose $y_{i+1}=1$ (pick).
    \end{enumerate}
    Formally:
    \begin{align*}
        \picwo(C_i=c) =\;& \picwo(C_{i+1}=c)\times\picwo(y_{i+1}=0 \mid C_{i+1}=c)\\
        &+ \picwo(C_{i+1}=c+w_{i+1})\times \picwo(y_{i+1}=1 \mid C_{i+1}=c+w_{i+1}).
    \end{align*}
    Plugging the definition of the sampling probabilities as $\picwo(y_k=1|C_k=\cdot)\triangleq\Qwo[k,\cdot]$ and the induction hypothesis, we get:
    \begin{align*}
        \picwo(C_i=c) &= \Ewo[i+1,c] (1 - \Qwo[i+1,c])\\
        &\quad+ \Ewo[i+1,c+w_{i+1}] \Qwo[i+1,c+w_{i+1}].
    \end{align*}
    This is exactly the backward recursion derived for $\Ewo$ in \cref{eq:backward_recursion}.
\end{itemize}
Thus, $\Ewo[i,c]$ represents the marginal probability that the remaining capacity is $c$ at step $i$. We therefore have:
\begin{align*}
    \EE_{Y\sim\picwo}[Y_i] =\mu_i= \sum_{c=0}^C \Ewo[i,c] \Qwo[i,c] = (\ycwo(\thetav))_i\,,
\end{align*}
which finally gives:
\begin{align*}
    \EE_{Y\sim\picwo}[Y] = \ycwo(\thetav).
\end{align*}

\paragraph{Recovery of the Gibbs distribution.}
Let $\Omega = -\gamma H^{s}$. The smoothed maximum becomes the \emph{log-sum-exp} function:
\begin{align*}
    \maxo(a, b) = \gamma \log (\exp(a/\gamma) + \exp(b/\gamma)).
\end{align*}
The forward recursion becomes:
\begin{align*}
    \Vwo[i,c] = \gamma \log \left( \exp(\Vwo[i-1,c]/\gamma) + \exp((\theta_i + \Vwo[i-1, c-w_i])/\gamma) \right).
\end{align*}
The local probability $\Qwo[i,c]$ becomes the standard softmax probability:
\begin{align*}
    \Qwo[i,c] &= \frac{\exp((\theta_i + \Vwo[i-1, c-w_i])/\gamma)}{\exp(\Vwo[i,c]/\gamma)}, \\
    1 - \Qwo[i,c] &= \frac{\exp(\Vwo[i-1, c]/\gamma)}{\exp(\Vwo[i,c]/\gamma)}.
\end{align*}
Now consider the probability of a full sampled vector $\y$. Let $c_i$ be the sequence of remaining capacities realized during sampling (with $c_n=C$ and $c_{i-1} = c_i - y_i w_i$).
\begin{align*}
    \picwo(\y) &= \prod_{i=n}^1 \picwo(y_i \mid c_i) \\
    &= \prod_{i=n}^1 \begin{cases}
        \Qwo[i, c_i] & \text{if } y_i=1 \\
        1-\Qwo[i, c_i] & \text{if } y_i=0
    \end{cases} \\
    &= \prod_{i=n}^1 \frac{\exp\left( (y_i \theta_i + \Vwo[i-1, c_{i-1}]) / \gamma \right)}{\exp(\Vwo[i, c_i]/\gamma)}.
\end{align*}
Note that the term $\Vwo[i-1, c_{i-1}]$ in the numerator of step $i$ cancels with the denominator term $\exp(\Vwo[i-1, c_{i-1}]/\gamma)$ of step $i-1$. This telescoping product yields:
\begin{align*}
    \picwo(\y) &= \frac{\exp(\Vwo[0, c_0]/\gamma)}{\exp(\Vwo[n, C]/\gamma)} \prod_{i=n}^1 \exp(y_i \theta_i / \gamma) \\
    &\propto\exp\left( \frac{\sum_{i=1}^n y_i \theta_i}{\gamma} \right) = \exp(\langle \thetav, \y \rangle / \gamma).
\end{align*}
\end{proof}

\subsection{Proof of \cref{lemma:limit}}
\label{proof:limit}

\begin{proof}
    Let $x,c\in\RR$. We have:
    \begin{align*}
        \maxo(x+c, 0)-c &= \max_{p\in[0,1]}\left\langle \begin{pmatrix}
            x+c\\
            0
        \end{pmatrix}, \begin{pmatrix}
            p\\
            1-p
        \end{pmatrix}\right\rangle - \Omega\left(\begin{pmatrix}
            p\\
            1-p
        \end{pmatrix}\right)-c\\
        &= \max_{p\in[0,1]}(x+c)p - \omega(p)-\omega(1-p)-c\\
        &= \max_{p\in[0,1]}xp + c(p-1)- \omega(p)-\omega(1-p).
    \end{align*}
    Define:
    \begin{align*}
        h_c(p)\triangleq xp + c(p-1)- \omega(p)-\omega(1-p).
    \end{align*}
    Since $\omega$ is lower semi-continuous, $p\mapsto\omega(p)+\omega(1-p)$ also is lower semi-continuous. Thus, $h_c$ is upper semi-continuous, as the difference between a continuous function and a lower semi-continuous one. The quantity $\max_{p\in[0,1]} h_c(p)$ is thus well-defined by Weierstrass' extreme value theorem, and $h_c$ attains its maximum on $[0,1]$. We are computing $\ell\triangleq \lim_{c\to+\infty}\max_{p\in[0,1]} h_c(p)$.
    
    \textbf{Lower bound.} By definition of the maximum, we have:
    \begin{align*}
        \max_{p\in[0,1]} h_c(p) &\geq h_c(1)\\
        &= x - \omega(1)-\omega(0)\\
        &= x - \Omega(\e_1)\\
        &= x.
    \end{align*}
    This gives us a lower bound $\ell \geq x$ for the limit.
    
    \textbf{Upper bound.} We now show that $\limsup_{c\to+\infty}\max_{p\in[0,1]} h_c(p) \leq x$. For any $c\in\RR$, let $p^\star_c\in \argmax_{{p\in[0,1]}} h_c(p)$, which is well-defined by upper semi-continuity of $h_c$. First, we show that $p^\star_c\xrightarrow[c\to+\infty]{}1$. From the lower bound, we know $h_c(p^\star_c)\geq h_c(1)=x$. Substituting the definition of $h_c$, we have:
    \begin{align*}
        &xp^\star_c + c(p^\star_c-1)- \omega(p^\star_c)-\omega(1-p^\star_c) \geq x\\
        \iff &c(p^\star_c-1) \geq x(1-p^\star_c)+\omega(p^\star_c)+\omega(1-p^\star_c).
    \end{align*}
    Since the function $\alpha(p)\triangleq x(1-p)+\omega(p)+\omega(1-p)$ is lower semi-continuous on the compact set $[0,1]$, it is bounded below and attains its minimum. Thus, we have $c(p^\star_c-1)\geq K$, where $K\triangleq \min_{p\in[0,1]}\alpha(p)>-\infty$.

    Since $p^\star_c\in[0,1]$, we have $p^\star_c-1\leq0$, giving:
    \begin{align*}
        \frac{K}{c}\leq p^\star_c-1\leq 0, \quad\text{(for $c>0$)}
    \end{align*}
    which implies $p^\star_c-1\xrightarrow[c\to+\infty]{}0$, i.e., $\lim_{c\to+\infty}p^\star_c=1$.
    
    We now find the limit of $h_c(p^\star_c)$. Define $\beta(p)\triangleq xp-\omega(p)-\omega(1-p)$. Since $\omega$ is lower semi-continuous, $\beta$ is upper semi-continuous. We can write $h_c(p)=\beta(p)+c(p-1)$. Since $p^\star_c\in[0,1]$, we have $c(p^\star_c-1)\leq 0$ for $c\geq0$, giving the inequality:
    \begin{align*}
        h_c(p^\star_c) = \beta(p^\star_c) + c(p^\star_c-1)\leq \beta(p^\star_c).
    \end{align*}
    Now we can take the $\limsup$:
    \begin{align*}
        \limsup_{c\to+\infty} h_c(p^\star_c) \leq \limsup_{c\to+\infty} \beta(p^\star_c).
    \end{align*}
    By upper semi-continuity of $\beta$, since $\lim_{c\to+\infty}p^\star_c=1$, we have:
    \begin{align*}
        \limsup_{c\to+\infty} \beta(p^\star_c) \leq \beta(1) = x-\omega(1)-\omega(0)=x-\Omega(\e_1)=x.
    \end{align*}

    Since we have $\liminf_{c\to+\infty}h_c(p^\star_c)\geq x$ from the lower bound and $\limsup_{c\to+\infty}h_c(p^\star_c)\leq x$, we conclude that the limit exists and is equal to $x$:
    \begin{align*}
        \ell = \lim_{c\to+\infty}\max_{p\in[0,1]}h_c(p)= \lim_{c\to+\infty}h_c(p^\star_c) = x.
    \end{align*}
\end{proof}

\subsection{Proof of \cref{lemma:equiv}}
\label{proof:equiv}

\begin{proof}
    $\bullet$ ($\implies$)
    Assume that $f\circ\sigma = f$ for all $\sigma\in S_n$. 
    Fix $\sigma\in S_n$ and $\x\in\RR^n$. 
    For any direction $\v\in\RR^n$, by the chain rule we have:
    \begin{align*}
        \langle \nabla(f\circ\sigma)(\x), \v\rangle 
        &= \langle\nabla f(\sigma(\x)),\nabla\sigma(\x)\v\rangle .
    \end{align*}
    Since $\sigma$ acts as a linear operator, $\nabla\sigma(\x)$ acts simply as the permutation $\sigma$ itself, giving:
    \begin{align*}
       \langle \nabla(f\circ\sigma)(\x), \v\rangle 
        = \langle\nabla f(\sigma(\x)),\sigma(\v)\rangle .
    \end{align*}
    Since by assumption $f\circ\sigma=f$, we have $\nabla(f\circ\sigma)(\x)=\nabla f(\x)$.
    Therefore:
    \begin{align*}
         \langle \nabla f(\x), \v\rangle
        &= \langle\nabla f(\sigma(\x)),\sigma(\v)\rangle\\
        &= \langle\sigma^{-1}(\nabla f(\sigma(\x))),\v\rangle.
    \end{align*}
    Since this holds for all $\v\in\RR^n$, we get in fact:
    \begin{align*}
        \forall \x\in\RR^n, \nabla f(\x) &= \sigma^{-1}(\nabla f(\sigma(\x)))\\
        \iff \forall \x\in\RR^n, \sigma (\nabla f(\x)) &= \nabla f(\sigma(\x)).
    \end{align*}
   
    Thus, $\nabla f\circ\sigma = \sigma\circ\nabla f$ for all $\sigma\in S_n$.

    $\bullet$ ($\impliedby$)
    Conversely, assume that $\nabla f\circ\sigma = \sigma\circ\nabla f$ for all $\sigma\in S_n$. 
    Fix $\sigma\in S_n$ and define:
    \begin{align*}
        g(\x) \triangleq (f\circ\sigma)(\x) - f(\x).
    \end{align*}
    Then, by the chain rule:
    \begin{align*}
        \nabla g(\x) 
        &= \sigma^{-1} (\nabla f(\sigma(\x))) - \nabla f(\x).
    \end{align*}
    Using the equivariance assumption $\nabla f(\sigma(\x)) = \sigma(\nabla f(\x))$, we have:
    \begin{align*}
        \nabla g(\x) 
        &= \sigma^{-1} (\nabla f(\sigma(\x))) - \nabla f(\x)\\
        &= \nabla f(\x) - \nabla f(\x) = 0.
    \end{align*}
    Hence, $\nabla g(\x) = 0$ for all $\x\in\RR^n$, which implies that $g$ is constant. 
    To determine this constant, note that we clearly have $g(\0) = 0$, since $\sigma(\0)=\0$. 
    Therefore:
    \begin{align*}
        f(\sigma(\x)) = f(\x), \quad \forall \x\in\RR^n.
    \end{align*}
    Thus, $f\circ\sigma = f$ for all $\sigma\in S_n$.
\end{proof}

\subsection{Proof of \cref{lemma:value_bound}}
\label{proof:value_bound}

\begin{proof}
    We proceed by induction on $i$. For $i=0$, the feasible states are $c \in \{0,\dots,C\}$ (Knapsack) or $c=0$ (Top-$k$), where $\Vwo[0,c] = \Vw[0,c] = 0$.
    Assume the bound holds at $i-1 \in \{0,\dots,n-1\}$, and consider a feasible state $(i, c)$.

    If $w_i > c$ (which includes $c = 0$ in Top-$k$), $\Vwo[i,c] = \Vwo[i-1,c]$ and $\Vw[i,c] = \Vw[i-1,c]$, so the error remains bounded by $(i-1)M_\Omega \leq i M_\Omega$.
    In Top-$k$ when $c = i$, the skip parent $(i-1, i)$ equals $-\infty$, so $\Vko[i,i] = \theta_i + \Vko[i-1,i-1] - \Omega(\e_1)$ while $\Vk[i,i] = \theta_i + \Vk[i-1,i-1]$, giving $|\Vko[i,i] - \Vk[i,i]| \leq (i-1)M_\Omega + |\Omega(\e_1)| \leq i M_\Omega$.

    Otherwise, both $a = \theta_i + \Vw[i-1, c-w_i]$ and $b = \Vw[i-1, c]$ are finite, and we have that $\Vwo[i,c] = \maxo(a+\epsilon_a, b+\epsilon_b)$ with $|\epsilon_a|, |\epsilon_b| \leq (i-1)M_\Omega$ by the induction hypothesis. By the triangle inequality, $|\maxo(\x) - \max(\x)| \leq M_\Omega$, and $1$-Lipschitz continuity of $\max$ in $\|\cdot\|_\infty$:
    \begin{align*}
        |\Vwo[i,c] - \Vw[i,c]|
        &\leq |\maxo(a+\epsilon_a, b+\epsilon_b) - \max(a+\epsilon_a, b+\epsilon_b)|\\
        &\quad+ |\max(a+\epsilon_a, b+\epsilon_b) - \max(a, b)| \\
        &\leq M_\Omega + \max(|\epsilon_a|, |\epsilon_b|)\\
        &\leq M_\Omega + (i-1)M_\Omega\\
        &= i M_\Omega. \qedhere
    \end{align*}
\end{proof}

\newpage    

\section{Algorithmic Derivations}
\label{sec:algo_derivations}

\subsection{Backtracking as a backward pass}
\label{sec:backtracking}

We now turn to the derivation of the backward pass, which enables to compute $\ycw(\thetav)=\nabla\maxcw(\thetav)$ and $\yk(\thetav)=\nabla\maxk(\thetav)$. In this unregularized case, we recover the idea of backtracking. We will write everything in the Knapsack case, as everything holds for the Top-$k$ case by letting $C=k$ and $\w=\1$ (the slight change due to the difference in padding values of the DP table does not impact differentiation).

Let $\Ew[i, c]\triangleq \frac{\partial \Vw[n,C]}{\partial \Vw[i,c]}$ be the main object of interest in this derivation, and define:
\begin{align*}
    \Qw[i,c] &\triangleq \frac{\partial \Vw[i,c]}{\partial\theta_i} \\
    &= \begin{cases}
        0 \quad \text{if $w_i>c$,}\\
        \mathbbm{1}\left(\left\{\max\left(\theta_i+\Vw[i-1,c-w_i]\;,\,\Vw[i-1,c]\right) = \theta_i+\Vw[i-1,c-w_i]\right\}\right) &\text{else,}\\
    \end{cases}
\end{align*}
so that the matrix $\QQw$ stores a track of the maximizers in \cref{eq:recursion_knapsack} during the forward pass (we have $\Qw[i,c]=1\iff$ item $i$ is needed to reach value $\Vw[i,c]$). As $\Vw[i,c]$ only directly influences $\Vw[i+1,c]$ and $\Vw[i+1,c+w_{i+1}]$, we have:

\begin{align*}
    \Ew[i,c] &= \frac{\partial \Vw[n,C]}{\partial \Vw[i+1,c]}\times\frac{\partial \Vw[i+1,c]}{\partial \Vw[i,c]} + \frac{\partial \Vw[n,C]}{\partial \Vw[i+1,c+w_{i+1}]}\times\frac{\partial \Vw[i+1,c+w_{i+1}]}{\partial \Vw[i,c]}\\
    &= \Ew[i+1,c]\times\frac{\partial \Vw[i+1,c]}{\partial \Vw[i,c]} +  \Ew[i+1,c+w_{i+1}]\times\frac{\partial \Vw[i+1,c+w_{i+1}]}{\partial \Vw[i,c]}.
\end{align*}

Then, from the recursion in \cref{eq:recursion_knapsack} (for $c \geq 0$, with $c + w_{i+1} \leq C$ for the second line), we have:
\begin{align*}
    \Vw[i+1,c] &= \begin{cases}
        \Vw[i,c] &\text{if $w_{i+1}>c$,}\\
        \max\left(\theta_{i+1}+ \Vw[i, c-w_{i+1}]\;,\;\Vw[i,c]\right) &\text{else,}
    \end{cases}\\
    \Vw[i+1,c+w_{i+1}] &= \max\left(\theta_{i+1}+ \Vw[i, c]\;,\;\Vw[i,c+w_{i+1}]\right),
\end{align*}
which further gives:

\begin{align*}
    \frac{\partial \Vw[i+1,c]}{\partial \Vw[i,c]} &= \begin{cases}
        1 \quad\quad \text{if $w_{i+1}>c$,} \\
        \mathbbm{1}\left(\left\{\max\left(\theta_{i+1}+ \Vw[i, c-w_{i+1}],\Vw[i,c]\right) = \Vw[i,c]\right\}\right)  &\text{else,}
    \end{cases}\\
    &= 1 - \Qw[i+1,c]\,,  \\
     \frac{\partial \Vw[i+1,c+w_{i+1}]}{\partial \Vw[i,c]} &= \mathbbm{1}\left(\left\{\max\left(\theta_{i+1}+ \Vw[i,c],\Vw[i,c+w_{i+1}]\right) = \theta_{i+1}+\Vw[i,c]\right\}\right)\\
     &= \Qw[i+1,c+w_{i+1}]\,.
\end{align*}
Thus, we finally have the following backward recursion:

\begin{equation*}
    \Ew[i,c] = \Ew[i+1,c]\times(1-\Qw[i+1,c]) + \Ew[i+1,c+w_{i+1}]\times \Qw[i+1,c+w_{i+1}].
\end{equation*}
The recursion is initialized with $\Ew[n,C]=\frac{\partial \Vw[n,C]}{\partial \Vw[n,C]}=1$, and $\forall c<C,\;\Ew[n,c]=0$.

The hard maximizing mask $\ycw(\thetav)=\nabla\maxcw(\thetav)=\frac{\partial \Vw[n,C]}{\partial\thetav}$ is then recovered by noting that since $\theta_i$ only directly influences the row $\Vw[i,\colon]$, we have:
\begin{align*}
    \frac{\partial \Vw[n,C]}{\partial \theta_i} &= \sum_{c=0}^C \frac{\partial \Vw[n,C]}{\partial \Vw[i,c]}\times\frac{\partial \Vw[i,c]}{\partial\theta_i} \\
 &= \sum_{c=0}^C\Ew[i,c]\times \Qw[i,c],
\end{align*}
giving the following compact expression for the maximizing mask:
\begin{align*}
    \ycw(\thetav) = (\EEw[1\colon,\colon]\circ\QQw[1\colon,\colon])\cdot  \1,
\end{align*}
where $\circ$ denotes the Hadamard product and $\1\in\RR^{C+1}$ is the all-ones vector.

\subsection{Derivation of the layer}
\label{sec:layer_derivation}

We now turn to the derivation of the backward pass to compute the relaxed layers. We will write everything in the Knapsack case, with $\ycwo(\thetav) = \nabla\maxcwo(\thetav)$. All results also hold for the Top-$k$ case, by setting $\w=\1$ and $C=k$ (the only slight change due to the difference in the padding values for $\VVwo$ does not impact differentiation).

We first recall here the smoothed recursion given in \cref{eq:smoothed_recursion}:
\begin{align*}
    &\Vwo[i,c] = \begin{cases}
        \Vwo[i-1, c] &\text{if $w_{i}>c$,}\\
        \maxo\left(\theta_{i}+\Vwo[i-1, c-w_{i}],\; \Vwo[i-1, c] \right) &\text{else.}
    \end{cases}
\end{align*}

Define the following objects:
\begin{align*}
    \Ewo[i,c]&\triangleq \frac{\partial \Vwo[n,C]}{\partial \Vwo[i,c]},\\
    \Qwo[i,c]&\triangleq\frac{\partial\Vwo[i,c]}{\partial\theta_i}.
\end{align*}

The matrix $\QQwo$ is computed and stored during the forward pass (see \cref{algo:value}). As we have
\begin{align*}
    \nabla_{\x}\maxo(x_1, x_2) \in \triangle^2,\quad\text{i.e.,}\quad\frac{\partial \maxo(x_1,x_2)}{\partial x_1}+\frac{\partial \maxo(x_1,x_2)}{\partial x_2}  =1,
\end{align*}
one can easily check that we have:
\begin{align}
    \frac{\partial \Vwo[i,c]}{\partial\Vwo[i-1,c]} &=1- \Qwo[i,c]\label{eq:dv_2}\,,\\
    \frac{\partial \Vwo[i,c]}{\partial\Vwo[i-1,c-w_i]} &=\frac{\partial \Vwo[i,c]}{\partial\theta_i}= \Qwo[i,c]. \label{eq:dv_1}
\end{align}
These identities will prove useful in the following. We can now derive the computation of $\Ewo[i,c]$. As $\Vwo[i,c]$ only directly influences $\Vwo[i+1,c]$ and $\Vwo[i+1,c+w_{i+1}]$, we have (similarly to the unregularized case, see \cref{sec:backtracking}):

\begin{align*}
    \Ewo[i,c] = \Ewo[i+1,c]\times \frac{\partial \Vwo[i+1,c]}{\partial \Vwo[i,c]} + \Ewo[i+1,c+w_{i+1}]\times \frac{\partial \Vwo[i+1,c+w_{i+1}]}{\partial \Vwo[i,c]}.
\end{align*}
Moreover, using \cref{eq:dv_1,eq:dv_2}, we have:

\begin{align*}
    \frac{\partial \Vwo[i+1,c]}{\partial \Vwo[i,c]} &= 1-\Qwo[i+1,c]\,,\\
    \frac{\partial \Vwo[i+1,c+w_{i+1}]}{\partial \Vwo[i,c]} &= \Qwo[i+1,c+w_{i+1}]\,.
\end{align*}

Thus, we have the following backward recursion:

\begin{equation}
    \label{eq:backward_recursion}
    \Ewo[i,c] = \Ewo[i+1,c]\times (1-\Qwo[i+1,c]) + \Ewo[i+1,c+w_{i+1}]\times \Qwo[i+1,c+w_{i+1}].
\end{equation}
The recursion is initialized with $\Ewo[n,C] = \frac{\partial \Vwo[n,C]}{\partial \Vwo[n,C]} = 1$ and $\forall c<C, \;\Ewo[n,c]=0$. The gradient $\ycwo(\thetav)=\nabla\maxcwo(\thetav)=\frac{\partial \Vwo[n,C]}{\partial\thetav}$ is then recovered by noting that, since $\theta_i$ only directly influences the row $\Vwo[i,\colon]$, we have:
\begin{align*}
    \frac{\partial \Vwo[n,C]}{\partial\theta_i} = \sum_{c=0}^C \frac{\partial \Vwo[n,C]}{\partial \Vwo[i,c]}\times \frac{\partial \Vwo[i,c]}{\partial\theta_i} = \sum_{c=0}^C \Ewo[i,c] \times  \Qwo[i,c].
\end{align*}
Finally, we get the following compact expression:
\begin{align*}
    \ycwo(\thetav) = (\EEwo[1\colon,\colon]\circ\QQwo[1\colon,\colon])\cdot  \1,
\end{align*}
where $\circ$ denotes the Hadamard product and $\1\in\RR^{C+1}$ is the all-ones vector.

\subsection{Derivation of the directional derivative}
\label{sec:directional_derivation}

We now turn to the computation of the directional derivative $\langle \nabla\maxcwo(\thetav),\, \z\rangle=\langle \ycwo(\thetav),\, \z\rangle$ of the soft value in direction $\z\in\RR^n$ (i.e., the inner product between the layer and a cotangent vector $\z$). We will write everything in the Knapsack case, with $\ycwo(\thetav) = \nabla\maxcwo(\thetav)$. All results also hold for the Top-$k$ case, by setting $\w=\1$ and $C=k$ (the only slight change due to the difference in the padding values for $\VVwo$ does not impact differentiation).

Let $\z\in\RR^n$ be a direction. We define here:
\begin{align*}
    \Vdwo[i,c]\triangleq \left\langle \frac{\partial\Vwo[i,c]}{\partial\thetav},\z \right\rangle.
\end{align*}

Since $\Vwo[i,c]$ only directly depends on $\Vwo[i-1,c-w_i]$, $\Vwo[i-1,c]$ and $\theta_i$, we have:

\begin{align*}
    \Vdwo[i,c] &= \left\langle \frac{\partial\Vwo[i,c]}{\partial\thetav},\z \right\rangle \\
    &= \biggl\langle \frac{\partial \Vwo[i,c]}{\partial \Vwo[i-1,c-w_i]}\times\frac{\partial\Vwo[i-1,c-w_i]}{\partial\thetav} \\
    &+ \frac{\partial  \Vwo[i,c]}{\partial  \Vwo[i-1,c]}\times\frac{\partial\Vwo[i-1,c]}{\partial\thetav} + \frac{\partial \Vwo[i,c]}{\partial \theta_i}\times\frac{\partial\theta_i}{\partial\thetav}\,,\, \z\biggr\rangle \\
    &= \frac{\partial \Vwo[i,c]}{\partial \Vwo[i-1,c-w_i]} \times\Vdwo[i-1,c-w_i] \\
    &+ \frac{\partial \Vwo[i,c]}{\partial \Vwo[i-1,c]}\times\Vdwo[i-1,c] + \frac{\partial \Vwo[i,c]}{\partial \theta_i}\times z_i .
\end{align*}
Then, using \cref{eq:dv_1,eq:dv_2} gives the following forward recursion:
\begin{align}
\label{eq:vdot_recursion}
\Vdwo[i,c] &= \left(\Vdwo[i-1,c-w_i]+z_i\right)\times\Qwo[i,c] + \Vdwo[i-1,c]\times (1-\Qwo[i,c]),  
\end{align}

which is initialized with $\Vdwo[0,\colon] = 0$, and ends at $\Vdwo[n,C]= \left\langle \frac{\partial\Vwo[n,C]}{\partial\thetav},\z \right\rangle = \langle \ycwo(\thetav), \z\rangle$.

\subsection{Derivation of the vector-Jacobian product}
\label{sec:vjp_derivation}

Here, we derive the computation of the vector-Jacobian product $\z^\top(\nabla_\thetav\ycwo(\thetav))$ for a given cotangent vector $\z\in\RR^n$. As stated in \cref{sec:continuous_forward}, since we have $\ycwo=\nabla\maxcwo$, the Jacobian $\nabla\ycwo(\thetav)$ is equal to the Hessian $\nabla^2\maxcwo(\thetav)$, so it is in fact symmetric for all $\thetav\in\RR^n$. Thus, we can view the VJP $\z^\top(\nabla_\thetav\ycwo(\thetav))$ as the corresponding JVP $(\nabla_\thetav\ycwo(\thetav))\z$, for any tangent/cotangent vector $\z\in\RR^n$. Further, we use the fact that $(\nabla_\thetav\ycwo(\thetav))\z = \nabla_\thetav\langle \ycwo(\thetav), \z\rangle$, and focus on the differentiation of the directional derivative $\langle\ycwo(\thetav),\z\rangle= \Vdwo[n,C]$. Thus, our goal is to compute $\frac{\partial\Vdwo[n,C]}{\partial\thetav}$.

As $\theta_i$ only directly influences the rows $\Vwo[i,\colon]$ and $\Vdwo[i,\colon]$, we have:
\begin{equation}
\label{eq:colors}
    \frac{\partial\Vdwo[n,C]}{\partial\theta_i} = \sum_{c=0}^C  \textcolor{orange}{\frac{\partial \Vdwo[n,C]}{\partial\Vwo[i,c]}}\times\textcolor{green}{\frac{\partial\Vwo[i,c]}{\partial\theta_i}} + \textcolor{blue}{\frac{\partial \Vdwo[n,C]}{\partial\Vdwo[i,c]}}\times\textcolor{red}{\frac{\partial\Vdwo[i,c]}{\partial\theta_i}}.
\end{equation}

We now turn to the derivation of every element in \cref{eq:colors} needed to compute $\frac{\partial\Vdwo[n,C]}{\partial\theta_i}$.

To begin with, the second term is already computed as we simply have $\textcolor{green}{\frac{\partial\Vwo[i,c]}{\partial\theta_i}} = \Qwo[i,c]$.

Now, we compute the third term $\textcolor{blue}{\frac{\partial \Vdwo[n,C]}{\partial\Vdwo[i,c]}}$. As $\Vdwo[i,c]$ only directly influences $\Vdwo[i+1,c]$ and $\Vdwo[i+1,c+w_{i+1}]$, we have the recursion:
\begin{align*}
    \textcolor{blue}{\frac{\partial \Vdwo[n,C]}{\partial\Vdwo[i,c]}} = \frac{\partial \Vdwo[n,C]}{\partial\Vdwo[i+1,c]}\times\frac{\partial \Vdwo[i+1,c]}{\partial\Vdwo[i,c]} + \frac{\partial \Vdwo[n,C]}{\partial\Vdwo[i+1,c+w_{i+1}]}\times\frac{\partial \Vdwo[i+1,c+w_{i+1}]}{\partial\Vdwo[i,c]}.
\end{align*}

However, \cref{eq:vdot_recursion} gives:
\begin{align}
    \Vdwo[i+1,c] &= \left(\Vdwo[i,c-w_{i+1}]+z_{i+1}\right)\times\Qwo[i+1,c] \notag\\
    &+ \Vdwo[i,c]\times (1-\Qwo[i+1,c]),\label{eq:Vdot_iplus_c}\\[10pt]
    \Vdwo[i+1,c+w_{i+1}] &= \left(\Vdwo[i,c]+z_{i+1}\right)\times\Qwo[i+1,c+w_{i+1}]\notag\\
    &+ \Vdwo[i,c+w_{i+1}]\times (1-\Qwo[i+1,c+w_{i+1}])\label{eq:Vdot_iplus_cplus}.
\end{align}

Thus, as $\Vdwo[i,c]$ only directly influences $\Vdwo[i+1,c]$ and $\Vdwo[i+1,c+w_{i+1}]$, we get, by differentiating \cref{eq:Vdot_iplus_c,eq:Vdot_iplus_cplus}:

\begin{align*}
    \frac{\partial\Vdwo[i+1,c]}{\partial\Vdwo[i,c]} &= 1-\Qwo[i+1,c],\\
    \frac{\partial\Vdwo[i+1,c+w_{i+1}]}{\partial\Vdwo[i,c]} &= \Qwo[i+1,c+w_{i+1}].
\end{align*}
Thus, we have in fact the following recursion:
\begin{align*}
    \textcolor{blue}{\frac{\partial \Vdwo[n,C]}{\partial\Vdwo[i,c]}} &= \frac{\partial \Vdwo[n,C]}{\partial\Vdwo[i+1,c]}\times(1-\Qwo[i+1,c])\\
    &\quad + \frac{\partial \Vdwo[n,C]}{\partial\Vdwo[i+1,c+w_{i+1}]}\times\Qwo[i+1,c+w_{i+1}],
\end{align*}
and $\textcolor{blue}{\frac{\partial \Vdwo[n,C]}{\partial\Vdwo[i,c]}}$ is defined by the exact same recursion as $\Ewo[i,c]$, which is given by \cref{eq:backward_recursion}. As they are both initialized by $\frac{\partial \Vdwo[n,C]}{\partial\Vdwo[n,C]} = \Ewo[n,C] = 1$ and $\forall c<C,\;\frac{\partial \Vdwo[n,C]}{\partial\Vdwo[n,c]} = \Ewo[n,c] = 0$, the two sequences are in fact equal:
\begin{align*}
    \textcolor{blue}{\frac{\partial \Vdwo[n,C]}{\partial\Vdwo[i,c]}} = \Ewo[i,c].
\end{align*}

Now, we derive the red term $\textcolor{red}{\frac{\partial\Vdwo[i,c]}{\partial\theta_i}}$. As $\theta_i$ does not influence $\Vdwo[i-1,c-w_i]$, $\Vdwo[i-1,c]$, $\Vwo[i-1,c-w_i]$, or $\Vwo[i-1,c]$, the differentiation of the forward recursion in \cref{eq:vdot_recursion} gives:

\begin{align*}
    \textcolor{red}{\frac{\partial\Vdwo[i,c]}{\partial\theta_i}} &= \left(\Vdwo[i-1,c-w_i]+z_i\right)\times \frac{\partial\Qwo[i,c]}{\partial\theta_i} - \Vdwo[i-1,c]\times \frac{\partial\Qwo[i,c]}{\partial\theta_i}\\
    &= \left(\Vdwo[i-1,c-w_i]- \Vdwo[i-1,c] +z_i\right)\times \frac{\partial\Qwo[i,c]}{\partial\theta_i}.\\
\end{align*}

Finally, for the orange term, let us define:
\begin{align*}
    \Edwo[i,c]\triangleq \textcolor{orange}{\frac{\partial \Vdwo[n,C]}{\partial\Vwo[i,c]}}.
\end{align*}

As $\Vwo[i,c]$ only directly influences $\Vwo[i+1,c]$, $\Vwo[i+1,c+w_{i+1}]$, $\Vdwo[i+1,c]$ and $\Vdwo[i+1,c+w_{i+1}]$, we have:
\begin{align*}
    \Edwo[i,c] &= \textcolor{orange}{\frac{\partial \Vdwo[n,C]}{\partial\Vwo[i,c]}}\\
    &= \frac{\partial \Vdwo[n,C]}{\partial\Vwo[i+1,c]}\times \frac{\partial \Vwo[i+1,c]}{\partial\Vwo[i,c]} + \frac{\partial \Vdwo[n,C]}{\partial\Vwo[i+1,c+w_{i+1}]}\times \frac{\partial \Vwo[i+1,c+w_{i+1}]}{\partial\Vwo[i,c]}\\
    &+ \frac{\partial \Vdwo[n,C]}{\partial\Vdwo[i+1,c]}\times \frac{\partial \Vdwo[i+1,c]}{\partial\Vwo[i,c]} + \frac{\partial \Vdwo[n,C]}{\partial\Vdwo[i+1,c+w_{i+1}]}\times \frac{\partial\Vdwo[i+1,c+w_{i+1}]}{\partial\Vwo[i,c]}\\
    &= \Edwo[i+1,c] \times (1-\Qwo[i+1,c]) + \Edwo[i+1,c+w_{i+1}]\times 
    \Qwo[i+1,c+w_{i+1}]\\
    &+ \Ewo[i+1,c]\times \frac{\partial \Vdwo[i+1,c]}{\partial\Vwo[i,c]} + \Ewo[i+1,c+w_{i+1}]\times \frac{\partial\Vdwo[i+1,c+w_{i+1}]}{\partial\Vwo[i,c]}.\\
\end{align*}

Thus, we need to compute the two terms $\frac{\partial \Vdwo[i+1,c]}{\partial\Vwo[i,c]}$ and $\frac{\partial\Vdwo[i+1,c+w_{i+1}]}{\partial\Vwo[i,c]}$ in order to get a recursive formulation of $\Edwo[i,c]= \textcolor{orange}{\frac{\partial \Vdwo[n,C]}{\partial\Vwo[i,c]}}$.

As $\Vwo[i,c]$ does not influence $\Vdwo[i,c-w_{i+1}], \,\Vdwo[i,c]$ or $z_{i+1}$, differentiating \cref{eq:Vdot_iplus_c} gives on the one hand:

\begin{align*}
    \frac{\partial\Vdwo[i+1,c]}{\partial\Vwo[i,c]} &= \left(\Vdwo[i,c-w_{i+1}] - \Vdwo[i,c]+z_{i+1}\right)\times\frac{\partial\Qwo[i+1,c]}{\partial\Vwo[i,c]}.
\end{align*}

On the other hand, as $\Vwo[i,c]$ does not influence $\Vdwo[i,c]$ or $\Vdwo[i,c+w_{i+1}]$, differentiating \cref{eq:Vdot_iplus_cplus} gives:

\begin{align*}
    \frac{\partial \Vdwo[i+1,c+w_{i+1}]}{\partial\Vwo[i,c]} &= \left(\Vdwo[i,c] - \Vdwo[i,c+w_{i+1}]+z_{i+1}\right)\times\frac{\partial\Qwo[i+1,c+w_{i+1}]}{\partial\Vwo[i,c]}.
\end{align*}

Thus, $\Edwo[i,c]= \textcolor{orange}{\frac{\partial \Vdwo[n,C]}{\partial\Vwo[i,c]}}$ is given by the following backward recursion:

\begin{align*}
    \Edwo[i,c] &= \Edwo[i+1,c] \times (1-\Qwo[i+1,c])   + \Edwo[i+1,c+w_{i+1}]\times 
    \Qwo[i+1,c+w_{i+1}]\\
    &+\Ewo[i+1,c]\times 
    \left(\Vdwo[i,c-w_{i+1}] - \Vdwo[i,c]+z_{i+1}\right)\times\frac{\partial\Qwo[i+1,c]}{\partial\Vwo[i,c]}\\
    &+ \Ewo[i+1,c+w_{i+1}]\times \left(\Vdwo[i,c] - \Vdwo[i,c+w_{i+1}]+z_{i+1}\right)\times\frac{\partial\Qwo[i+1,c+w_{i+1}]}{\partial\Vwo[i,c]}.
\end{align*}

Thus, to summarize our results, we have (for any choice of $\Omega$):

\begin{align*}
    \textcolor{orange}{\frac{\partial \Vdwo[n,C]}{\partial\Vwo[i,c]}} &= \Edwo[i,c]\\
    &= \Edwo[i+1,c] \times (1-\Qwo[i+1,c])   + \Edwo[i+1,c+w_{i+1}]\times 
    \Qwo[i+1,c+w_{i+1}]\\
    &+\Ewo[i+1,c]\times 
    \left(\Vdwo[i,c-w_{i+1}] - \Vdwo[i,c]+z_{i+1}\right)\times\frac{\partial\Qwo[i+1,c]}{\partial\Vwo[i,c]}\\
    &+ \Ewo[i+1,c+w_{i+1}]\times \left(\Vdwo[i,c] - \Vdwo[i,c+w_{i+1}]+z_{i+1}\right)\\
    &\quad\quad\times\frac{\partial\Qwo[i+1,c+w_{i+1}]}{\partial\Vwo[i,c]}\,,\\
    \textcolor{green}{\frac{\partial\Vwo[i,c]}{\partial\theta_i}} &= \Qwo[i,c]\,,\\
    \textcolor{blue}{\frac{\partial \Vdwo[n,C]}{\partial\Vdwo[i,c]}} &= \Ewo[i,c]\,,\\
    \textcolor{red}{\frac{\partial\Vdwo[i,c]}{\partial\theta_i}} 
    &= \left(\Vdwo[i-1,c-w_i]- \Vdwo[i-1,c] +z_i\right)\times \frac{\partial\Qwo[i,c]}{\partial\theta_i}\,.\\
\end{align*}

\newpage

\section{Pseudo-code}
\label{sec:pseudo_code}

\begin{algorithm}
\caption{Computation of the VJP $\z^\top(\nabla_\thetav\ycwo(\thetav))$\,, for general $\Omega$}
\label{algo:vjp}
\begin{algorithmic}
\INPUT Item values $\thetav \in \RR^n$, item weights $\w\in\NN^n$, capacity $C \in \NN$, regularization function $\Omega$, cotangent vector $\z\in\RR^n$, outputs $\VVwo,\QQwo,\EEwo$ of \cref{algo:layer}.
\OUTPUT Vector-Jacobian product $\z^\top(\nabla_\thetav\ycwo(\thetav))$.
\STATE Initialize $\VVdwo\gets\0\in\RR^{(n+1)\times(C+1)}$, $\EEdwo\gets\0\in\RR^{(n+1)\times(C+1)}$, and $\mathrm{VJP}\gets\0\in\RR^n$.

\STATE \COMMENT{forward recursion to compute $\VVdwo$}
\FOR{$i=1$ {\bfseries to} $n$}
    \STATE \COMMENT{the loop on $c$ is parallelizable (wavefront computation)}
    \FOR{$c=0$ {\bfseries to} $C$}
        \STATE $\Vdwo[i,c] \gets \left(\Vdwo[i-1,c-w_i]+z_i\right)\times\Qwo[i,c] + \Vdwo[i-1,c]\times (1-\Qwo[i,c])$
    \ENDFOR
\ENDFOR

\STATE \COMMENT{backward recursion to compute $\EEdwo$}
\FOR{$i=n-1$ {\bfseries to} $1$}
    \STATE \COMMENT{the loop on $c$ is parallelizable (wavefront computation)}
    \FOR{$c=0$ {\bfseries to} $C$}
    \STATE $\begin{aligned}
        \Edwo[i,c] \gets &\left(\Edwo[i+1,c] \times (1-\Qwo[i+1,c])\right)\\
        &+ \left(\Edwo[i+1,c+w_{i+1}]\times 
        \Qwo[i+1,c+w_{i+1}]\right)\\
        &+\left(\Ewo[i+1,c]\times 
        \left(\Vdwo[i,c-w_{i+1}] - \Vdwo[i,c]+z_{i+1}\right)\times\frac{\partial\Qwo[i+1,c]}{\partial\Vwo[i,c]}\right)\\
        &+ \biggl(\Ewo[i+1,c+w_{i+1}]\times \left(\Vdwo[i,c] - \Vdwo[i,c+w_{i+1}]+z_{i+1}\right)\\
        &\quad\quad \times\frac{\partial\Qwo[i+1,c+w_{i+1}]}{\partial\Vwo[i,c]}\biggr)
        \end{aligned}$
    \ENDFOR
\ENDFOR

\STATE \COMMENT{Final VJP assembly (both the loops on $i$ and $c$ are parallelizable)}
\FOR{$i=1$ {\bfseries to} $n$}
    \FOR{$c=0$ {\bfseries to} $C$}
    \STATE \COMMENT{colors match the ones used in \cref{sec:vjp_derivation}}
    \STATE $\begin{aligned}
        \mathrm{VJP}[i]\gets\mathrm{VJP}[i]&+\left(\textcolor{orange}{\Edwo[i,c]}\times\textcolor{green}{\Qwo[i,c]}\right)\\
        &+ \left(\textcolor{blue}{\Ewo[i,c]}\times\textcolor{red}{\left(\Vdwo[i-1,c-w_i]- \Vdwo[i-1,c] +z_i\right)}\textcolor{red}{\times\frac{\partial\Qwo[i,c]}{\partial\theta_i}}\right)
    \end{aligned}$
    \ENDFOR
\ENDFOR

\STATE $\z^\top(\nabla_\thetav\ycwo(\thetav))\gets\mathrm{VJP}$

\STATE \textbf{return} $\z^\top(\nabla_\thetav\ycwo(\thetav))$
\end{algorithmic}
\end{algorithm}

\newpage

\section{Additional Material}
\label{sec:additional_material}

\subsection{Visualization of the DP recursions}
\label{sec:visualization_dp_recursions}

\begin{figure}[H]
\centering
\scalebox{0.99}{\begin{tikzpicture}[
    cell/.style={draw, minimum width=1.4cm, minimum height=0.8cm, anchor=center},
    textcell/.style={minimum width=1cm, minimum height=1cm, anchor=center, align=left},
    arrow/.style={blue, thick, ->, >=Latex}
]

\matrix (m) [matrix of nodes, nodes={cell}, column sep=-\pgflinewidth, row sep=-\pgflinewidth] {
  0 & 0 & 0 & 0 \\
  0 & \footnotesize$V^{\!\w}\![1,1]$ & \footnotesize$V^{\w}[1,2]$ & \footnotesize$V^{\!\w}\![1,3]$ \\
  0 & \footnotesize$V^{\!\w}\![2,1]$ & \footnotesize$V^{\!\w}\![2,2]$ & \footnotesize$V^{\!\w}\![2,3]$ \\
  0 & \footnotesize$V^{\!\w}\![3,1]$ & \footnotesize$V^{\!\w}\![3,2]$ & \footnotesize$V^{\!\w}\![3,3]$ \\
  0 & \footnotesize$V^{\!\w}\![4,1]$ & \footnotesize$V^{\!\w}\![4,2]$ & \footnotesize$V^{\!\w}\![n,C]$ \\
};

\node[textcell] at ([xshift=1.5cm]m-1-4.east) {$[C=3]$};
\node[textcell] at ([xshift=1.5cm]m-2-4.east) {(\small$\theta_1=2,\;w_1=2$)};
\node[textcell] at ([xshift=1.5cm]m-3-4.east) {(\small$\theta_2=1,\;w_2=1$)};
\node[textcell] at ([xshift=1.5cm]m-4-4.east) {(\small$\theta_3=-1,\;w_3=3$)};
\node[textcell] at ([xshift=1.5cm]m-5-4.east) {(\small$\theta_4=3,\;w_4=2$)};

\end{tikzpicture}}
\scalebox{0.99}{\begin{tikzpicture}[
    cell/.style={draw, minimum width=1.4cm, minimum height=0.8cm, anchor=center},
    textcell/.style={minimum width=1cm, minimum height=1cm, anchor=center, align=left},
    arrow/.style={blue, thick, ->, >=Latex}
]

\matrix (m) [matrix of nodes, nodes={cell}, column sep=-\pgflinewidth, row sep=-\pgflinewidth] {
  0 & 0 & 0 & 0 \\
  0 & 0 & 2 & 2 \\
  0 & * & * & * \\
  0 & * & * & * \\
  0 & * & * & * \\
};

\node[textcell] at ([xshift=1.5cm]m-1-4.east) {$[C=3]$};
\node[textcell] at ([xshift=1.5cm]m-2-4.east) {(\small$\theta_1=2,\;w_1=2$)};
\node[textcell] at ([xshift=1.5cm]m-3-4.east) {(\small$\theta_2=1,\;w_2=1$)};
\node[textcell] at ([xshift=1.5cm]m-4-4.east) {(\small$\theta_3=-1,\;w_3=3$)};
\node[textcell] at ([xshift=1.5cm]m-5-4.east) {(\small$\theta_4=3,\;w_4=2$)};

\draw[arrow, shorten >=5pt, shorten <=7pt] (m-1-2.center) -- (m-2-2.center);
\draw[arrow, shorten >=10pt, shorten <=10pt] (m-1-1.center) -- (m-2-3.center);

\draw[arrow, shorten >=5pt, shorten <=7pt] (m-1-3.center) -- (m-2-3.center);
\draw[arrow, shorten >=10pt, shorten <=10pt] (m-1-2.center) -- (m-2-4.center);

\draw[arrow, shorten >=5pt, shorten <=7pt] (m-1-4.center) -- (m-2-4.center);

\end{tikzpicture}}

\scalebox{0.99}{\begin{tikzpicture}[
    cell/.style={draw, minimum width=1.4cm, minimum height=0.8cm, anchor=center},
    textcell/.style={minimum width=1cm, minimum height=1cm, anchor=center, align=left},
    arrow/.style={blue, thick, ->, >=Latex}
]

\matrix (m) [matrix of nodes, nodes={cell}, column sep=-\pgflinewidth, row sep=-\pgflinewidth] {
  0 & 0 & 0 & 0 \\
  0 & 0 & 2 & 2 \\
  0 & 1 & 2 & 3 \\
  0 & * & * & * \\
  0 & * & * & * \\
};

\node[textcell] at ([xshift=1.5cm]m-1-4.east) {$[C=3]$};
\node[textcell] at ([xshift=1.5cm]m-2-4.east) {(\small$\theta_1=2,\;w_1=2$)};
\node[textcell] at ([xshift=1.5cm]m-3-4.east) {(\small$\theta_2=1,\;w_2=1$)};
\node[textcell] at ([xshift=1.5cm]m-4-4.east) {(\small$\theta_3=-1,\;w_3=3$)};
\node[textcell] at ([xshift=1.5cm]m-5-4.east) {(\small$\theta_4=3,\;w_4=2$)};

\draw[arrow, shorten >=5pt, shorten <=7pt] (m-2-2.center) -- (m-3-2.center);
\draw[arrow, shorten >=12pt, shorten <=12pt] (m-2-1.center) -- (m-3-2.center);

\draw[arrow, shorten >=5pt, shorten <=7pt] (m-2-3.center) --  (m-3-3.center);
\draw[arrow, shorten >=12pt, shorten <=12pt] (m-2-2.center) -- (m-3-3.center);

\draw[arrow, shorten >=5pt, shorten <=7pt] (m-2-4.center) -- (m-3-4.center);
\draw[arrow, shorten >=12pt, shorten <=12pt] (m-2-3.center) -- (m-3-4.center);

\draw[arrow, gray, dashed, shorten >=5pt, shorten <=7pt] (m-1-2.center) -- (m-2-2.center);
\draw[arrow, gray, dashed, shorten >=10pt, shorten <=10pt] (m-1-1.center) -- (m-2-3.center);

\draw[arrow, gray, dashed, shorten >=5pt, shorten <=7pt] (m-1-3.center) -- (m-2-3.center);
\draw[arrow, gray, dashed, shorten >=10pt, shorten <=10pt] (m-1-2.center) -- (m-2-4.center);

\draw[arrow, gray, dashed, shorten >=5pt, shorten <=7pt] (m-1-4.center) -- (m-2-4.center);

\end{tikzpicture}}
\scalebox{0.99}{\begin{tikzpicture}[
    cell/.style={draw, minimum width=1.4cm, minimum height=0.8cm, anchor=center},
    textcell/.style={minimum width=1cm, minimum height=1cm, anchor=center, align=left},
    arrow/.style={blue, thick, ->, >=Latex}
]

\matrix (m) [matrix of nodes, nodes={cell}, column sep=-\pgflinewidth, row sep=-\pgflinewidth] {
  0 & 0 & 0 & 0 \\
  0 & 0 & 2 & 2 \\
  0 & 1 & 2 & 3 \\
  0 & 1 & 2 & 3 \\
  0 & 1 & 3 & 4 \\
};

\node[textcell] at ([xshift=1.5cm]m-1-4.east) {$[C=3]$};
\node[textcell] at ([xshift=1.5cm]m-2-4.east) {(\small$\theta_1=2,\;w_1=2$)};
\node[textcell] at ([xshift=1.5cm]m-3-4.east) {(\small$\theta_2=1,\;w_2=1$)};
\node[textcell] at ([xshift=1.5cm]m-4-4.east) {(\small$\theta_3=-1,\;w_3=3$)};
\node[textcell] at ([xshift=1.5cm]m-5-4.east) {(\small$\theta_4=3,\;w_4=2$)};

\draw[arrow, shorten >=5pt, shorten <=7pt] (m-4-2.center) -- (m-5-2.center);
\draw[arrow, shorten >=5pt, shorten <=7pt] (m-4-3.center) -- (m-5-3.center);
\draw[arrow, shorten >=5pt, shorten <=7pt] (m-4-4.center) -- (m-5-4.center); 

\draw[arrow, shorten >=5pt, shorten <=7pt] (m-4-1.center) -- (m-5-3.center);
\draw[arrow, shorten >=5pt, shorten <=7pt] (m-4-2.center) -- (m-5-4.center);

\draw[arrow, gray, dashed, shorten >=5pt, shorten <=7pt] (m-1-2.center) -- (m-2-2.center);
\draw[arrow, gray, dashed, shorten >=10pt, shorten <=10pt] (m-1-1.center) -- (m-2-3.center);

\draw[arrow, gray, dashed, shorten >=5pt, shorten <=7pt] (m-1-3.center) -- (m-2-3.center);
\draw[arrow, gray, dashed, shorten >=10pt, shorten <=10pt] (m-1-2.center) -- (m-2-4.center);

\draw[arrow, gray, dashed, shorten >=5pt, shorten <=7pt] (m-1-4.center) -- (m-2-4.center);

\draw[arrow, gray, dashed, shorten >=5pt, shorten <=7pt] (m-2-2.center) -- (m-3-2.center);
\draw[arrow, gray, dashed, shorten >=12pt, shorten <=12pt] (m-2-1.center) -- (m-3-2.center);

\draw[arrow, gray, dashed, shorten >=5pt, shorten <=7pt] (m-2-3.center) --  (m-3-3.center);
\draw[arrow, gray, dashed, shorten >=12pt, shorten <=12pt] (m-2-2.center) -- (m-3-3.center);

\draw[arrow, gray, dashed, shorten >=5pt, shorten <=7pt] (m-2-4.center) -- (m-3-4.center);
\draw[arrow, gray, dashed, shorten >=12pt, shorten <=12pt] (m-2-3.center) -- (m-3-4.center);

\draw[arrow, gray, dashed, shorten >=5pt, shorten <=7pt] (m-3-2.center) -- (m-4-2.center);
\draw[arrow, gray, dashed, shorten >=5pt, shorten <=7pt] (m-3-3.center) -- (m-4-3.center);
\draw[arrow, gray, dashed, shorten >=5pt, shorten <=7pt] (m-3-4.center) -- (m-4-4.center);

\draw[arrow, gray, dashed, shorten >=5pt, shorten <=7pt] (m-3-1.center) -- (m-4-4.center);

\end{tikzpicture}}
\caption{Illustration of the DP recursion table for the Knapsack problem, with $\thetav=(2,1,-1,3)^\top$, $\w=(2,1,3,2)^\top$ and $C=3$. Arrows indicate the influence of values on others, illustrating the DAG structure underlying the recursion. See \cref{sec:dag_ddp} for more details.}
\label{fig:knapsack_dp_tables}
\end{figure}

\begin{figure}[H]
\centering
\scalebox{0.99}{\begin{tikzpicture}[
    cell/.style={draw, minimum width=1.4cm, minimum height=0.8cm, anchor=center},
    textcell/.style={minimum width=1cm, minimum height=1cm, anchor=center, align=left},
    arrow/.style={blue, thick, ->, >=Latex}
]

\matrix (m) [matrix of nodes, nodes={cell}, column sep=-\pgflinewidth, row sep=-\pgflinewidth] {
  0 & $-\infty$ & $-\infty$ & $-\infty$ \\
  0 & \footnotesize$\Vk[1,1]$ & \footnotesize$\Vk[1,2]$ & \footnotesize$\Vk[1,3]$ \\
  0 & \footnotesize$\Vk[2,1]$ & \footnotesize$\Vk[2,2]$ & \footnotesize$\Vk[2,3]$ \\
  0 & \footnotesize$\Vk[3,1]$ & \footnotesize$\Vk[3,2]$ & \footnotesize$\Vk[3,3]$ \\
  0 & \footnotesize$\Vk[4,1]$ & \footnotesize$\Vk[4,2]$ & \footnotesize$\Vk[4,3]$ \\
  0 & \footnotesize$\Vk[5,1]$ & \footnotesize$\Vk[5,2]$ & \footnotesize$\Vk[n,k]$ \\
};

\node[textcell] at ([xshift=0.9cm]m-1-4.east) {$[k=3]$};
\node[textcell] at ([xshift=0.9cm]m-2-4.east) {(\small$\theta_1=3$)};
\node[textcell] at ([xshift=0.9cm]m-3-4.east) {(\small$\theta_2=-1$)};
\node[textcell] at ([xshift=0.9cm]m-4-4.east) {(\small$\theta_3=4$)};
\node[textcell] at ([xshift=0.9cm]m-5-4.east) {(\small$\theta_4=-2$)};
\node[textcell] at ([xshift=0.9cm]m-6-4.east) {(\small$\theta_5=2$)};

\end{tikzpicture}}
\hspace{0.2cm}
\scalebox{0.99}{\begin{tikzpicture}[
    cell/.style={draw, minimum width=1.4cm, minimum height=0.8cm, anchor=center},
    textcell/.style={minimum width=1cm, minimum height=1cm, anchor=center, align=left},
    arrow/.style={blue, thick, ->, >=Latex}
]

\matrix (m) [matrix of nodes, nodes={cell}, column sep=-\pgflinewidth, row sep=-\pgflinewidth] {
  0 & $-\infty$ & $-\infty$ & $-\infty$ \\
  0 & 3 & $-\infty$ & $-\infty$ \\
  0 & * & * & * \\
  0 & * & * & * \\
  0 & * & * & * \\
  0 & * & * & * \\
};

\node[textcell] at ([xshift=0.9cm]m-1-4.east) {$[k=3]$};
\node[textcell] at ([xshift=0.9cm]m-2-4.east) {(\small$\theta_1=3$)};
\node[textcell] at ([xshift=0.9cm]m-3-4.east) {(\small$\theta_2=-1$)};
\node[textcell] at ([xshift=0.9cm]m-4-4.east) {(\small$\theta_3=4$)};
\node[textcell] at ([xshift=0.9cm]m-5-4.east) {(\small$\theta_4=-2$)};
\node[textcell] at ([xshift=0.9cm]m-6-4.east) {(\small$\theta_5=2$)};

\draw[arrow, shorten >=5pt, shorten <=7pt] (m-1-2.center) -- (m-2-2.center);
\draw[arrow, shorten >=12pt, shorten <=12pt] (m-1-1.center) -- (m-2-2.center);

\draw[arrow, shorten >=5pt, shorten <=7pt] (m-1-3.center) -- (m-2-3.center);
\draw[arrow, shorten >=12pt, shorten <=12pt] (m-1-2.center) -- (m-2-3.center);

\draw[arrow, shorten >=5pt, shorten <=7pt] (m-1-4.center) -- (m-2-4.center);
\draw[arrow, shorten >=12pt, shorten <=12pt] (m-1-3.center) -- (m-2-4.center);

\end{tikzpicture}}

\scalebox{0.99}{\begin{tikzpicture}[
    cell/.style={draw, minimum width=1.4cm, minimum height=0.8cm, anchor=center},
    textcell/.style={minimum width=1cm, minimum height=1cm, anchor=center, align=left},
    arrow/.style={blue, thick, ->, >=Latex}
]

\matrix (m) [matrix of nodes, nodes={cell}, column sep=-\pgflinewidth, row sep=-\pgflinewidth] {
  0 & $-\infty$ & $-\infty$ & $-\infty$ \\
  0 & 3 & $-\infty$ & $-\infty$ \\
  0 & 3 & 2 & $-\infty$ \\
  0 & * & * & * \\
  0 & * & * & * \\
  0 & * & * & * \\
};

\node[textcell] at ([xshift=0.9cm]m-1-4.east) {$[k=3]$};
\node[textcell] at ([xshift=0.9cm]m-2-4.east) {(\small$\theta_1=3$)};
\node[textcell] at ([xshift=0.9cm]m-3-4.east) {(\small$\theta_2=-1$)};
\node[textcell] at ([xshift=0.9cm]m-4-4.east) {(\small$\theta_3=4$)};
\node[textcell] at ([xshift=0.9cm]m-5-4.east) {(\small$\theta_4=-2$)};
\node[textcell] at ([xshift=0.9cm]m-6-4.east) {(\small$\theta_5=2$)};

\draw[arrow, shorten >=5pt, shorten <=7pt] (m-2-2.center) -- (m-3-2.center);
\draw[arrow, shorten >=12pt, shorten <=12pt] (m-2-1.center) -- (m-3-2.center);

\draw[arrow, shorten >=5pt, shorten <=7pt] (m-2-3.center) --  (m-3-3.center);
\draw[arrow, shorten >=12pt, shorten <=12pt] (m-2-2.center) -- (m-3-3.center);

\draw[arrow, shorten >=5pt, shorten <=7pt] (m-2-4.center) -- (m-3-4.center);
\draw[arrow, shorten >=12pt, shorten <=12pt] (m-2-3.center) -- (m-3-4.center);

\draw[arrow, gray, dashed, shorten >=5pt, shorten <=7pt] (m-1-2.center) -- (m-2-2.center);
\draw[arrow, gray, dashed, shorten >=12pt, shorten <=12pt] (m-1-1.center) -- (m-2-2.center);

\draw[arrow, gray, dashed, shorten >=5pt, shorten <=7pt] (m-1-3.center) -- (m-2-3.center);
\draw[arrow, gray, dashed, shorten >=12pt, shorten <=12pt] (m-1-2.center) -- (m-2-3.center);

\draw[arrow, gray, dashed, shorten >=5pt, shorten <=7pt] (m-1-4.center) -- (m-2-4.center);
\draw[arrow, gray, dashed, shorten >=12pt, shorten <=12pt] (m-1-3.center) -- (m-2-4.center);

\end{tikzpicture}}
\hspace{0.2cm}
\scalebox{0.99}{\begin{tikzpicture}[
    cell/.style={draw, minimum width=1.4cm, minimum height=0.8cm, anchor=center},
    textcell/.style={minimum width=1cm, minimum height=1cm, anchor=center, align=left},
    arrow/.style={blue, thick, ->, >=Latex}
]

\matrix (m) [matrix of nodes, nodes={cell}, column sep=-\pgflinewidth, row sep=-\pgflinewidth] {
  0 & $-\infty$ & $-\infty$ & $-\infty$ \\
  0 & 3 & $-\infty$ & $-\infty$ \\
  0 & 3 & 2 & $-\infty$ \\
  0 & 4 & 7 & 6 \\
  0 & 4 & 7 & 6 \\
  0 & 4 & 7 & 9 \\
};

\node[textcell] at ([xshift=0.9cm]m-1-4.east) {$[k=3]$};
\node[textcell] at ([xshift=0.9cm]m-2-4.east) {(\small$\theta_1=3$)};
\node[textcell] at ([xshift=0.9cm]m-3-4.east) {(\small$\theta_2=-1$)};
\node[textcell] at ([xshift=0.9cm]m-4-4.east) {(\small$\theta_3=4$)};
\node[textcell] at ([xshift=0.9cm]m-5-4.east) {(\small$\theta_4=-2$)};
\node[textcell] at ([xshift=0.9cm]m-6-4.east) {(\small$\theta_5=2$)};

\draw[arrow, shorten >=5pt, shorten <=7pt] (m-5-2.center) -- (m-6-2.center);
\draw[arrow, shorten >=12pt, shorten <=12pt] (m-5-1.center) -- (m-6-2.center);

\draw[arrow, shorten >=5pt, shorten <=7pt] (m-5-3.center) -- (m-6-3.center);
\draw[arrow, shorten >=12pt, shorten <=12pt] (m-5-2.center) -- (m-6-3.center);

\draw[arrow, shorten >=5pt, shorten <=7pt] (m-5-4.center) -- (m-6-4.center);
\draw[arrow, shorten >=12pt, shorten <=12pt] (m-5-3.center) -- (m-6-4.center);

\draw[arrow, gray, dashed, shorten >=5pt, shorten <=7pt] (m-1-2.center) -- (m-2-2.center);
\draw[arrow, gray, dashed, shorten >=12pt, shorten <=12pt] (m-1-1.center) -- (m-2-2.center);

\draw[arrow, gray, dashed, shorten >=5pt, shorten <=7pt] (m-1-3.center) -- (m-2-3.center);
\draw[arrow, gray, dashed, shorten >=12pt, shorten <=12pt] (m-1-2.center) -- (m-2-3.center);

\draw[arrow, gray, dashed, shorten >=5pt, shorten <=7pt] (m-1-4.center) -- (m-2-4.center);
\draw[arrow, gray, dashed, shorten >=12pt, shorten <=12pt] (m-1-3.center) -- (m-2-4.center);

\draw[arrow, gray, dashed, shorten >=5pt, shorten <=7pt] (m-2-2.center) -- (m-3-2.center);
\draw[arrow, gray, dashed, shorten >=12pt, shorten <=12pt] (m-2-1.center) -- (m-3-2.center);

\draw[arrow, gray, dashed, shorten >=5pt, shorten <=7pt] (m-2-3.center) --  (m-3-3.center);
\draw[arrow, gray, dashed, shorten >=12pt, shorten <=12pt] (m-2-2.center) -- (m-3-3.center);

\draw[arrow, gray, dashed, shorten >=5pt, shorten <=7pt] (m-2-4.center) -- (m-3-4.center);
\draw[arrow, gray, dashed, shorten >=12pt, shorten <=12pt] (m-2-3.center) -- (m-3-4.center);

\draw[arrow, gray, dashed, shorten >=5pt, shorten <=7pt] (m-3-2.center) -- (m-4-2.center);
\draw[arrow, gray, dashed, shorten >=12pt, shorten <=12pt] (m-3-1.center) -- (m-4-2.center);

\draw[arrow, gray, dashed, shorten >=5pt, shorten <=7pt] (m-3-3.center) --  (m-4-3.center);
\draw[arrow, gray, dashed, shorten >=12pt, shorten <=12pt] (m-3-2.center) -- (m-4-3.center);

\draw[arrow, gray, dashed, shorten >=5pt, shorten <=7pt] (m-3-4.center) -- (m-4-4.center);
\draw[arrow, gray, dashed, shorten >=12pt, shorten <=12pt] (m-3-3.center) -- (m-4-4.center);

\draw[arrow, gray, dashed, shorten >=5pt, shorten <=7pt] (m-4-2.center) -- (m-5-2.center);
\draw[arrow, gray, dashed, shorten >=12pt, shorten <=12pt] (m-4-1.center) -- (m-5-2.center);

\draw[arrow, gray, dashed, shorten >=5pt, shorten <=7pt] (m-4-3.center) --  (m-5-3.center);
\draw[arrow, gray, dashed, shorten >=12pt, shorten <=12pt] (m-4-2.center) -- (m-5-3.center);

\draw[arrow, gray, dashed, shorten >=5pt, shorten <=7pt] (m-4-4.center) -- (m-5-4.center);
\draw[arrow, gray, dashed, shorten >=12pt, shorten <=12pt] (m-4-3.center) -- (m-5-4.center);

\end{tikzpicture}}
\caption{Illustration of the DP recursion table for the Top-$k$ problem, with $\thetav=(3,-1,4,-2,2)^\top$ and $k=3$. Arrows indicate the influence of values on others, illustrating the DAG structure underlying the recursion. See \cref{sec:dag_ddp} for more details.}
\label{fig:topk_dp_tables}
\end{figure}

\subsection{Visualization of \cref{prop:equivariance,prop:sparsity}}

\begin{figure}[H]
    \centering
    \includegraphics[width=0.99\linewidth]{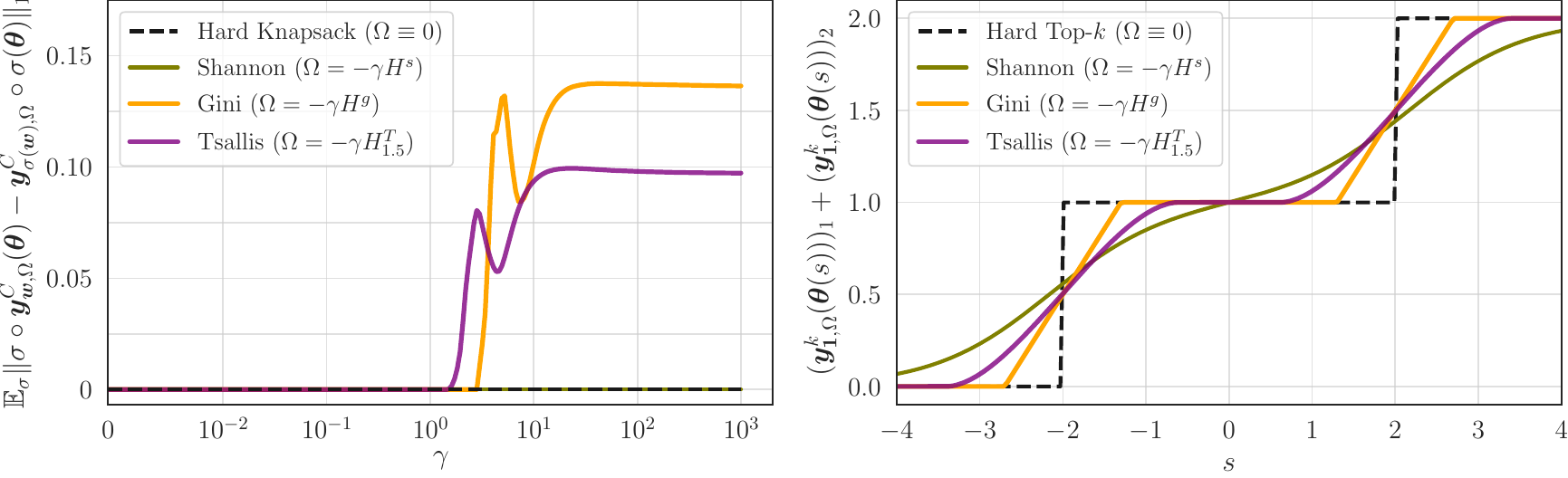}
    \caption{\textbf{Illustration of \cref{prop:equivariance,prop:sparsity}}. On the \textbf{left}, we measure the $\ell_1$ distance between $\sigma\circ\ycwo(\thetav)$ and $\y^C_{\sigma(\w),\Omega}\circ\sigma(\thetav)$, averaged on every possible permutation $\sigma\in S_n$. We use $\thetav=(1,2,3,4,5,6)^\top$, $\w=(6,5,4,3,2,1)^\top$, and $C=10$, for varying regularization weight $\gamma$. The Shannon entropy is the only one yielding a permutation-equivariant operator with $\gamma>0$.\newline On the \textbf{right}, we use $\gamma=0.7$, $\w=\1$ and $k=2$, and define $\thetav(s)\triangleq (s-1, s, 1, -2)^\top$ for $s\in[-4,4]$. We plot the sum of the first two components of the relaxed operator $\yko(\thetav(s))$. The hard Top-$k$ mask ($\Omega\equiv 0$) is piecewise-constant. Gini and Tsallis regularization yield a sparse and differentiable (\textit{a.e.} for Gini) operator $\yko$, while Shannon regularization yields a dense and differentiable $\yko$.}
    \label{fig:equivariance_and_sparsity}
\end{figure}

\subsection{Mapping with the Differentiable Dynamic Programming framework}
\label{sec:dag_ddp}

In this section, we formally cast our proposed smoothed recursions as instances of the differentiable dynamic programming (DDP) framework introduced by \citet{mensch_differentiable_2018}.

\paragraph{DDP framework.} The DDP framework considers optimization problems framed as finding a highest-scoring path on a Directed Acyclic Graph (DAG) $\cG=(\cV,\cE)$. The edge scores are parameterized by $\Thetav\in\RR^{|\cV|\times|\cV|}$, where $\Theta_{u,v}$ is the score of edge $(u,v)\in\cE$. The value of a node $v$, denoted $V_v(\Thetav)$, is defined recursively by the Bellman equation:
\begin{align*}
    V_v(\Thetav) \triangleq \max_{u\in\cP_v} \left( \Theta_{u,v} + V_u(\Thetav) \right),
\end{align*}
where $\cP_v$ denotes the set of parents of node $v$. The smoothed recursion is obtained by replacing the $\max$ operator with $\max_\Omega$.

\paragraph{Knapsack DAG construction.}
To map the Knapsack problem defined by weights $\w$ and capacity $C$ to this framework, we construct a DAG where nodes represent the states of the dynamic program.
\begin{itemize}
    \item \textbf{Nodes ($\cV$).} The set of nodes is the grid of DP states $\cV = \{ (i,c) \mid 0 \leq i \leq n, \, 0 \leq c \leq C \}$. The source node is $(0,0)$ and the sink node is $(n,C)$.
    
    \item \textbf{Edges ($\cE$) and Scores ($\Thetav$).} The transitions from row $i-1$ to row $i$ correspond to the decision of selecting item $i$ or not. For each node $(i,c)$ with $i \in \{1,\dots,n\}$ and $c \in \{0,\dots,C\}$:
    \begin{enumerate}
        \item \textbf{Skip item $i$:} An edge exists from $(i-1, c)$ to $(i,c)$ representing the decision $y_i=0$. The score is $\Theta_{(i-1,c), (i,c)} = 0$.
        \item \textbf{Pick item $i$:} If $c \geq w_i$, an edge exists from $(i-1, c-w_i)$ to $(i,c)$ representing the decision $y_i=1$. The score is $\Theta_{(i-1, c-w_i), (i,c)} = \theta_i$.
    \end{enumerate}
    This local construction is illustrated in \cref{fig:knapsack_dag_construction}.
    \item \textbf{Boundary conditions.} As explained in \cref{sec:dynamic}, boundary conditions of the form $V_{(0,c)}(\Thetav) = 0$ for all $c\in\{0,\dots,C\}$ are imposed on the first row to initialize the recursion (note that whenever $w_i \geq 1$ for all $i\in[n]$, only the skip edge exists at $c=0$, which automatically propagates $V_{(i,0)}(\Thetav) = 0$ down the first column as well).
    
    In order to formally cast this as an instance of DDP (where only the source node $(0,0)$ is assigned a fixed value of $0$), we add artificial edges between the source node $(0,0)$ and every boundary node $(0,c)$ for $c\in\{1,\dots,C\}$, with edge scores $\Thetav_{(0,0),(0,c)}=0$. 
\end{itemize}

Under this construction, the DDP recursion exactly matches the Knapsack recursion. Indeed, for $i\in\{1,\dots,n\}$, $c\in\{0,\dots,C\}$ (and $c'\in\{1,\dots,C\}$ on row $0$), we have:
\begin{align*}
    V_{(i,c)}(\Thetav) &= \max\left( \underbrace{\underbrace{0}_{\Thetav_{(i-1,c),(i,c)}} + V_{(i-1,c)}(\Thetav)}_{\text{Skip}}, \underbrace{\underbrace{\theta_i}_{\Thetav_{(i-1,c-w_i),(i,c)}} + V_{(i-1, c-w_i)}(\Thetav)}_{\text{Pick (if $c\geq w_i$)}} \right),\\
    V_{(0,c')}(\Thetav) &= \max\left(\underbrace{0}_{\Thetav_{(0,0),(0,c')}} + V_{(0,0)}(\Thetav)\right) = V_{(0,0)}(\Thetav) = 0.
\end{align*}

A full example of the recursion with corresponding edges is drawn in \cref{fig:knapsack_dp_tables}.

\begin{figure}[t]
\centering
\begin{tikzpicture}[
    scale=1.0,
    state/.style={circle, draw=black, minimum size=0.8cm, inner sep=0pt},
    >={Latex[length=2mm, width=2mm]}
]

\node[state] (prev_c) at (0, 2) {$\scriptstyle i\!-\!1, c$};
\node[state] (prev_cw) at (-3, 2) {$\scriptstyle i\!-\!1, c\!-\!w_i$};

\node[state] (curr_c) at (0, 0) {$\scriptstyle i, c$};

\draw[->, thick] (prev_c) -- (curr_c) node[midway, right, font=\small] {Skip ($0$)};

\draw[->, thick] (prev_cw) -- (curr_c) node[midway, left, font=\small] {Pick ($\theta_i$)};

\node at (-1.5, 2) {$\dots$};
\node at (-1.5, 0) {$\dots$};

\node[align=left, anchor=west] at (2, 1) {
    \textbf{Transitions to} $\bm{(i,c)}$:\\
    $\bullet$ From $(i\!-\!1, c)$: item $i$ not selected.\\
    \textcolor{gray}{Edge score: $0$}\\
    $\bullet$ From $(i\!-\!1, c\!-\!w_i)$: item $i$ selected.\\
    \textcolor{gray}{Edge score: $\theta_i$}
};

\end{tikzpicture}
\caption{Local DAG structure for the Knapsack DDP formulation. The value $V_{(i,c)}(\Thetav)$ is computed by taking the maximum (or smoothed maximum) over incoming edges.}
\label{fig:knapsack_dag_construction}
\end{figure}

\paragraph{Top-$k$ adaptation.}
The Top-$k$ problem ($\w=\1, C=k$) follows the same graph topology. As explained in \cref{sec:dynamic}, the constraint that exactly $k$ items must be chosen is enforced via the boundary conditions $V_{(0,0)}(\Thetav) = 0$ and $V_{(0,j)}(\Thetav) = -\infty$ for all $j\in\{1,\dots,k\}$ (while $w_i = 1 > 0$ ensures only the skip edge exists at $j=0$, propagating $V_{(i,0)}(\Thetav) = 0$ for all $i\in[n]$). An example showing how these boundary conditions effectively propagate and enforce the Top-$k$ constraint is depicted in \cref{fig:topk_dp_tables}.

In order to formally cast this as an instance of DDP, we add artificial edges between the source node $(0,0)$ and every boundary node $(0,j)$ for $j\in[k]$ with edge scores $\Thetav_{(0,0),(0,j)}=-\infty$.

Under this construction, the DDP recursion exactly matches the Top-$k$ recursion. Indeed, for $i\in\{1,\dots,n\},j\in\{1,\dots,k\}$, we have:
\begin{align*}
    V_{(i,j)}(\Thetav) &= \max\left( \underbrace{\underbrace{0}_{\Thetav_{(i-1,j),(i,j)}} + V_{(i-1,j)}(\Thetav)}_{\text{Skip}}, \underbrace{\underbrace{\theta_i}_{\Thetav_{(i-1,j-1),(i,j)}} + V_{(i-1, j-1)}(\Thetav)}_{\text{Pick}} \right),\\
    V_{(i,0)}(\Thetav) &= \max\left(\underbrace{0}_{\Thetav_{(i-1,0),(i,0)}} + V_{(i-1,0)}(\Thetav)\right) = V_{(i-1,0)}(\Thetav) = 0,\\
    V_{(0,j)}(\Thetav) &= \max\left(\underbrace{-\infty}_{\Thetav_{(0,0),(0,j)}} + V_{(0,0)}(\Thetav)\right) = -\infty.
\end{align*}

\subsection{Properties of the relaxed values and operators}
\label{sec:properties}

The relaxed values and operators have the following properties:

\begin{enumerate}
    \item The smoothed output values $\thetav\mapsto\maxcwo(\thetav)$ and $\thetav\mapsto\maxko(\thetav)$ are convex in $\thetav$.
    \item Define $\Omega_d:\triangle^{d}\to\RR$ as $\Omega_d(\q)\triangleq \sum_{i=1}^d\omega(q_i)$ for any integer $d$. Then, the difference between the smoothed and the unregularized values are bounded above and below, for all $\thetav\in\RR^n$:
    \begin{align*}
        (N_C-1)L_{N_C}&\leq \maxcw(\thetav)-\mathsf{max}^C_{\w,\Omega_2}(\thetav)\leq (N_C-1)U_{N_C},\\
        (N_k-1)L_{N_k}&\leq \maxk(\thetav)-\mathsf{max}^k_{\bm{1},\Omega_2}(\thetav)\leq (N_k-1)U_{N_k},
    \end{align*}
    where $L_d$ and $U_d$ are lower and upper bounds of $\Omega_d$ on $\triangle^d$, and with $N_C\triangleq (C+1)(n+1)$,   $N_k\triangleq(k+1)(n+1)$.
    \item Scaling the regularizer by $\gamma>0$ acts as a temperature parameter on the input scores $\thetav\in\RR^n$:
    \begin{align*}
        \mathsf{max}^C_{\w,\gamma\Omega}(\thetav) &= \gamma\,\maxcwo(\thetav/\gamma), \qquad &\y_{\w,\,\gamma\Omega}^C(\thetav) &= \ycwo(\thetav/\gamma),\\
        \mathsf{max}^k_{\bm{1},\gamma\Omega}(\thetav) &= \gamma\,\maxko(\thetav/\gamma), \qquad &\y_{\bm{1},\,\gamma\Omega}^k(\thetav) &= \yko(\thetav/\gamma).
    \end{align*}
    \item The proposed layers converge to the original, hard operators for vanishing regularization:
    \begin{align*}
        \y_{\w,\,\gamma\Omega}^C(\thetav)\xrightarrow[\gamma\to0^+]{}\ycw(\thetav).
    \end{align*}
    \item For a fixed $\gamma>0$ and Shannon regularization $\Omega=-H^{s}$, we can bound the bias induced by regularization on the forward pass of the operators in $\ell_1$ distance as:
    \begin{align*}
        \|\y_{\w,\,\gamma\Omega}^C(\thetav)-\ycw(\thetav)\|_1\leq n|\Ycw|\exp(-\delta/\gamma),
    \end{align*}
    where $\delta\triangleq\max_{\y\in\Ycw}\langle\thetav,\y\rangle-\max_{\y'\in\Ycw\setminus\{\ycw(\thetav)\}}\langle\thetav,\y'\rangle$ is the hard optimality gap between the best and second best solutions (note that we have $\delta>0$ almost everywhere with respect to $\thetav\in\RR^n$). Thus, this forward bias decays exponentially with the regularization strength $\gamma$.
\end{enumerate}

\begin{proof}
    \textbf{Properties 1, 2, and 4.} Properties 1, 2, and 4 stem from a direct application of results in \citet[Proposition 2]{mensch_differentiable_2018} to our special DAG and edge weights, after mapping our framework to the DDP framework as described in \cref{sec:dag_ddp}. The constants $N_C$ and $N_k$ emerge from the size of the corresponding DAG.

    \textbf{Property 3.} For any $\gamma > 0$ and $\x \in \RR^2$, we have:
    \begin{align*}
        \mathsf{max}_{\gamma\Omega}(\x) = \max_{\q \in \triangle^2}\{\langle \x, \q \rangle - \gamma\Omega(\q)\} = \gamma \max_{\q \in \triangle^2}\left\{\left\langle\frac{\x}{\gamma},\q\right\rangle-\Omega(\q)\right\} = \gamma\,\maxo(\x/\gamma).
    \end{align*}
    Since the DP boundary values ($0$ and $-\infty$) are invariant under multiplication by $\gamma > 0$, induction on $i$ in \cref{eq:smoothed_recursion} shows that evaluating the $\gamma\Omega$-smoothed DP table at $\thetav$ multiplies every entry of the $\Omega$-smoothed DP table at $\thetav/\gamma$ by $\gamma$, yielding the equalities $\mathsf{max}^C_{\w,\gamma\Omega}(\thetav) = \gamma\,\maxcwo(\thetav/\gamma)$ and $\mathsf{max}^k_{\bm{1},\gamma\Omega}(\thetav) = \gamma\,\maxko(\thetav/\gamma)$. Differentiating both sides with respect to $\thetav$ via the chain rule further gives $\y_{\w,\,\gamma\Omega}^C(\thetav) = \ycwo(\thetav/\gamma)$ and $\y_{\bm{1},\,\gamma\Omega}^k(\thetav) = \yko(\thetav/\gamma)$.

    \textbf{Property 5.} For the fifth property, we can write:
    \begin{align*}
        \y_{\w,\,\gamma\Omega}^C(\thetav)-\ycw(\thetav)&=\sum_{\y\in\Ycw}\frac{\exp(\langle\thetav,\y\rangle/\gamma)}{\sum_{\y'\in\Ycw}\exp(\langle\thetav,\y'\rangle/\gamma)}\y - \ycw(\thetav)\\
        &= \sum_{\y\in\Ycw\setminus\{\ycw(\thetav)\}}\frac{\exp(\langle\thetav,\y\rangle/\gamma)}{\sum_{\y'\in\Ycw}\exp(\langle\thetav,\y'\rangle/\gamma)}(\y - \ycw(\thetav))
    \end{align*}
    and then use the triangular inequality to get:
    \begin{align*}
        \|\y_{\w,\,\gamma\Omega}^C(\thetav)-\ycw(\thetav)\|_1&\leq\sum_{\y\in\Ycw\setminus\{\ycw(\thetav)\}}\|\y - \ycw(\thetav)\|_1\frac{\exp(\langle\thetav,\y\rangle/\gamma)}{\sum_{\y'\in\Ycw}\exp(\langle\thetav,\y'\rangle/\gamma)}\\
        &\leq \sum_{\y\in\Ycw\setminus\{\ycw(\thetav)\}}\|\y - \ycw(\thetav)\|_1\frac{\exp(\max_{\y''\in\Ycw\setminus\{\ycw(\thetav)\}}\langle\thetav,\y''\rangle/\gamma)}{\exp(\max_{\y'\in\Ycw}\langle\thetav,\y'\rangle/\gamma)}\\
        &\leq n|\Ycw|\exp(-\delta/\gamma). \qedhere
    \end{align*}
\end{proof}

\subsection{Properties of the surrogate gradients for stochastic forward passes}
\label{sec:surrogate_properties}

\paragraph{Unbiased first-order estimator.}
Let $\cL:\conv(\Ycw)\to\RR$ denote the composition of the downstream computation graph $f_{W_2}$ with any loss function. The sampling operation $\thetav\mapsto \y\sim\picwo$ performed in the stochastic forward pass of \cref{sec:stochastic_forward} is inherently non-differentiable. Thus, there is no ``exact gradient'' to compare against. However, following \citet{liu_bridging_2023}, we can show that our surrogate gradient is still principled as it is an \emph{unbiased first-order estimator} of the gradient of the expectation of the loss, $J(\thetav)\triangleq \EE_{Y\sim\picwo}[\cL(Y)]$.

Let $\hat{\nabla}_{\text{surr}}\triangleq (\nabla_\thetav\ycwo(\thetav))\cdot\nabla\cL(Y)$ denote the proposed surrogate gradient, where $Y\sim\picwo$. First, notice that we have $\sum_{\y \in \Ycw}\nabla_{\thetav}\picwo(\y) = \nabla_{\thetav}\sum_{\y \in \Ycw}\picwo(\y)  = \nabla_{\thetav}(1)=0$. Thus, we have:
$$ \nabla J(\thetav) = \sum_{\y \in \Ycw} \cL(\y) \nabla_{\thetav}\picwo(\y) = \sum_{\y \in \Ycw} \left( \cL(\y) - J(\thetav) \right) \nabla_{\thetav} \picwo(\y).$$

By expanding the expected loss $J(\thetav) = \sum_{\y' \in \Ycw} \picwo(\y') \cL(\y')$, we obtain:

$$ \nabla J(\thetav) = \sum_{\y \in \Ycw} \sum_{\y' \in \Ycw} \picwo(\y') \left( \cL(\y) - \cL(\y') \right) \nabla_{\thetav} \picwo(\y) $$

At this point, we define a first-order approximation of the true gradient, $\hat{\nabla}_{\text{1st-order}}$, by replacing the exact loss difference $\cL(\y) - \cL(\y')$ with its first-order Taylor expansion $\nabla \cL(\y')^\top (\y - \y')$:

$$ \hat{\nabla}_{\text{1st-order}} \triangleq \sum_{\y \in \Ycw} \sum_{\y' \in \Ycw} \picwo(\y') \left( \nabla \cL(\y')^\top (\y - \y') \right) \nabla_{\thetav} \picwo(\y) $$

To show that our surrogate gradient estimator $\hat{\nabla}_{\text{surr}}$ is an unbiased estimator of this first-order approximation, we rearrange the terms in $\hat{\nabla}_{\text{1st-order}}$ to isolate the sum over $\y$:

$$ \hat{\nabla}_{\text{1st-order}} = \sum_{\y' \in \Ycw} \picwo(\y') \sum_{\y \in \Ycw} \nabla_{\thetav} \picwo(\y) \left( (\y - \y')^\top \nabla \cL(\y') \right) $$

Distributing the dot product allows us to split the inner summation into two separate terms:

$$ \hat{\nabla}_{\text{1st-order}} \!=\!\!\! \sum_{\y' \in \Ycw} \!\!\picwo(\y') \left[ \left( \sum_{\y \in \Ycw} \!\!\nabla_{\thetav} \picwo(\y) \y^\top \right) \nabla \cL(\y')\! -\! \left( \sum_{\y \in \Ycw} \!\!\nabla_{\thetav} \picwo(\y) \right) \y'^\top \nabla \cL(\y') \right] $$

Recall that the sum of the gradients of the probabilities is zero ($\sum_{\y \in \Ycw} \nabla_{\thetav} \picwo(\y) = 0$). This eliminates the entire second term inside the bracket, simplifying the expression to:

$$ \hat{\nabla}_{\text{1st-order}} = \sum_{\y' \in \Ycw} \picwo(\y') \left( \sum_{\y \in \Ycw} \nabla_{\thetav} \picwo(\y) \y^\top \right) \nabla \cL(\y') $$

Next, we recognize that the expected value of the sample under the distribution is $\ycwo(\thetav) = \sum_{\y \in \Ycw} \picwo(\y) \y$. Its Jacobian with respect to $\thetav$ corresponds exactly to the inner summation:

$$ \nabla_{\thetav} \ycwo(\thetav) = \sum_{\y \in \Ycw} \nabla_{\thetav} \picwo(\y) \y^\top $$

Substituting this Jacobian back into our equation yields:

$$ \hat{\nabla}_{\text{1st-order}} = \sum_{\y' \in \Ycw} \picwo(\y') \left( \nabla_{\thetav} \ycwo(\thetav) \right) \cdot \nabla \cL(\y') $$

Finally, rewriting this sum over $\y'$ weighted by $\picwo(\y')$ as an expectation directly recovers the expected value of our surrogate gradient:

$$ \hat{\nabla}_{\text{1st-order}} = \EE_{Y \sim \picwo} \left[ (\nabla_{\thetav} \ycwo(\thetav)) \cdot \nabla \cL(Y) \right] = \EE_{Y \sim \picwo} \left[ \hat{\nabla}_{\text{surr}} \right].$$

This concludes the proof that our surrogate gradient $\hat{\nabla}_{\text{surr}}$ is a principled, unbiased first-order estimate of $\nabla J(\thetav)$.

\begin{figure}
    \centering
    \includegraphics[width=0.5\linewidth]{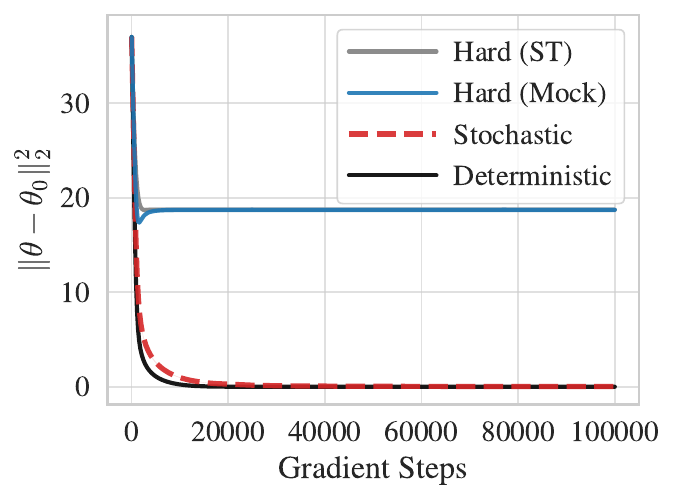}
    \caption{Comparison of our stochastic forward pass and surrogate gradient against a hard forward pass with straight-through backward, a hard forward pass with relaxed VJP backward (mock gradient), and the relaxed deterministic forward with relaxed VJP backward (deterministic layer from \cref{sec:continuous_forward}).}
    \label{fig:mock_gradient}
\end{figure}

\paragraph{Comparison to straight-through and mock gradient approaches.} In order to further validate our proposed surrogate gradient estimator, we compare it to two alternatives for gradient propagation when using hard item selections in the forward passes:
\begin{enumerate}
    \item a hard forward pass $\thetav\mapsto\ycw(\thetav)$ with a straight-through estimator in the backward (treating the layer as if it were the identity),
    \item a hard forward pass $\thetav\mapsto\ycw(\thetav)$ using the relaxed layer's VJP $\nabla_\thetav\ycwo(\thetav)$ in the backward (a ``mock gradient'' approach).
\end{enumerate}
We evaluate the layers on the following simple task: we generate a hidden, ground-truth parameter $\thetav_0\in\RR^n$ and initialize a learnable one $\thetav$. Then, we minimize the MSE between $\hat{\y}(\thetav)$ (where $\hat{\y}$ is the evaluated layer) and $\ycwo(\thetav_0)$ using gradient descent (where the gradient computation is custom and defined differently for each layer as described above), and record the squared $\ell_2$ distance between $\thetav$ and $\thetav_0$ along the optimization trajectory.

The results are gathered in \cref{fig:mock_gradient}. While the alternatives using hard forward passes exhibit a strong bias and do not converge to the true $\thetav_0$, the proposed stochastic forward pass from \cref{sec:stochastic_forward} almost matches the exact, deterministic forward pass from \cref{sec:continuous_forward}. 

\section{Additional Details on Numerical Experiments}

\subsection{Experimental details for \cref{sec:experiments_output}}
\label{sec:pyepo_appendix}

\paragraph{Relative regret metric.} For a data point $(\x^{(i)},\thetav^{(i)},\y^{(i)})$, the relative regret of a predicted solution $\ycw(\hat{\thetav}^{(i)})$ is defined as the relative difference between its objective value and the optimal one under the true item values $\thetav^{(i)}$, i.e.:
\begin{align*}
    \frac{\langle \thetav^{(i)}, \y^{(i)}\rangle - \langle \thetav^{(i)}, \ycw(\hat{\thetav}^{(i)})\rangle}{|\langle \thetav^{(i)}, \y^{(i)}\rangle |}.
\end{align*}

\paragraph{Data generation.}
We generate synthetic Knapsack instances using the \texttt{genData} protocol from the \texttt{PyEPO} library. 
For each instance, item weights $\w \in \NN^n$ are sampled uniformly from the interval $\{3,\dots, 8\}$.
The item values $\thetav$ are generated dependent on input features $\x \in \RR^p$ (with $\x \sim \mathcal{N}(0, 1)$) according to a polynomial relation with degree $\text{deg}=3$:
\begin{align*}
    \theta_{j} = \left\lceil \left( \left( \frac{1}{\sqrt{p}} (\B\x)_j + 3 \right)^{3} + 1 \right) \cdot 5 \cdot 3.5^{-3} \cdot \epsilon_j \right\rceil,
\end{align*}
where $\B_{ij} \sim \text{Bernoulli}(0.5)$ and $\epsilon_j \sim \mathcal{U}(0.7, 1.3)$ is a multiplicative noise term. 
The Knapsack capacity $C$ is set to $50\%$ of the total weight of all items: $C\triangleq\lfloor \frac{1}{2}\sum_{i=1}^n w_i\rfloor$.
We generate a total of $18,000$ samples, split into $10,000$ for training, $4,000$ for validation, and $4,000$ for testing.

\paragraph{Architecture and optimization.}
We utilize a standard feed-forward neural network with two hidden layers of dimension $64$ and ReLU activations. 
All models are trained for $30$ epochs using the Adam optimizer with a learning rate of $2 \times 10^{-3}$ and a batch size of $32$. All experiments were executed on the CPU of an Apple M1 Max processor with 64 GB of RAM.

\paragraph{Statistical significance.}
We quantify the computational overhead by measuring the wall-clock time required for a single gradient step (encompassing both forward and backward passes), averaged over $10^4$ repetitions to mitigate variance. We monitor performance of the methods on distinct datasets generated with $n \in \{10, 25, 50, 100\}$, each generated with the protocol described earlier. Each method is used to train a model on the training set, then the best model iteration is selected according to its relative regret on the validation set, and performance is finally reported using the relative regret on the test set. To ensure the robustness of our results, we perform a systematic evaluation across $10$ independent random seeds for every method.

\paragraph{Baselines details.} We compare against the following methods provided in the \texttt{PyEPO} benchmark:
\begin{itemize}
    \item \texttt{PFY}: The \textit{perturbed Fenchel-Young} loss \citep{berthet_learning_2020}, which relies on Monte-Carlo sampling of perturbed costs to estimate gradients.
    \item \texttt{DBB}: The \textit{differentiable black-box} \citep{pogancic_differentiation_2020} method, which computes gradients via piecewise affine interpolation.
    \item \texttt{NCE}: A surrogate loss based on \emph{noise-contrastive estimation} \citep{mulamba_contrastive_2021}.
    \item \texttt{NID}: Treats the optimal solution mapping of combinatorial minimization problems as a \emph{negative identity} function during the backward pass \citep{sahoo_backpropagation_2022}.
\end{itemize}
In order to enable a fair comparison, we use the same Numba DP-based implementation of the unregularized knapsack operator $\ycw$ as a solver for every baseline.

For \texttt{DBB} and \texttt{NID}, which are differentiation methods and not loss functions, we minimize the mean squared error on solutions $\frac{1}{2n}\|\ycw(\hat{\thetav}^{(i)})-\y^{(i)}\|_2^2$.

\paragraph{Hyperparameters.}
For the baselines and our method, we use the following hyperparameters:
\begin{itemize}[noitemsep, topsep=0pt]
    \item DP-based Fenchel-Young losses (ours): We use a regularization strength $\gamma=5.0$.
    \item \texttt{PFY:} We use $10$ Monte-Carlo samples and a perturbation noise $\sigma=5.0$.
    \item \texttt{DBB:} We use a smoothing parameter $\lambda=0.1$.
\end{itemize}

These hyperparameters were tuned using a grid search on the validation set.

\subsection{Derivations and experimental details for \cref{sec:experiments_dvae}}
\label{sec:dvae_appendix}

\paragraph{Derivation of the FY-DVAE objective.}
Standard DVAEs maximize the ELBO:
\begin{align*}
    \log p_{W_2}(\x) \geq \EE_{Y\sim q_{W_1}(\cdot\mid \x)}[\log p_{W_2}(\x\mid Y)] - D_{\text{KL}}(q_{W_1}(\cdot\mid \x) \mid\mid p(\cdot)).
\end{align*}
In the standard DVAE setting of, e.g., \citet{ahmed_simple_2023}, where the approximate posterior is parameterized as a Gibbs distribution $q_{W_1}(\y|\x) = \pi_{\thetav}(\y) \propto \exp(\langle \thetav, \y \rangle)$ with $\thetav=E_{W_1}(\x)$, and the prior $p$ is the uniform distribution on $k$-subsets, the KL divergence regularization term writes:
\begin{align*}
    D_{\text{KL}}(\pi_\thetav \mid\mid p) &\triangleq \sum_{\y\in\Yk}\pi_\thetav(\y)\log\left(\frac{\pi_\thetav(\y)}{p(\y)}\right)\\
    &= - \sum_{\y\in\Yk}\pi_\thetav(\y)\log\left(\frac{1}{|\Yk|}\right) + \sum_{\y\in\Yk}\pi_\thetav(\y)\log\left(\pi_\thetav(\y)\right) \\
    &= \log\left(|\Yk|\right) + \sum_{\y\in\Yk}\frac{\exp(\langle\thetav,\y\rangle)}{\sum_{\y'\in\Yk}\exp(\langle\thetav,\y'\rangle)}\left(\langle\thetav,\y\rangle - \log\sum_{\y'\in\Yk}\exp(\langle\thetav,\y'\rangle)\right)\\
    &= \log\left(|\Yk|\right) - \log\sum_{\y'\in\Yk}\exp(\langle\thetav,\y'\rangle) + \left\langle \sum_{\y\in\Yk}\frac{\exp(\langle\thetav,\y\rangle)}{\sum_{\y'\in\Yk}\exp(\langle\thetav,\y'\rangle)}\y,\;\thetav\right\rangle\\
    &= A(\0) - A(\thetav) + \langle\nabla A(\thetav) ,\thetav\rangle,
\end{align*}
where $A(\thetav) = \log \sum_{\y\in\Yk} \exp(\langle \thetav, \y \rangle)$ is the cumulant function of $\pi_\thetav$. We recognize this expression as a Fenchel-Young loss generated by the convex conjugate of the cumulant, $A^*$. Indeed, we have:
\begin{align*}
    L_{A^*}(\0\,; \nabla A(\thetav)) &\triangleq A^{**}(\0) + A^*(\nabla A(\thetav)) - \langle \0, \nabla A(\thetav) \rangle \quad \text{\citep{blondel_learning_2020}} \\
    &= A(\0) + (\langle \thetav, \nabla A(\thetav) \rangle - A(\thetav)) \quad \text{($A^{**}=A$ and Fenchel's equality)} \\
    &= A(\0) - A(\thetav) + \langle \thetav, \nabla A(\thetav) \rangle \\
    &= D_{\text{KL}}(\pi_\thetav \mid\mid p).
\end{align*}
Here, we used the fact that $A$ is convex and lower semi-continuous to apply the Fenchel-Moreau biconjugation theorem, and Fenchel's equality $A^*(\nabla A(\thetav)) + A(\thetav) = \langle \thetav, \nabla A(\thetav) \rangle$.

We generalize this to our smoothed DP setting by using a generalized approximate posterior $q_{W_1}(\y|\x)=\piko(\y)$, and by replacing $A^*$ with the generalized moment polytope regularization function $\Omegak$, defined in \cref{sec:output_layer} as the conjugate of the smoothed value $\maxko$. The resulting regularization term is the Fenchel-Young loss between the uniform prior parameters $\0$ and the relaxed item selection $\yko(\thetav)$:
\begin{align*}
    L_{\Omegak}(\0\,; \yko(\thetav)) &\triangleq (\Omegak)^*(\0)  + \Omegak(\yko(\thetav)) - \langle \0, \yko(\thetav) \rangle \\
    &= (\langle \thetav, \yko(\thetav) \rangle - \maxko(\thetav)) + \maxko(\0) \\
    &= \maxko(\0) - \maxko(\thetav) + \langle \thetav, \yko(\thetav) \rangle.
\end{align*}
By construction, as $\Omega=-H^{s}$ yields $\maxko=A$ and $\piko=\pi_\thetav$ (see \cref{sec:closed_form,prop:distribution}), this formulation recovers the standard KL divergence objective used in \citet{ahmed_simple_2023} as a special case, while enabling the use of regularizers that incentivize sparse latent representations (e.g., Gini or $1.5$-Tsallis). This framework also extends the Fenchel-Young variational inference framework of \citet{sklaviadis_fenchelyoung_2025}, which introduced Fenchel-Young regularization for learning continuous VAEs, to the discrete setting of latent distributions on $k$-subsets. Moreover, the gradient of this Fenchel-Young regularization term is given by:
\begin{align*}
    \nabla_\thetav L_{\Omegak}(\0\,; \yko(\thetav)) = (\nabla_\thetav\yko(\thetav))\cdot\thetav,
\end{align*}
that is, it is simply the JVP of the relaxed layer $\yko$ in the direction of its input $\thetav$, efficiently computed via \cref{algo:vjp}.

\paragraph{Dataset generation.}
We construct each data point in our Stacked MNIST dataset by sampling $k=3$ distinct digit classes $\{c_1, \dots, c_k\} \subset \{0, \dots, 9\}$ without replacement. For each class $c_j$, we sample a random image $\x^{(j)}$ from the standard MNIST dataset. The input $\x$ is the pixel-wise average: $\x = \frac{1}{k} \sum_{j=1}^k \x^{(j)}$. The ground truth is the $k$-hot vector $\y \in \{0,1\}^{10}$ indicating the present digits. We generate $120,000$ training examples and $20,000$ test examples.

\paragraph{Architecture.}
The model architecture captures both the presence and the style of the digits using a decomposed latent space with dimensions $n=10$ for the discrete part and $d_z=32$ for the continuous part.
The encoder $E_{W_1}$ is a convolutional neural network consisting of a stack of strided convolution layers (kernel size $3\times3$, stride $2$, ReLU activation), which progressively downsample the spatial dimensions while increasing channel depth. The resulting flattened feature map is fed into two separate dense projection heads: one outputting the selection logits $\thetav\in \RR^{10}$ for the discrete mask $\y$, and another outputting the parameters $[\muv, \log\sigmav^2] \in \RR^{10 \times 2d_z}$ for the $n=10$ independent Gaussian style posteriors $\z_i \sim \cN(\muv_i, \text{diag}(\sigmav_i^2))$.

The mapping from selection logits $\thetav$ to mixing coefficients $\y$ in the discrete bottleneck depends on the considered method:
\begin{itemize}
    \item for the hard Top-$k$ baseline, the mapping is simply $\y=\yk(\thetav)\in\Yk$,
    \item for the Gumbel Top-$k$ baseline, we sample i.i.d. Gumbel noise $Z_i\sim \text{Gumbel}(0,1)$ for $i\in[10]$ and take the Top-$k$ of the perturbed values $\y=\yk(\thetav + \tau \Z)\in\Yk$,
    \item for the specialized deterministic Top-$k$ baselines, we apply the operators of \citet{sander_fast_2023} (FDS), \citet{struski_lapsum_2025} (LapSum), and \citet{xie_differentiable_2020} (SOFT, using optimal transport regularization strength $\epsilon=0.005$ selected via a hyperparameter sweep),
    \item for our deterministic DP-based layers, we directly use the regularized Top-$k$ operator $\y=\yko(\thetav)\in\conv(\Yk)$ using \cref{algo:layer},
    \item and for our stochastic DP-based layers, we sample from the underlying distribution via $\y\sim\piko$ using \cref{algo:sample}.
\end{itemize}

We maintain a set of learned embeddings $\mathbf{E} = \{\e_1, \dots, \e_{10}\}$ with $\e_i \in \RR^{d_z}$, representing the prototypes for each digit class.
The decoder $D_{W_2}$ is a \emph{shared} de-convolutional network applied independently to each expert. For each item $i$, it receives the concatenation $[\z_i, \e_i] \in \RR^{2 d_z}$ of the sampled style code and the prototype embedding. This input is mapped via a dense layer to a low-resolution feature map, then upsampled via a stack of transposed convolution layers (kernel size $3\times3$, stride $2$, ReLU activation) to reach the target image resolution. A final convolution layer followed by a sigmoid activation produces the candidate reconstruction $\hat{\x}_i \in [0,1]^{H \times W}$. The final output is the weighted average of these candidates:
\begin{align*}
    \hat{\x} = \frac{1}{k} \sum_{i=1}^{10} y_i \cdot \sigma\left( D_{W_2}([\z_i, \e_i]) \right) = \frac{1}{k} \sum_{i=1}^{10} y_i \cdot \hat{\x}_i.
\end{align*}

\paragraph{Training details.}
The models are trained using the Adam optimizer with a learning rate of $3 \times 10^{-4}$ and a batch size of $64$ for $50$ epochs. The full training objective is given by:
\begin{align*}
    \cL(\x\,; W_1, W_2) =\; &\|\x - \hat{\x}\|_2^2 + \beta_{\text{KL}} \sum_{i=1}^{10} D_{\text{KL}}(\cN(\muv_i, \text{diag}(\sigmav_i^2)) || \cN(\0, \I)) + \beta_{\text{FY}} \cdot L_{\Omegak}(\0\,; \yko(\thetav)).
\end{align*}
We use a fixed weight $\beta_{\text{KL}}=10^{-5}$ for the continuous latent regularization.
We anneal the discrete Fenchel-Young regularization strength $\beta_{\text{FY}}$ linearly from $0$ to a base value of $10^{-5}$ over the first two-thirds of the training epochs. All displayed metrics are averaged over $30$ trainings with different random seeds. All experiments were executed on TPUv2 with one device.

\paragraph{Final test metrics.}
To complement the training curves in \cref{fig:curves_dvae} with exact precision, we report the final test reconstruction error and KL divergence for all evaluated methods in \cref{tab:dvae_final_metrics}.

\begin{table}[h]
\centering
\caption{\textbf{Final DVAE metrics} (test reconstruction error $\times 10^2$ and continuous latent KL divergence) averaged across $30$ random seeds,  sorted by reconstruction error in ascending order.}
\label{tab:dvae_final_metrics}
\begin{tabular}{lcc}
\toprule
\textbf{Method} & \textbf{Test Reconstruction\ Error $\times 10^2$} & \textbf{KL Divergence} \\
\midrule
Stochastic Tsallis & $0.2315$ & $155.14$ \\
Deterministic Gini & $0.2315$ & $155.18$ \\
LapSum & $0.2316$ & $155.24$ \\
Stochastic Gini & $0.2319$ & $155.05$ \\
Deterministic Tsallis & $0.2321$ & $155.06$ \\
FDS & $0.2323$ & $155.15$ \\
Deterministic Shannon & $0.2684$ & $149.45$ \\
SOFT ($\epsilon=0.005$) & $0.2824$ & $222.72$ \\
Stochastic Shannon & $0.3830$ & $156.11$ \\
Hard-ST & $2.0949$ & $1.6741 \times 10^{13}$ \\
Gumbel-ST & $2.1203$ & $1.0777 \times 10^{13}$ \\
\bottomrule
\end{tabular}
\end{table}

\paragraph{Ablation on the Fenchel-Young regularization weight $\beta_{\text{FY}}$.}
On the considered Stacked MNIST task, setting $\beta_{\text{FY}} = 0$ (no discrete regularization) yields test metrics that are nearly identical to our default hyperparameter ($\beta_{\text{FY}} = 10^{-5}$). This reflects a deliberate hyperparameter choice governed by the usual rate-distortion trade-off in VAEs: to ensure a fair comparison against purely deterministic, unregularized Top-$k$ baselines such as FDS and LapSum, we tuned $\beta_{\text{FY}}$ to a small value so that the model could prioritize reconstruction. \cref{tab:dvae_fy_ablation} illustrates this dynamic across increasing values of $\beta_{\text{FY}}$: larger weights enforce the uniform prior over $k$-subsets more strongly, which opposes latent sparsity (increasing the average support size $\|\y\|_0$ toward $n=10$) and degrades test reconstruction error.

Consequently, the primary role of the Fenchel-Young regularization term $L_{\Omegak}(\0\,; \yko(\thetav))$ here is theoretical completeness rather than an empirical gain on this specific benchmark: as shown above, it generalizes the standard discrete KL divergence against a uniform prior used in \citet{ahmed_simple_2023} and extends \citet{sklaviadis_fenchelyoung_2025} to discrete $k$-subset bottlenecks, thereby grounding the model within a proper DVAE framework rather than an unregularized discrete-bottleneck autoencoder.

\begin{table}[h]
\centering
\caption{\textbf{Ablation on the Fenchel-Young regularization weight $\beta_{\text{FY}}$} in the FY-DVAE objective.}
\label{tab:dvae_fy_ablation}
\begin{tabular}{lccc}
\toprule
$\beta_{\text{FY}}$ & \textbf{Test Reconstruction\ Error $\times 10^2$} & \textbf{Fenchel-Young Loss} & \textbf{Average\ $\|\y\|_0$} \\
\midrule
$0.0$ (Ablated) & $0.231$ & $1.31$ & $3.0$ \\
$10^{-5}$ (Ours) & $0.232$ & $1.31$ & $3.0$ \\
$3 \times 10^{-5}$ & $0.236$ & $1.26$ & $3.3$ \\
$5 \times 10^{-5}$ & $0.260$ & $0.81$ & $6.3$ \\
$7 \times 10^{-5}$ & $0.279$ & $0.31$ & $8.8$ \\
$10^{-4}$ & $0.284$ & $0.09$ & $9.6$ \\
\bottomrule
\end{tabular}
\end{table}

\subsection{Experimental details for \cref{sec:experiments_srl}}
\label{sec:srl_appendix}

\paragraph{Problem formulation.} 
The agent manages a store containing $n=20$ items over a finite horizon of $T=80$ time steps. Each item $i$ is assigned a static weight $w_i$ drawn uniformly from $[5]$, a static price $s_i>0$, and possesses a dynamic feature vector $\x^{(t,i)}$. At each time step $t$, the agent selects a subset of items (an assortment) $\y^{(t)} \in\{0,1\}^n$ to display. This selection is subject to two operational constraints:
\begin{enumerate}
    \item \textbf{Knapsack constraint:} the total weight of displayed items must not exceed the store's capacity, i.e.:
    \begin{align*}
        \sum_{i=1}^n w_i \,y^{(t)}_i \leq C\triangleq40.
    \end{align*}
    \item \textbf{Inventory constraint:} an item $i$ may only be displayed if it remains in stock ($I^{(t)}_i \geq 1$). All items are initialized with a uniform inventory of $I^{(0)}_i\triangleq I^{(0)}=5$.
\end{enumerate}
In our precise setting, there are exactly $|\Ycw|=989034$ feasible assortments when every item is in stock.

\paragraph{Customer choice model.} 
Customer behavior is simulated using a multinomial logit model. Given an assortment $\y^{(t)}$, the customer purchases item $i \in \{j\in[n]\mid y^{(t)}_j = 1\}$ or selects the ``no-purchase'' option ($i=0$) according to the following probabilities:
\begin{equation}
    P(i|\y^{(t)}) = \frac{\exp(u^{(t)}_i)}{1 + \sum_{j:y^{(t)}_j = 1} \exp(u^{(t)}_j)},
\end{equation}
where $u^{(t)}_i\!\!\!=\! \langle \phiv, \x^{(t,i)} \rangle$ represents the utility of item $i$ derived from a hidden linear customer model $\phiv$.

\paragraph{Endogenous uncertainty.} 
The environment features strong endogeneity in both feature evolution and state transitions. First, the inventory evolves deterministically: $I^{(t+1)}_i = I^{(t)}_i - 1$ if item $i$ is purchased, creating a coupling between immediate rewards and future availability. Second, item features evolve based on customer decisions. Specifically, if item $i$ is purchased, its ``satisfaction'' feature (part of $\x^{(t,i)}$) increases by $1\%$. Furthermore, a ``hype'' feature evolves based on a rolling window of the last 5 purchases: the most recently purchased item receives a hype boost of $+2\times10^{-2}$, while items purchased $2$ to $5$ steps ago suffer a decay of $-5\times 10^{-3}$. Finally, a time feature increments linearly by $9/T$ at each step.

\paragraph{State space.} 
The state $s_t$ provided to the agent is a concatenation of static and dynamic information resulting in a feature dimension of $12 \times n$. Specifically, the 12-dimensional vector $\x^{(t,i)}$ for item $i$ is constructed as follows: indices $x^{(t,i)}_0, x^{(t,i)}_1$ contain static latent attributes; $x^{(t,i)}_2$ is the dynamic hype; $x^{(t,i)}_3$ is the dynamic satisfaction; $x^{(t,i)}_4$ is the static price $s_i$; and $x^{(t,i)}_5$ is the current time step. These base features are augmented by their dynamics: $x^{(t,i)}_6,x^{(t,i)}_7$ contain the one-step difference in hype and satisfaction, while $x^{(t,i)}_8,x^{(t,i)}_9$ contain their cumulative change since $t=0$. Finally, $x^{(t,i)}_{10}$ contains the normalized inventory $I^{(t)}_i/I^{(0)}$ and $x^{(t,i)}_{11}$ contains the normalized weight $w_i/C$. The reward $r_t$ is defined as the price $s_i$ of the purchased item $i$, or $0$ if no purchase occurs. The objective is to maximize cumulative revenue $\sum_{t=1}^{T} r_t$.

\paragraph{Greedy policy.} As a simple baseline, we include a greedy actor, which always displays available items that maximize the total price:
\begin{align*}
    \y^{(t)}_{\text{greedy}} \triangleq &\argmax_{\y \in \{0,1\}^n} \;\sum_{i=1}^n s_i y_i \\
    &\quad\quad\quad\quad \text{s.t.} \quad \sum_{i=1}^n w_i y_i \leq C, \\
    &\quad\quad\quad\quad \text{and}\quad  \forall i \in [n]\mid y_i =1,\; I^{(t)}_i \geq 1.
\end{align*}
This is effectively implemented using a regular Knapsack operator, by using an artificial weights vector $\hat{\w}^{(t)}$ such that $\hat{w}^{(t)}_i=w_i$ if $I_i^{(t)}\geq 1$, and $\hat{w}^{(t)}_i=C+1$ if $I_i^{(t)}=0$. We then have $\y^{(t)}_{\text{greedy}} = \y^C_{\hat{\w}^{(t)}}(\s)$.

\paragraph{Expert policy.} 
Finding the optimal policy is computationally intractable. We define a myopic expert policy baseline as follows: at each time step $t$, the expert computes the assortment $\y^{(t)}$ that maximizes the expected immediate revenue, subject to current constraints:
\begin{align*}
    \y^{(t)}_{\text{expert}} \triangleq &\argmax_{\y \in \{0,1\}^n} \sum_{i\in[n]\mid y_i=1} s_i \cdot P(i|\y)\\
    &\quad\quad \quad\quad\quad \text{s.t.} \quad \sum_{i=1}^n w_i y_i \leq C, \\
    &\quad\quad \quad\quad\quad \text{and}\quad  \forall i \in [n]\mid y_i =1,\; I^{(t)}_i \geq 1.
\end{align*}
To compute $\y^{(t)}_{\text{expert}}$, we enumerate every feasible item selection (there are $989034$ of them in our configuration), and compute the corresponding purchase probabilities. Naturally, this baseline requires access to hidden information (the customer model $\phiv$ defining purchase probabilities $P(i|\y)$), and is computationally costly.

\paragraph{SRL actors.} The actors in the SRL-based methods are parameterized by a feed-forward neural network $f_{W_1}$ mapping states $\x^{(t)}\in\RR^{12\times n}$ to item logits $\thetav\in\RR^n$, which are then mapped to feasible assortments by the hard Knapsack operator $\y^C_{\hat{\w}^{(t)}}(\thetav)$, where $\hat{w}^{(t)}_i=w_i$ if $I_i^{(t)}\geq 1$, and $\hat{w}^{(t)}_i=C+1$ if $I_i^{(t)}=0$ (effectively enforcing the inventory constraint).

\paragraph{Exploration and gradient computation.} In standard SRL \citep{hoppe_structured_2025}, $K$ exploration targets are generated by perturbing the predicted logits $\thetav$ with a random variable $Z$ (which follows, e.g., a centered Gaussian distribution in $\RR^n$) to get samples:
\begin{align*}
    \y^{(k)}\triangleq \y^C_{\hat{\w}^{(t)}}(\thetav + \varepsilon Z^{(k)})\in \cY^C_{\hat{\w}^{(t)}}\;,
\end{align*}

where $\varepsilon>0$ is a hyperparameter and the $Z^{(k)}$ are i.i.d. samples. Then, the target $\hat{\y}$ is defined by aggregating these samples via a softmax operation using Q-values predicted by a critic network $\psi_{W_2}$:
\begin{align*}
    \hat{\y}\triangleq  \sum_{k=1}^K\frac{\exp\left(\frac{\psi_{W_2}(\x, \y^{(k)})}{\tau}\right)}{\sum_{k'=1}^K \exp\left(\frac{\psi_{W_2}(\x, \y^{(k')})}{\tau}\right)}\cdot \y^{(k)}\in \conv\left(\cY^C_{\hat{\w}^{(t)}}\right).
\end{align*}
Finally, a perturbation-based Fenchel-Young loss \citep{berthet_learning_2020} is used to compute gradients with respect to $\thetav$, which are then backpropagated to upstream weights $W_1$. This gradient is estimated via Monte-Carlo as:
\begin{align*}
    \nabla_\thetav L_\varepsilon(\thetav\,;\hat{\y}) \approx \frac{1}{M}\sum_{m=1}^M \y^C_{\hat{\w}^{(t)}}(\thetav + \varepsilon Z^{(m)}) - \hat{\y},
\end{align*}
where the $Z^{(m)}$ are again i.i.d. samples.

What we propose for our DP-based SRL agents instead, is to use \cref{algo:sample} to produce $K$ exploration actions:
\begin{align*}
    \y^{(k)}\sim \pi^{\hat{\w}^{(t)},C}_{\thetav,\Omega}\in \cY^C_{\hat{\w}^{(t)}}\;,
\end{align*}
and to use our DP-based Fenchel-Young losses to compute exact gradients
\begin{align*}
    \nabla_\thetav L_{\Omega^C_{\hat{\w}^{(t)}}}(\thetav\,;\hat{\y}) = \y^C_{\hat{\w}^{(t)},\Omega}(\thetav) - \hat{\y},
\end{align*}
using \cref{algo:layer}. Importantly, we only need to call \cref{algo:value} once to produce intermediate outputs for \cref{algo:layer,algo:sample}, yielding computational gains.

\paragraph{PPO baseline.}
We adapt proximal policy optimization (PPO) \citep{schulman_proximal_2017} to the combinatorial setting following the COaML-pipeline framework detailed in \citet[Section C.2.2]{hoppe_structured_2025}. In this formulation, the PPO agent does not directly output discrete actions. Instead, the actor network $f_{W_1}$ predicts a latent score vector $\thetav \in \RR^n$, which parameterizes a multivariate Gaussian policy $\pi_{\thetav}(\etav) = \cN(\etav \mid \thetav, \sigma^2 \I)$. During the rollout, a continuous score vector $\etav$ is sampled from this distribution and passed to the hard Knapsack solver to produce the discrete item selection $\y^{(t)} = \y^C_{\hat{\w}^{(t)}}(\etav)$. From the perspective of the PPO algorithm, the actor's action is the sampled continuous vector $\etav$, the combinatorial solver is treated as a deterministic transition function within the environment dynamics, and the optimization objective is to adjust the mean $\thetav$ of the score distribution to maximize returns. The policy ratio and clipping are computed on the continuous densities of the perturbed scores $\etav$. We anneal the exploration standard deviation $\sigma$ linearly during training.

\paragraph{Hyperparameters.}
We train all agents for $10^4$ episodes, performing $10$ update iterations with a batch size of $32$ at the end of each episode for SRL, and $100$ iterations for PPO to compensate for its on-policy data efficiency. We maintain a replay buffer with a capacity of $10^4$ transitions and use the Adam optimizer with a learning rate of $8 \times 10^{-5}$ for all networks. For the SRL agents, we set the perturbation noise scale $\varepsilon$ and the regularization strength $\gamma$ to the same value, annealing it linearly from $5.0$ to $1.0$ over the course of training. We use $K=32$ exploration samples for target action creation, and the softmax aggregation temperature $\tau$ is annealed linearly from $1.0$ to $0.5$. For PPO, we use a clipping ratio of $\epsilon=0.2$, and the exploration standard deviation $\sigma$ is annealed linearly from $2.0$ to $1.0$. All experiments were executed on the CPU of an Apple M1 Max processor with 64 GB of RAM.

\section{Explicit Derivations}
\label{sec:explicit_derivations}

In this section, we provide explicit derivations for the quantities $\Qwo[i,c]$, $\Vwo[i,c]$, and the partial derivatives required to instantiate \cref{algo:value,algo:layer,algo:sample,algo:vjp}. The obtained closed forms are summarized in \cref{tab:instantiations}.

Recall the definition of the smoothed value function $\Vwo[i,c]$ from \cref{eq:smoothed_recursion}:

\begin{align*}
    &\Vwo[i,c] \triangleq \begin{cases}
        \Vwo[i-1, c] &\text{if $w_{i}>c$,}\\[5pt]
        \maxo\bigl(\theta_{i}+\Vwo[i-1, c-w_{i}],\;\Vwo[i-1, c]\bigr) &\text{else.}
    \end{cases}
\end{align*}

We focus on the non-trivial case where $w_i \le c$. We define the local inputs $a$ and $b$ as:
\begin{align*}
    a &\triangleq \theta_i + \Vwo[i-1, c-w_i], \\
    b &\triangleq \Vwo[i-1, c].
\end{align*}
The smoothed maximum is defined as the optimal value of the regularized objective:
\begin{align*}
    \maxo(a, b) \triangleq \max_{q \in [0,1]} J(q\,; a, b), \quad \text{where } J(q\,; a, b) \triangleq q a + (1-q) b - \Omega(\q).
\end{align*}
Here, $\q = (q, 1-q)^\top$ and $\Omega(\q) = \omega(q) + \omega(1-q)$ is a separable strictly convex regularization function. We denote the unique maximizer by $q^*(a, b)$.
The partial derivatives required for the algorithms are given by Danskin's theorem and the chain rule:
\begin{align*}
    \Qwo[i,c] &= q^*(a,b), \\
    \frac{\partial \Qwo[i,c]}{\partial \theta_i} &= \frac{\partial q^*}{\partial a}(a, b).
\end{align*}
In the following, we derive $q^*$ and $\frac{\partial q^*}{\partial a}$ for specific choices of $\Omega$. The resulting explicit expressions are summarized in \cref{tab:instantiations}, together with the corresponding expression of $\Vwo[i,c]=\maxo(a,b)=J(q^*(a,b)\,;a,b)$.

\subsection{General derivative computation}

The optimization problem $\max_{q \in [0,1]} J(q\,; a, b)$ is equivalent to the minimization problem:
\begin{align*}
    \min_{q \in \RR} \quad & \Phi(q) \triangleq \omega(q) + \omega(1-q) - q a - (1-q) b \\
    \text{s.t.} \quad & g_1(q) \triangleq -q \le 0, \nonumber \\
    & g_2(q) \triangleq q - 1 \le 0. \nonumber
\end{align*}
Let $\mu, \lambda \ge 0$ be the KKT multipliers associated with $g_1$ and $g_2$ respectively. The KKT stationarity condition necessitates:
\begin{align}
    \label{eq:general_stationarity}
    \omega'(q^*) - \omega'(1-q^*) - (a-b) - \mu + \lambda = 0.
\end{align}

We now distinguish two regimes for the sensitivity $\frac{\partial q^*}{\partial a}$.

\paragraph{Case 1: interior solution.}
Assume the optimal solution lies in the strict interior $q^* \in (0,1)$. The box constraints are inactive and we have $\mu = \lambda = 0$. The stationarity condition \cref{eq:general_stationarity} simplifies to:
\begin{align*}
    \omega'(q^*)-\omega'(1-q^*) = a - b.
\end{align*}
Differentiating with respect to $a$ yields:
\begin{align}
\label{eq:partial_stationarity}
    \frac{\partial q^*}{\partial a} = \frac{1}{\omega''(q^*)+\omega''(1-q^*)}.
\end{align}

\paragraph{Case 2: boundary solution.}
We now consider the case where the solution saturates at a boundary. This occurs if and only if the regularizer has bounded gradients at the endpoints.
Assume $\lim_{q \to 1^-} \omega'(q) = L < \infty$ and $\lim_{q \to 0^+} \omega'(q) = l > -\infty$.
Let $\tau_\Omega \triangleq L - l$.
We prove that if $a-b \ge \tau_\Omega$, then $q^*=1$ and $\frac{\partial q^*}{\partial a} = 0$.

Let $q^*=1$. The active constraint is $g_2(q) \le 0$, so $\mu=0$ and $\lambda \ge 0$. The stationarity condition \cref{eq:general_stationarity} becomes:
\begin{align*}
    \lambda &= (a-b) - (\omega'(1) - \omega'(0)) \\
    &= (a-b) - \tau_\Omega.
\end{align*}
Dual feasibility requires $\lambda \ge 0$, which holds if and only if $a-b \ge \tau_\Omega$.
Since the problem is strictly convex, the KKT conditions are sufficient for optimality. Thus, for all $(a,b)$ such that $a-b > \tau_\Omega$, the unique minimizer is $q^*=1$.
In this open region, $q^*$ is constant with respect to $a$, implying:
\begin{align*}
    \frac{\partial q^*}{\partial a} = 0.
\end{align*}
A symmetric argument shows that if $a-b \le -\tau_\Omega$, then $q^*=0$ and the derivative is zero.

\subsection{Shannon entropy}
We consider $\Omega(\q) = -\gamma H^{s}(\q) \triangleq \gamma \sum_{i=1}^2q_i\log q_i$, which implies $\omega(q) = \gamma q \log q$. The gradients of $\omega$ become unbounded at the boundaries ($\lim_{q\to 0} \omega'(q) = -\infty$), forcing the solution to be strictly interior for any $a, b\in\RR$.

\paragraph{Maximizer.}

The stationarity condition \cref{eq:general_stationarity} yields:
\begin{align*}
    \gamma \log q^* - \gamma \log(1-q^*) = a-b \iff \gamma \log \left(\frac{q^*}{1-q^*}\right) = a-b.
\end{align*}
Solving for $q^*$ yields the logistic function $q^* = \sigma(\frac{a-b}{\gamma})\triangleq (1+\exp((b-a)/\gamma))^{-1}$.

\paragraph{Partial derivative.}

The derivative is computed via \cref{eq:partial_stationarity} derived in Case 1:
\begin{align*}
    \frac{\partial q^*}{\partial a} = \frac{1}{\omega''(q^*) + \omega''(1-q^*)} = \left( \frac{\gamma}{q^*} + \frac{\gamma}{1-q^*} \right)^{-1} = \frac{q^*(1-q^*)}{\gamma}.
\end{align*}

\paragraph{Smoothed maximum.}
Substituting the optimal probability $q^*$ back into the objective function $J$ yields the standard \emph{log-sum-exp} as expected:
\begin{align*}
    \maxo(a,b) = \gamma \log \left( \exp(a/\gamma) + \exp(b/\gamma) \right).
\end{align*}

\subsection{Gini entropy}
We consider $\Omega(\q) = -\gamma H^{g}(\q) \triangleq \frac{\gamma}{2} (\|\q\|_2^2 - 1)$, which implies $\omega(q) = \frac{\gamma}{2} q^2 - \frac{\gamma}{4}$.

\paragraph{Maximizer.}

In this case, we have:
\begin{align*}
    \omega'(q) - \omega'(1-q) = \gamma q - \gamma(1-q) = 2\gamma q - \gamma,
\end{align*}
so that \cref{eq:general_stationarity} becomes:
\begin{align*}
    2\gamma q^* - \gamma - (a-b) - \mu + \lambda = 0 \implies q^* = \frac{a-b+\gamma + \mu - \lambda}{2\gamma}.
\end{align*}
We analyze the solution based on the active constraints using the complementary slackness conditions $\mu q^* = 0$ and $\lambda (q^* - 1) = 0$.

\begin{enumerate}
    \item \textbf{Inactive constraints ($\mu=0, \lambda=0$):}
    we have $q^* = \frac{a-b+\gamma}{2\gamma}$.
    Primal feasibility $0 < q^* < 1$ requires $0 < \frac{a-b+\gamma}{2\gamma} < 1$, which simplifies to $-\gamma < a-b < \gamma$.
    In this region, $\frac{\partial q^*}{\partial a} = \frac{1}{2\gamma}$.

    \item \textbf{Active lower bound ($q^*=0$):}
    we have $\lambda = 0$ and $\mu \ge 0$.
    The stationarity condition yields $0 = \frac{a-b+\gamma + \mu}{2\gamma}$, which implies $\mu = -(a-b+\gamma)$.
    Dual feasibility $\mu \ge 0$ requires $a-b+\gamma \le 0 \iff a-b \le -\gamma$.
    In this region, $q^*$ is constant, so $\frac{\partial q^*}{\partial a} = 0$.

    \item \textbf{Active upper bound ($q^*=1$):}
    we have $\mu = 0$ and $\lambda \ge 0$.
    The stationarity condition yields $1 = \frac{a-b+\gamma - \lambda}{2\gamma}$, which implies $2\gamma = a-b+\gamma - \lambda$, or $\lambda = a-b-\gamma$.
    Dual feasibility $\lambda \ge 0$ requires $a-b-\gamma \ge 0 \iff a-b \ge \gamma$.
    In this region, $q^*$ is constant, so $\frac{\partial q^*}{\partial a} = 0$.
\end{enumerate}

Combining these three cases, we recover the projection onto $[0,1]$:
\begin{align*}
    q^* = \clip_{[0,1]}\left( \frac{a-b+\gamma}{2\gamma} \right).
\end{align*}

\paragraph{Partial derivative.}

The partial derivative is constant and only non-zero on the non-clipped region:
\begin{align*}
    \frac{\partial q^*}{\partial a} = \frac{1}{2\gamma} \cdot \mathbbm{1}_{\{|a-b| < \gamma\}}.
\end{align*}

\paragraph{Smoothed maximum.}

The regularization term simplifies to $\Omega(\q) = \gamma(q^2 - q)$ for any $\q=(q,(1-q))^\top$. Substituting the optimal $q^*$ into the objective function $J$ yields:
\begin{align*}
    \maxo(a,b) &= q^* a + (1-q^*) b - \gamma((q^*)^2 - q^*) \\
    &= (a-b)q^* + b - \gamma q^*(q^* - 1).
\end{align*}
When saturation occurs (i.e., $|a-b| \geq \gamma$), we have $q^* \in \{0, 1\}$. In this case, the regularization term vanishes and we recover the hard maximum $\max(a,b)$. In the interior region (i.e., $|a-b| < \gamma$), substituting $q^* = \frac{a-b+\gamma}{2\gamma}$ yields:
\begin{align*}
\maxo(a,b) &= (a-b)\left(\frac{a-b+\gamma}{2\gamma}\right) + b - \gamma \left(\frac{a-b+\gamma}{2\gamma}\right)\left(\frac{a-b-\gamma}{2\gamma}\right) \\
&= \frac{(a-b)^2 + \gamma(a-b)}{2\gamma} + b - \gamma\left(\frac{(a-b)^2 - \gamma^2}{4\gamma^2}\right) \\
&= \frac{(a-b)^2}{2\gamma} + \frac{a-b}{2} + b - \frac{(a-b)^2}{4\gamma} + \frac{\gamma^2}{4\gamma} \\
&= \frac{(a-b)^2}{4\gamma} + \frac{a+b}{2} + \frac{\gamma}{4}.
\end{align*}
\vspace{1cm}

\subsection{1.5-Tsallis Entropy}
We consider $\Omega(\q) = -\gamma H_{1.5}^{T}(\q) \triangleq \gamma\sum_{i=1}^2\frac{q_i^{\frac{3}{2}}-q_i}{\frac{3}{2}(\frac{3}{2}-1)}$, i.e., $\omega(q) = \frac{4\gamma}{3} q^{\frac{3}{2}} - \frac{2\gamma}{3}$ and $\omega'(q) = 2\gamma \sqrt{q}$.

\paragraph{Maximizer.}

The stationarity condition for an interior solution ($q^* \in (0,1)$) requires:
\begin{align*}
    2\gamma\sqrt{q^*} - 2\gamma\sqrt{1-q^*} = a-b.
\end{align*}
Let $C \triangleq (a-b)/2\gamma$. We seek to solve $\sqrt{q^*} - \sqrt{1-q^*} = C$.
The function $f(q) = \sqrt{q} - \sqrt{1-q}$ is strictly increasing on $[0,1]$ with range $[-1, 1]$.
\begin{itemize}
    \item If $|C| > 1$ (i.e., $|a-b| > 2\gamma$), no interior solution exists. The monotonicity of the objective forces the solution to the boundary: $q^*=1$ if $C > 1$, and $q^*=0$ if $C < -1$.
    \item If $|C| \le 1$, we square the stationarity equation: $q^* - 2\sqrt{q^*(1-q^*)} + (1-q^*) = C^2$, which rearranges to $2\sqrt{q^*(1-q^*)} = 1 - C^2$. Squaring again yields the quadratic equation $4(q^*)^2 - 4q^* + (1 - C^2)^2 = 0$.
    The roots are $q_{\pm} = (1 \pm |C|\sqrt{2 - C^2})/2$.
    To satisfy the sign of the original equation $\sqrt{q^*} - \sqrt{1-q^*} = C$, we must select the root such that $\mathrm{sgn}(\sqrt{q^*} - \sqrt{1-q^*})=\mathrm{sgn}(q^* - \frac{1}{2}) = \mathrm{sgn}(C)$. Since $q_+\geq \frac{1}{2}$ and $q_-\leq \frac{1}{2}$, we must then choose $q_+$ when $C\geq 0$ and $q_-$ when $C\leq 0$. This yields $q^* = \frac{1 + C\sqrt{2 - C^2}}{2}$.
\end{itemize}
Combining the boundary and interior cases and denoting $\bar{C} \triangleq \clip_{[-1,1]}(C)$, the closed-form solution is $q^* = \frac{1}{2} \left( 1 + \bar{C}\sqrt{2 - \bar{C}^2} \right)$.

\paragraph{Partial derivative.}

Using $\omega''(q) = \gamma q^{-\frac{1}{2}}$, the derivative in the interior (where $|\bar{C}| < 1$) is:
\begin{align*}
    \frac{\partial q^*}{\partial a} = \left(\frac{\gamma}{\sqrt{q^*}} + \frac{\gamma}{\sqrt{1-q^*}}\right)^{-1}.
\end{align*}
If $|\bar{C}| = 1$, the derivative is 0.

\paragraph{Smoothed maximum.}

The value is obtained by substituting the optimal $q^*$ into the objective. Using the definition $\Omega(\q) = \frac{4\gamma}{3} (q^{\frac{3}{2}} + (1-q)^{\frac{3}{2}} - 1)$, we obtain:
\begin{align*}
    \maxo(a,b) &= q^* a + (1-q^*) b - \Omega(\q^*)= q^* a + (1-q^*) b + \frac{4\gamma}{3}\left( 1 - (q^*)^{\frac{3}{2}} - (1-q^*)^{\frac{3}{2}} \right).
\end{align*}
Unlike the Gini case, this expression does not admit a simple polynomial form in terms of $a$ and $b$, and is computed using the closed-form solution for $q^*$ derived above.

\begin{table*}
\caption{Instantiations of the quantities required for \cref{algo:value} and \cref{algo:vjp}.\\
We use $\Delta_{i,c}\triangleq\Vwo[i-1, c] - (\theta_i + \Vwo[i-1, c-w_i])$, and $\sigma(t)\triangleq(1+\exp(-t))^{-1}$ denotes the logistic function.
For the $1.5$-Tsallis case, we use the intermediate variable $\bar{C} \triangleq \clip_{[-1,1]}\left(-\Delta_{i,c}/2\gamma\right)$.}
\label{tab:instantiations}
\begin{center}
\begin{small}
\begin{sc}
\renewcommand{\arraystretch}{2.2}
\setlength{\tabcolsep}{8pt}
\scalebox{0.9}{
    \begin{tabular}{|l|c|c|c|}
    \hline
    & \textbf{Shannon ($\Omega=-\gamma H^{s}$)}
    & \textbf{Gini ($\Omega=-\gamma H^{g}$)}
    & \textbf{1.5-Tsallis ($\Omega = -\gamma H_{1.5}^{T}$)} \\
    \hline
    $\Qwo[i,c]$
    & $\sigma(-\Delta_{i,c}/\gamma)$
    & $\clip_{[0,1]}\left( \frac{-\Delta_{i,c} + \gamma}{2\gamma} \right)$
    & $\frac{1}{2}\left(1 + \bar{C}\sqrt{2-\bar{C}^2}\right)$
    \\
    \hline
    $\Vwo[i,c]$
    & \makecell{$\Vwo[i-1,c] $ \\$+ \gamma \log\left(1 + \exp(-\Delta_{i,c}/\gamma)\right)$}
    & \makecell{$\Vwo[i-1,c] - \Delta_{i,c}\Qwo[i,c]$ \\ $+ \gamma\Qwo[i,c](1 - \Qwo[i,c])$}
    & \makecell{$\Vwo[i-1,c] - \Delta_{i,c}\Qwo[i,c]$ \\ $+ \frac{4\gamma}{3}\biggl(1 - \Qwo[i,c]^{\frac{3}{2}} $ \\ $- (1-\Qwo[i,c])^{\frac{3}{2}}\biggr)$}
    \\
    \hline
    $\frac{\partial \Qwo[i,c]}{\partial \theta_i}$
    & $\frac{1}{\gamma}\Qwo[i,c](1-\Qwo[i,c])$
    & $\frac{1}{2\gamma}\mathbbm{1}_{\{0 < \Qwo[i,c] < 1\}}$
    & $\frac{1}{\gamma}\left(\frac{1}{\sqrt{\Qwo[i,c]}} + \frac{1}{\sqrt{1-\Qwo[i,c]}}\right)^{-1}$
    \\
    \hline
    $\frac{\partial \Qwo[i,c]}{\partial \Vwo[i-1, c]}$
    & $-\frac{1}{\gamma}\Qwo[i,c](1-\Qwo[i,c])$
    & $-\frac{1}{2\gamma}\mathbbm{1}_{\{0 < \Qwo[i,c] < 1\}}$
    & $-\frac{1}{\gamma}\left(\frac{1}{\sqrt{\Qwo[i,c]}} + \frac{1}{\sqrt{1-\Qwo[i,c]}}\right)^{-1}$
    \\
    \hline
    \end{tabular}
} %
\end{sc}
\end{small}
\end{center}
\vskip -0.1in
\end{table*}

\end{document}